\PassOptionsToPackage{table}{xcolor}
\documentclass[10pt]{article}

\usepackage[preprint]{tmlr}

\usepackage{amsmath,amsfonts,bm}

\def\eqref#1{equation~\ref{#1}}

\def\1{\bm{1}}

\DeclareMathAlphabet{\mathsfit}{\encodingdefault}{\sfdefault}{m}{sl}
\SetMathAlphabet{\mathsfit}{bold}{\encodingdefault}{\sfdefault}{bx}{n}

\newcommand{\E}{\mathbb{E}}

\newcommand{\Var}{\mathrm{Var}}

\newcommand{\Cov}{\mathrm{Cov}}

\DeclareMathOperator*{\argmin}{arg\,min}

\usepackage[utf8]{inputenc} 
\usepackage{hyperref}       
\usepackage{url}            
\usepackage{booktabs}       
\usepackage{amsfonts}       
\usepackage{nicefrac}       
\usepackage{microtype}      
\usepackage{xcolor}         
\usepackage{graphicx}

\usepackage{amsmath}
\usepackage{amssymb}
\usepackage{amsthm}
\usepackage{bm}
\usepackage{enumitem}
\usepackage{mathtools}
\usepackage{multirow}
\usepackage[percent]{overpic}
\usepackage{pifont}
\usepackage{caption}
\usepackage[font=small]{subcaption}
\usepackage{tikz}
\usepackage{wrapfig}
\usepackage{algorithm}
\usepackage{algpseudocode}
\usepackage{etoc}
\definecolor{appendixtocblue}{RGB}{63,145,190}
\makeatletter
\@ifundefined{theHALG@line}
  {\newcommand{\theHALG@line}{\thealgorithm.\arabic{ALG@line}}}
  {\renewcommand{\theHALG@line}{\thealgorithm.\arabic{ALG@line}}}
\makeatother

\usepackage[capitalize,noabbrev]{cleveref}

\theoremstyle{plain}

\theoremstyle{definition}

\theoremstyle{remark}

\newcommand{\dd}[2]{\frac{d #1}{d #2}}
\newcommand{\pp}[2]{\frac{\partial #1}{\partial #2}}

\newcommand{\diag}{\mathrm{diag}}

\title{Flow Matching with Uncertainty Quantification and Guidance}

\author{\name Juyeop Han \email juyeop@mit.edu \\
      \addr MIT
      \AND
      \name Lukas Lao Beyer \email llb@mit.edu \\
      \addr MIT
      \AND
      \name Sertac Karaman \email sertac@mit.edu \\
      \addr MIT}

\def\month{MM}
\def\year{YYYY}
\def\openreview{\url{https://openreview.net/forum?id=XXXX}}

\begin{document}

\etocsettagdepth{main}{part}
\etocsettagdepth{appendix}{subsection}
\etocdepthtag.toc{main}

\maketitle

\begin{abstract}
Despite the remarkable success of sampling-based generative models such as flow matching, they can still produce samples of inconsistent or degraded quality.
To assess sample reliability and improve sample fidelity, we propose \textbf{uncertainty-aware flow matching (UA-Flow)}, a lightweight extension of flow matching that predicts the velocity field together with heteroscedastic uncertainty.
UA-Flow estimates spatially localized sample uncertainty by propagating velocity uncertainty through the flow dynamics.
The resulting estimates serve as reliability signals that reflect data-manifold support for each generated sample or region.
We further use them to steer generation via uncertainty-aware guidance (UG) and uncertainty-aware classifier-free guidance (UCFG).
Experiments on ImageNet and CIFAR-10 show that UA-Flow’s uncertainty tracks sample fidelity more consistently than element-wise baselines and that these signals also carry information about data-manifold support.
On ImageNet, UG increases precision at the cost of recall and improves FID, while UCFG maintains lower FID than CFG at large guidance scales.
\end{abstract}

\section{Introduction}
\label{sec:introduction}

In recent years, sampling-based generative models such as diffusion~\citep{Sohl15Deep, Ho20Denoising, song21score} and flow matching~\citep{Lipman23Flowmatching, Liu23Rectified, Albergo23Interpolants} have achieved remarkable success across a wide range of domains, most particularly in image and video generation~\citep{Dhariwal21Diffusion, Ho22Video} as well as in sequential decision-making~\citep{Janner22Diffuser, Chi25Diffusionpolicy, Black24pi0}.
Despite this progress, these models often produce samples of inconsistent quality.
As a result, using reliably generated samples in downstream applications remains challenging.


To address this issue, the uncertainty associated with each generated sample can be interpreted as a measure of the reliability of the generation process.
Several works have explored uncertainty estimation for diffusion-based generative models by adapting techniques from existing uncertainty quantification literature for neural networks~\citep{Lakshminarayanan17Ensemble, Kendall17Uncertainties, Ritter18Laplace}.
In sampling-based generative modeling, uncertainty plays two central roles:
(i) it provides a signal for assessing the quality or reliability of individual generated samples,
and (ii) it can guide the generation process to improve sample quality~\citep{DeVita25Aleatoric}.
However, most prior works~\citep{Kou23Bayesdiff,Jazbec25Genunc} primarily focus on the first role, using uncertainty in a post-hoc manner for sample filtering or selection.
Moreover, existing approaches are often domain-specific~\citep{Sun23Conformal, Franchi2025PUNC} and quantify uncertainty over conditional inputs~\citep{Berry2024DECU}.

We propose \textbf{uncertainty-aware flow matching (UA-Flow)}, a lightweight extension of flow matching that models heteroscedastic uncertainty in the velocity field.
The discrepancy between the learned and true velocity fields is probabilistically modeled as velocity uncertainty.
By propagating this velocity uncertainty through the flow dynamics, UA-Flow provides principled sample-level uncertainty estimates with minimal additional overhead.
Because uncertainty is modeled element-wise at the velocity level of general flow matching, our approach provides spatially localized uncertainty within each generated sample.
Unlike methods that depend on domain-specific adaptations such as the CLIP encoder used by~\citep{Jazbec25Genunc}, this formulation is, in principle, not specialized to a particular data domain.
Moreover, we can use the learned velocity uncertainty for uncertainty-aware guided sampling, which improves generation quality and is not explicitly considered in closely related prior work~\citep{Kou23Bayesdiff,Jazbec25Genunc}.
The deterministic sampling allows us to localize uncertainty to the learned velocity field and propagate it through the dynamics, in contrast to stochastic sampling.

We empirically validate the uncertainty estimates produced by our proposed approach in two settings across CIFAR-10, ImageNet-128 and ImageNet-256.
First, we provide evidence that UA-Flow’s uncertainty correlates with sample fidelity and the degree of support from the training data, with higher uncertainty indicating lower fidelity and weaker data support.
In particular, filtering out high-uncertainty samples yields better fidelity-oriented metrics than prior uncertainty-quantification baselines for sampling-based generative models~\citep{Kou23Bayesdiff, DeVita25Aleatoric}.
Moreover, high uncertainty in generated samples or regions is associated with greater distance from the data manifold and higher data rarity.
Second, we use predicted velocity uncertainty to guide sampling through uncertainty-aware guidance (UG) and uncertainty-aware classifier-free guidance (UCFG).
Our experiments show that UG improves precision at the cost of recall and reduces FID on ImageNet.
UCFG improves the best ImageNet FID over fixed CFG and limits degradation at large guidance scales.

\begin{figure}[t]
	\begin{subfigure}[t]{0.44\textwidth}
		\centering
		\begin{tikzpicture}
			\node[anchor=south west, inner sep=0] (img) at (0,0) {%
				\includegraphics[width=\textwidth]{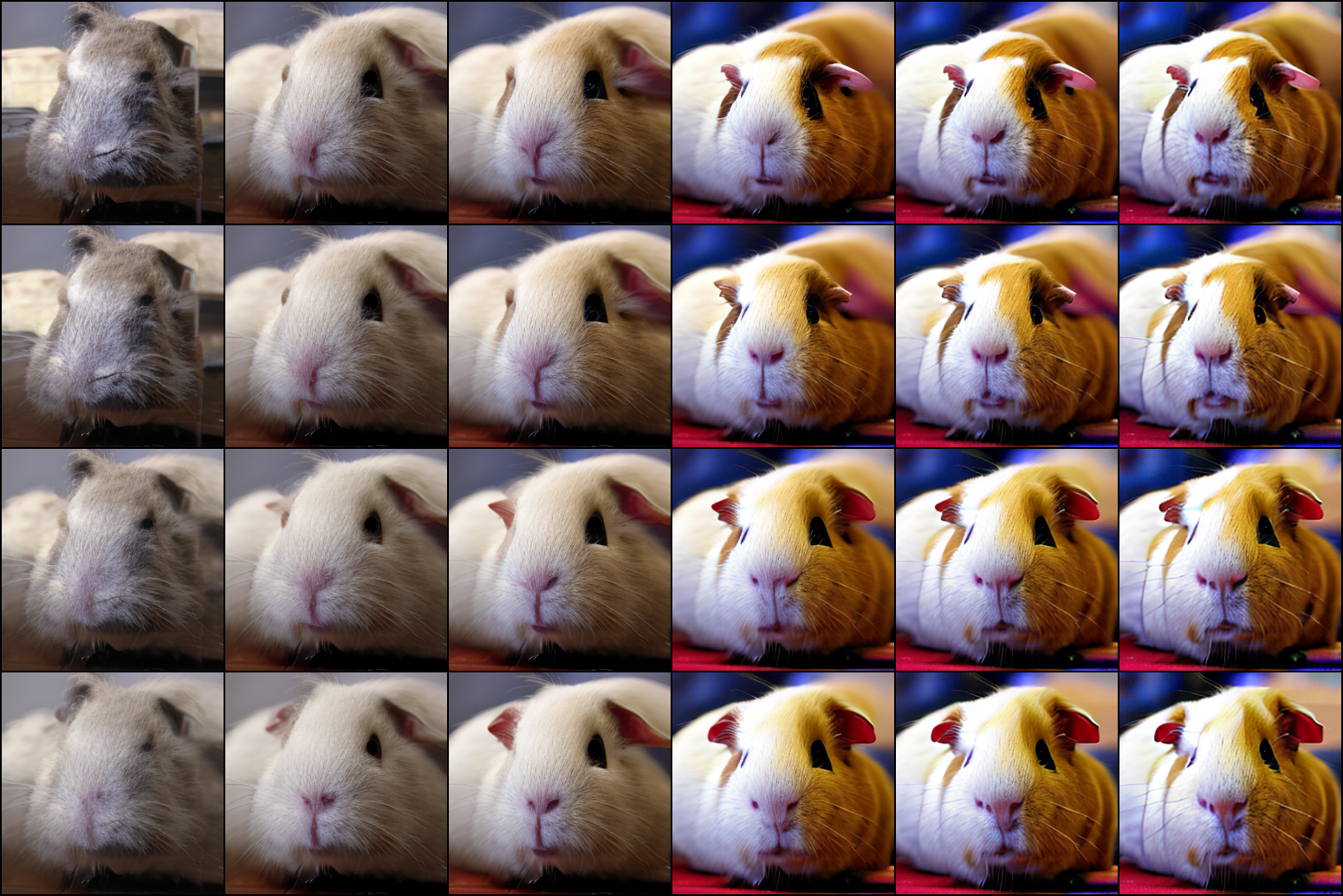}
			};
			\begin{scope}[x={(img.south east)}, y={(img.north west)}]
				
				\draw[->, line width=0.4pt] (0.03,1.03) -- (0.97,1.03);
				\node[anchor=south] at (0.50,1.04) {\small Max. UCFG scale $\lambda_{\max}$};
				\foreach \x/\lab in {0.083/0, 0.250/1, 0.417/2, 0.583/5, 0.750/10, 0.917/20}{
					\node[fill=white, inner sep=0.6pt] at (\x, 1.03) {\small $\lab$};
				}
				
				\draw[->, line width=0.4pt] (-0.02,0.95) -- (-0.02,0.05);
				\node[rotate=90, anchor=south] at (-0.040,0.50) {\small UG scale $w$};
				\foreach \y/\lab in {0.875/0, 0.625/10, 0.375/30, 0.125/50}{
					\node[fill=white, inner sep=0.6pt] at (-0.02,\y) {\rotatebox{90}{\small $\lab$}};
				}
			\end{scope}
		\end{tikzpicture}
		\caption{Generated samples.}
		\label{fig:cover_guidance_samples}
	\end{subfigure}
	\hspace{2.5em}
	\begin{subfigure}[t]{0.44\textwidth}
		\centering
		\begin{tikzpicture}
			\node[anchor=south west, inner sep=0] (img) at (0,0) {%
				\includegraphics[width=\textwidth]{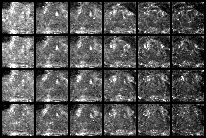}
			};
			\begin{scope}[x={(img.south east)}, y={(img.north west)}]
				\draw[->, line width=0.4pt] (0.03,1.03) -- (0.97,1.03);
				\node[anchor=south] at (0.50,1.04) {\small Max. UCFG scale $\lambda_{\max}$};
				\foreach \x/\lab in {0.083/0, 0.250/1, 0.417/2, 0.583/5, 0.750/10, 0.917/20}{
					\node[fill=white, inner sep=0.6pt] at (\x, 1.03) {\small $\lab$};
				}
				
				\draw[->, line width=0.4pt] (-0.02,0.95) -- (-0.02,0.05);
				\node[rotate=90, anchor=south] at (-0.040,0.50) {\small UG scale $w$};
				\foreach \y/\lab in {0.875/0, 0.625/10, 0.375/30, 0.125/50}{
					\node[fill=white, inner sep=0.6pt] at (-0.02,\y) {\rotatebox{90}{\small $\lab$}};
				}
			\end{scope}
		\end{tikzpicture}
		\caption{Latent pixel-wise uncertainties.}
		\label{fig:cover_guidance_uncertainty}
	\end{subfigure}
	\caption{
		\textbf{UG and UCFG sweep on ImageNet-256 (class \texttt{guinea pig}).}
		Left: generated samples; right: predicted latent pixel-wise uncertainty (brighter indicates higher uncertainty).
		Rows sweep the UG scale $w \in \{0, 10, 30, 50\}$, columns sweep the maximum UCFG scale $\lambda_{\max} \in \{0, 1, 2, 5, 10, 20\}$.
		Stronger guidance yields more class-consistent samples with lower predicted uncertainty.
	}
	\label{fig:cover_guidance}
\end{figure}

\section{Background}
\label{sec:background}

\textbf{Sampling-based generative models and guidance.}
Sampling-based generative models synthesize data by iteratively transforming samples from a simple base distribution into the data distribution.
Diffusion models implement this transformation through a gradual denoising process~\citep{Sohl15Deep,Ho20Denoising,song21score}.
Flow matching instead learns a deterministic ordinary differential equation (ODE) whose velocity field transports samples along a prescribed probability path~\citep{Lipman23Flowmatching,Liu23Rectified,Albergo23Interpolants}.
Across both families, guidance modifies the sampling dynamics to favor samples that better satisfy a condition.
Classifier guidance (CG) injects gradients from an external classifier or constraint functions into the sampling update~\citep{Dhariwal21Diffusion,Dao23Latent}, while classifier-free guidance (CFG) extrapolates conditional and unconditional predictions without a separate classifier~\citep{Ho22CFG,Zheng23Guided}.
A practical challenge is that strong guidance can reduce diversity or induce artifacts at high scales, motivating approaches that moderate the effective guidance signal or adapt its strength during sampling~\citep{Saharia22Imagen,Sadat24APG}.

\textbf{Uncertainty quantification for neural networks.}
A common approach to modeling aleatoric uncertainty is heteroscedastic regression, where the network jointly predicts a mean and an input-dependent variance under a Gaussian negative log-likelihood~\citep{Nix94Estimating,Kendall17Uncertainties}.
Epistemic uncertainty is typically approximated via Bayesian-inspired techniques~\citep{Gal16Dropout,Lakshminarayanan17Ensemble,Ritter18Laplace}.
While standard for classification and regression, applying these to sampling-based generative models requires specifying \emph{which} intermediate quantities are uncertain (e.g., score/velocity or conditional input) and \emph{how} uncertainty propagates through the sampling dynamics.

\textbf{Uncertainty quantification with sampling-based generative models.}
Most existing uncertainty quantification (UQ) methods for sampling-based generative models are diffusion-centric and can be broadly grouped by \emph{where} uncertainty is modeled.
Some quantify uncertainty in the conditional input of conditional diffusion~\citep{Berry2024DECU}.
Others develop diffusion-model uncertainty for domain-specific generation and decision-making, including trajectory planning and multi-agent forecasting~\citep{Sun23Conformal,Capellera25CVPR}, as well as text-to-image uncertainty analysis~\citep{Franchi2025PUNC}.
A third line targets uncertainty of the generated sample itself.
BayesDiff~\citep{Kou23Bayesdiff} uses Bayesian/Laplace-based estimators~\citep{Daxberger21Laplace} and propagates uncertainty through the diffusion dynamics.
Other methods rely on feature-space likelihoods based on CLIP encoders~\citep{Radford21CLIP,Jazbec25Genunc}, or pixel-wise aleatoric uncertainty for reliability scoring and uncertainty-guided sampling~\citep{DeVita25Aleatoric}.
Despite this progress, flow-matching-specific UQ remains underexplored, even though its deterministic ODE structure enables modeling uncertainty in the learned velocity field and propagating it through the dynamics.

\section{Uncertainty-Aware Flow Matching}
\label{sec:uncflow}

Flow matching learns a time-dependent velocity field that transports samples from a base distribution $\mathbf{x}_0\sim p_0$ to data $\mathbf{x}_1\sim p_1$ along a prescribed probability path.
We adopt the common affine path $\mathbf{x}_t = \alpha_t \mathbf{x}_1 + \beta_t \mathbf{x}_0$ over time $t \in [0, 1]$, which induces the conditional distribution $p_t(\mathbf{x}_t\mid \mathbf{x}_1)$ and the corresponding closed-form conditional target velocity $u_t(\mathbf{x}_t\mid \mathbf{x}_1)$~\citep{Lipman24Flowcode}.

Here, we interpret velocity uncertainty as the time and the state-dependent discrepancy between the learned mean velocity and the true marginal velocity.
If the learned mean exactly recovered the marginal velocity, this discrepancy and the modeled uncertainty would vanish.
In practice, however, various factors, such as finite training data, limited model capacity, and optimization error, produce local velocity errors.
As these errors accumulate along the ODE trajectory, they can divert generated samples away from regions supported by the data distribution.
UA-Flow models these local discrepancies probabilistically and propagates them through the flow dynamics to quantify sample-level uncertainty.

\subsection{Probabilistic Velocity Field Modeling}
\label{subsec:learning_mean_variance}
UA-Flow aims to learn the mean, $\bar{u}^{\theta}_t(\mathbf{x}_t) \in \mathbb{R}^n$, and the diagonal variance, $(\sigma_t^\theta(\mathbf{x}_t))^2 \in \mathbb{R}^n$, of the velocity field, using the target velocity $u_t(\mathbf{x}_t)$ as supervision.
We use \emph{velocity uncertainty} to refer to this predicted element-wise variance.
These predictions define a Gaussian velocity model, $u_t^\theta(\mathbf{x}_t)=\bar{u}_t^\theta(\mathbf{x}_t)+\sigma_t^\theta(\mathbf{x}_t)\odot\epsilon$ with $\epsilon\sim\mathcal{N}(0,I)$ and element-wise multiplication $\odot$.

For computational efficiency and representational simplicity, we estimate the variance element-wise.
This is common practice in uncertainty quantification for neural networks~\citep{Kendall17Uncertainties}.
We formulate the uncertainty-aware flow matching loss, $\mathcal{L}_{\mathrm{UFM}}(\theta)$, as a Gaussian negative log-likelihood (NLL) loss for the target velocity:
\begin{equation}
\label{eq:ufm}
\mathcal{L}_{\mathrm{UFM}}(\theta)
= \E_{t,p_t(\mathbf{x}_t)}\Big[
\frac{\big(\bar{u}^{\theta}_t(\mathbf{x}_t) - u_t(\mathbf{x}_t)\big)^2}{2(\sigma_t^\theta(\mathbf{x}_t))^2} + \log(\sigma_t^\theta(\mathbf{x}_t))
\Big].
\end{equation}
All operations in $\mathcal{L}_{\mathrm{UFM}}$ are applied element-wise.
This convention is used throughout the paper.

As in standard flow matching, the target velocity $u_t(\mathbf{x}_t)$ is not directly accessible, and training instead regresses the model to the conditional velocity $u_t(\mathbf{x}_t \mid \mathbf{x}_1)$ under $p_t(\mathbf{x}_t \mid \mathbf{x}_1)$.
Following the same principle, UA-Flow minimizes a conditional uncertainty-aware flow matching loss, denoted by $\mathcal{L}_{\mathrm{CUFM}}$, which reformulates $\mathcal{L}_{\mathrm{UFM}}$ in terms of the conditional velocity:
\begin{equation}
	\label{eq:cufm}
		\mathcal{L}_{\mathrm{CUFM}}(\theta) = 
		\mathbb{E}_{t,p_1(\mathbf{x}_1), p_t(\mathbf{x}_t \mid \mathbf{x}_1)}\Big[
		\frac{U_t(\mathbf{x}_t, \mathbf{x}_1)}{2(\sigma_t^\theta(\mathbf{x}_t))^2} +
		\frac{\big(\bar{u}^{\theta}_t(\mathbf{x}_t) - u_t(\mathbf{x}_t \mid \mathbf{x}_1)\big)^2}{2(\sigma_t^\theta(\mathbf{x}_t))^2} +
		\log(\sigma_t^\theta(\mathbf{x}_t))
		\Big],
\end{equation}
Here, $U_t(\mathbf{x}_t, \mathbf{x}_1):= \hat{u}_t(\mathbf{x}_t)^2 - u_t(\mathbf{x}_t \mid \mathbf{x}_1)^2$ approximates the exact element-wise correction
$U_t^\star(\mathbf{x}_t, \mathbf{x}_1) := u_t(\mathbf{x}_t)^2 - u_t(\mathbf{x}_t \mid \mathbf{x}_1)^2$.
In the mean squared error objective used for standard flow matching~\citep{Lipman23Flowmatching}, the analogous correction is independent of the parameter $\theta$ and can therefore be omitted without changing the optimizer.
In our Gaussian NLL objective, however, the correction is weighted by $1/[2(\sigma_t^\theta(\mathbf{x}_t))^2]$ and thus depends on the parameter $\theta$.

Because the marginal velocity $u_t(\mathbf{x}_t)$ is intractable, we approximate it using the following reweighted mini-batch estimator:
\begin{equation}
	\label{eq:u_hat}
	\hat{u}_t(\mathbf{x}_t) =
	\frac{\sum_{b=1}^B u_t(\mathbf{x}_t \mid \mathbf{x}_{1,b}) p_t(\mathbf{x}_t \mid \mathbf{x}_{1,b})}{\sum_{b=1}^B p_t(\mathbf{x}_t \mid \mathbf{x}_{1,b})}.
\end{equation}
Appendix~\ref{app:loss_deriv} details the derivation of $\mathcal{L}_{\mathrm{CUFM}}(\theta)$ as an approximation to $\mathcal{L}_{\mathrm{UFM}}(\theta)$ and shows that retaining $U_t(\mathbf{x}_t, \mathbf{x}_1)$ yields a closer surrogate than setting it to zero, even though $\hat{u}_t(\mathbf{x}_t)^2$ is not an unbiased estimator of $u_t(\mathbf{x}_t)^2$.
We empirically observe that the correction is largest early in sampling, and larger mini-batches improve its estimate on ImageNet-256 (\cref{subsec:correction_estimator_bias}).

\textbf{Remark.}
UA-Flow can be fine-tuned from a pre-trained flow matching model.
We find that this preserves generation quality as the model learns uncertainty.
For better mean estimates, we use the $\beta$-NLL loss~\citep{Seitzer22Betanll} with the factor $\mathrm{sg}\!\left[(\sigma_t^\theta(\mathbf{x}_t))^{2\beta}\right]$ for each element-wise NLL term.
Here, $\mathrm{sg}$ denotes stop-gradient and $\beta\in[0,1]$.

\subsection{Uncertainty Propagation through Flow Dynamics}
\label{subsec:uncertainty_propagation}

We aim to estimate the uncertainty of a generated sample from accumulated velocity uncertainty along the flow.
The network predicts the mean and variance $\bar u_t^\theta(\mathbf{x}_t)$ and $( \sigma_t^\theta(\mathbf{x}_t) )^2$ of $u_t^\theta(\mathbf{x}_t)$.
We first draw the initial state $\mathbf{x}_0$ from the base distribution $p_0$.
For each sample, we fix this initial state and set $\bar{\mathbf{x}}_0=\mathbf{x}_0$ and $\Var[\mathbf{x}_0]=\mathbf{0}$.
We then propagate the state mean $\bar{\mathbf{x}}_t$ and variance $\Var[\mathbf{x}_t]$.
The variance describes uncertainty from the Gaussian velocity model for this fixed initial state.

We interpret the endpoint mean $\bar{\mathbf{x}}_1$ as the generated sample and the propagated element-wise variance $\Var[\mathbf{x}_1]$ as its \emph{sample uncertainty map}.
The map retains uncertainty for each coordinate.
A \emph{sample-level uncertainty score} is a scalar derived from this map or from samples of the predicted endpoint distribution.
\Cref{sec:experiments} specifies the score for each experiment.

Both the mean and covariance approximations use the same first-order Taylor expansion about $\bar{\mathbf{x}}_t=\E[\mathbf{x}_t]$:
\begin{equation}
	\label{eq:mean_velocity_linearization}
	\bar u_t^\theta(\mathbf{x}_t)
	=\bar u_t^\theta(\bar{\mathbf{x}}_t)
	+J_t^\theta(\bar{\mathbf{x}}_t)(\mathbf{x}_t-\bar{\mathbf{x}}_t)
	+O(\|\mathbf{x}_t-\bar{\mathbf{x}}_t\|^2).
\end{equation}
Here, $J^{\theta}_t(\bar{\mathbf{x}}_t)\in\mathbb{R}^{n\times n}$ denotes the Jacobian of the mean velocity with respect to the state evaluated at $\bar{\mathbf{x}}_t$.
The linear term vanishes in expectation, and the higher-order term is neglected to obtain the mean dynamics
\begin{equation}
	\label{eq:mean_ode}
	\dd{\bar{\mathbf{x}}_t}{t}  = \bar{u}^{\theta}_t(\bar{\mathbf{x}}_t).
\end{equation}

For variance propagation, we use Euler discretization, $\mathbf{x}_{t+\Delta t}=\mathbf{x}_t + u^\theta_t(\mathbf{x}_t)\Delta t$ since higher-order solvers would increase analytical complexity and computational cost.
As in BayesDiff~\citep{Kou23Bayesdiff}, we approximate the evolution of the element-wise variance from $\mathbf{x}_t$ to $\mathbf{x}_{t+\Delta t}$ as
\begin{equation}
	\label{eq:variance_propagation}
		\Var[\mathbf{x}_{t + \Delta t}] \approx
		\Var[\mathbf{x}_t] +  (\sigma^\theta_t(\bar{\mathbf{x}}_t)\Delta t)^2 \\
		+ 2\Delta t \Cov(\mathbf{x}_t, u^{\theta}_t(\mathbf{x}_t)),
\end{equation}
Here, $(\sigma_t^\theta(\bar{\mathbf{x}}_t))^2$ is the predicted velocity variance evaluated at the mean state, and $\Cov(\mathbf{x}_t, u^{\theta}_t(\mathbf{x}_t)) \in \mathbb{R}^n$ denotes the element-wise covariance between the state and its velocity.
We regard all covariances as element-wise covariances for both computational and spatial tractability in high-dimensional space.
The velocity variance quantifies injected velocity noise, while the state-velocity covariance captures how state uncertainty couples with the local sensitivity of the velocity field.

By inserting the linear term from \cref{eq:mean_velocity_linearization} into the element-wise $\Cov(\mathbf{x}_t, u^{\theta}_t(\mathbf{x}_t))$, it is approximated as
\begin{equation}
	\label{eq:cov_approx_taylor}
	\Cov(\mathbf{x}_t, u^{\theta}_t(\mathbf{x}_t)) \approx \diag(J^{\theta}_t(\bar{\mathbf{x}}_t)) \odot \Var[\mathbf{x}_t]
\end{equation}
where $J^{\theta}_t(\bar{\mathbf{x}}_t) = \pp{\bar{u}^{\theta}_t}{\mathbf{x}_t} \Big|_{\bar{\mathbf{x}}_t} \in \mathbb{R}^{n \times n}$ denotes the Jacobian of the mean velocity with respect to the state, evaluated at $\bar{\mathbf{x}}_t$.
Also, $\odot$ represents element-wise multiplication.
Since forming $\diag(J^{\theta}_t(\bar{\mathbf{x}}_t))$ explicitly is intractable in high dimensions, we approximate $\diag(J^{\theta}_t(\bar{\mathbf{x}}_t)) \odot \Var[\mathbf{x}_t]$ using Hutchinson's diagonal estimator~\citep{Bekas07Estimator, Dharangutte23Tight}.
Consequently, the covariance $\Cov(\mathbf{x}_t, u^{\theta}_t)$ can be estimated as:
\begin{equation}
	\label{eq:cov_approx_jvp}
	\Cov(\mathbf{x}_t, u^{\theta}_t(\mathbf{x}_t)) \approx \frac{1}{S}\sum_{i=1}^S(\boldsymbol{\sigma}^x_t \odot \mathbf{r}_i) \odot (J^{\theta}_t(\bar{\mathbf{x}}_t)(\boldsymbol{\sigma}^x_t \odot \mathbf{r}_i))
\end{equation}
with $\boldsymbol{\sigma}^x_t = \sqrt{\Var[\mathbf{x}_t]}$.
Each $\mathbf{r}_i\in\mathbb{R}^n$ is a Rademacher vector with independent entries drawn uniformly from $\{-1,+1\}$.
Here, $S$ is the number of Monte Carlo probes.

\begin{algorithm}[t]
		\caption{Estimating $\operatorname{diag}(J^{\theta}_t(\bar{\mathbf{x}}_t))\odot \Var[\mathbf{x}_t]$}
		\label{alg:jacobian_diag_sigma2}
		\begin{algorithmic}[1]
			\Statex \textbf{Input:}
			\Statex \begin{tabular}{@{}l@{~}l@{\hspace{0.9em}}l@{~}l@{}}
				$\bar{u}^\theta_t(\cdot)$: & velocity mean  & $\bar{\mathbf{x}}_t$: & state mean \\
				$\Var[\mathbf{x}_t]$:      & state variance & $S$:                  & number of probes
			\end{tabular}
			\State $\boldsymbol{\sigma}^x_t \gets \sqrt{\Var[\mathbf{x}_t]}$, $\mathbf{g} \gets \mathbf{0}$
			\For{$k = 1,\dots,S$}
			\State Sample $\mathbf{r}^{(k)} \in \{\pm 1\}^n$
			\State $\mathbf{v}^{(k)} \gets \boldsymbol{\sigma}^x_t \odot \mathbf{r}^{(k)}$
			\State $\mathbf{Jv}^{(k)} \gets \mathrm{JVP}(\bar{u}^\theta_t, \bar{\mathbf{x}}_t, \mathbf{v}^{(k)})$
			\State $\mathbf{g} \gets \mathbf{g} + \mathbf{v}^{(k)} \odot \mathbf{Jv}^{(k)}$
			\EndFor
			\State \Return $\mathbf{g}/S$
		\end{algorithmic}
\end{algorithm}

Using Jacobian-vector products (JVPs)~\citep{Baydin18Automatic}, $J^\theta_t(\bar{\mathbf{x}}_t)(\boldsymbol{\sigma}^x_t \odot \mathbf{r}_i)$ can be computed efficiently via automatic differentiation without forming $J^\theta_t$ explicitly.
Algorithm~\ref{alg:jacobian_diag_sigma2} summarizes the Monte Carlo estimator corresponding to \cref{eq:cov_approx_jvp}.

We provide full derivations of \cref{eq:mean_ode,eq:variance_propagation,eq:cov_approx_taylor,eq:cov_approx_jvp}, as well as alternative covariance approximations, in Appendix~\ref{app:variance_propagation}.
For completeness, we also derive the variance propagation rule under Heun's 2nd-order (Heun2) discretization in \cref{app:heun_variance_propagation}.
\Cref{app:variance_propagation_method} compares the two methods empirically.

\subsection{Uncertainty-Aware Guidance for Flow Matching}
\label{subsec:uncertainty_aware_guidance}

We incorporate the predicted velocity uncertainty into guided sampling by modifying the mean dynamics in \cref{eq:mean_ode}.
We present two mechanisms: (i) an uncertainty-based pseudo-likelihood whose gradient is used as a guidance term, and (ii) an adaptive choice of the CFG scale that reduces the predicted variance of the extrapolated velocity.

\textbf{Uncertainty-aware guidance (UG).}
Standard guidance steers generation toward a condition $y$ by adding $b_t w \nabla_{\mathbf{x}_t}\log p_t(y\mid \mathbf{x}_t)$ to the velocity field $u^\theta_t(\mathbf{x}_t)$, where $w\ge 0$ is the guidance scale and $b_t = -\frac{\dot{\beta}_t\beta_t\alpha_t - \dot{\alpha}_t\beta_t^2}{\alpha_t}$ is the path-dependent coefficient for the affine path $\mathbf{x}_t = \alpha_t \mathbf{x}_1 + \beta_t \mathbf{x}_0$~\cite{Dhariwal21Diffusion, Lipman24Flowcode}.

To bias sampling toward low-uncertainty regions, we define a pseudo-likelihood over the predicted variance given the state,
\begin{equation}
	\label{eq:ucg_pseudo_likelihood}
	\tilde{p}_t \big( (\sigma_t^\theta)^2 \mid  \mathbf{x}_t \big) \propto \exp\!\big(f((\sigma^{\theta}_t(\mathbf{x}_t))^2)\big),
\end{equation}
where the scalar function $f$ has larger values when the predicted velocity variance is smaller.
Thus, $\tilde{p}_t$ favors states with low predicted uncertainty.
We substitute $\nabla_{\mathbf{x}_t}\log \tilde{p}_t \big( (\sigma_t^\theta)^2 \mid  \mathbf{x}_t \big) = \nabla_{\mathbf{x}_t}f((\sigma_t^\theta(\mathbf{x}_t))^2)$ into the classifier-guidance template to obtain the UG mean velocity:
\begin{equation}
	\label{eq:uncertainty_classifier_guidance}
	\bar{u}_{t,\mathrm{UG}}^{\theta}(\bar{\mathbf{x}}_t) = \bar{u}_t^{\theta}(\bar{\mathbf{x}}_t) + b_t w \nabla_{\bar{\mathbf{x}}_t}f((\sigma^{\theta}_t(\bar{\mathbf{x}}_t))^2),
\end{equation}
which steers the trajectory toward low-uncertainty regions.
In our experiments, we use the negative squared mean of the element-wise predicted variances, $f(\sigma^2) = -\big(\tfrac{1}{n}\sum_{i=1}^n \sigma_i^2\big)^2$.
\cref{app:f_sigma} shows that UG is robust to the choice of $f$.

\textbf{Uncertainty-aware classifier-free guidance (UCFG).}
When UA-Flow is trained with classifier-free conditioning, the CFG extrapolated mean velocity is:
\begin{equation}
	\label{eq:cfg_mean_velocity}
	\bar{u}_{t,\mathrm{CFG}}^{\theta}(\bar{\mathbf{x}}_t \mid y)
	= (1 + \lambda)\bar{u}_t^{\theta}(\bar{\mathbf{x}}_t \mid y)
	- \lambda\bar{u}_t^{\theta}(\bar{\mathbf{x}}_t \mid \varnothing),
\end{equation}
where $\lambda \ge 0$ is the CFG scale and $\varnothing$ denotes the null condition.
In the CFG discussion, \emph{conditional} and \emph{unconditional} refer to predictions with class condition $y$ and null condition $\varnothing$, respectively.
The conditional target velocity in \cref{subsec:learning_mean_variance} is conditioned on a data endpoint $\mathbf{x}_1$.

Let $\sigma_{t,y}^{\theta}(\bar{\mathbf{x}}_t)$ and $\sigma_{t,\varnothing}^{\theta}(\bar{\mathbf{x}}_t)$ denote the predicted element-wise standard deviations of the conditional and unconditional velocities, respectively.
For coordinate $i$, the variance of the CFG velocity is
\begin{equation}
	\label{eq:cfg_var_general}
	\Var\big[u_{t,\mathrm{CFG}}^\theta(\mathbf{x}_t \mid y)\big]_i = (1+\lambda)^2 (\sigma_{t,y,i}^\theta(\bar{\mathbf{x}}_t))^2+
	\lambda^2(\sigma_{t,\varnothing,i}^\theta(\bar{\mathbf{x}}_t))^2
	-2\lambda(1+\lambda)r_i\sigma_{t,y,i}^\theta(\bar{\mathbf{x}}_t)\sigma_{t,\varnothing,i}^\theta(\bar{\mathbf{x}}_t),
\end{equation}
where $r_i$ is the correlation coefficient between the conditional and unconditional velocity components.
The two velocity predictions come from the same network at the same state and differ only in conditioning, motivating strong correlation approximation $r_i\approx1$.
Under this approximation, the variance of the CFG velocity is
\begin{equation}
	\label{eq:cfg_var_approx}
	\Var\big[u_{t,\mathrm{CFG}}^\theta(\mathbf{x}_t \mid y)\big]
	\approx
	\big((1 + \lambda)\sigma_{t,y}^{\theta}(\bar{\mathbf{x}}_t) - \lambda\sigma_{t,\varnothing}^{\theta}(\bar{\mathbf{x}}_t)\big)^2.
\end{equation}
Empirically, the predicted conditional and unconditional mean velocities have Pearson correlation $r=0.993$ on ImageNet-256.
We compute this correlation from element-wise values pooled across all sampling steps.
The correlation at every individual step exceeds $0.98$ (\cref{app:cfg_velocity_correlation}).

For each sample, we choose one scalar $\lambda$ to minimize the total predicted variance of the extrapolated velocity.
We cap this scale at $\lambda_{\max}$ to prevent the extrapolated velocity from diverging:
\begin{equation}
	\label{eq:lambda_clamp}
	\lambda^* = \min(\lambda_{\mathrm{opt}}, \lambda_{\max}),
\end{equation}
where
\begin{equation} \label{eq:lambda_opt_argmin}
	\lambda_{\mathrm{opt}} = \argmin_{\lambda \ge 0}\; \sum_{i=1}^n \big((1 + \lambda)\sigma_{t,y,i}^{\theta}(\bar{\mathbf{x}}_t) - \lambda\sigma_{t,\varnothing,i}^{\theta}(\bar{\mathbf{x}}_t)\big)^2. 
\end{equation}
Here, the sum is over the coordinates of one sample.
The closed-form solution for $\lambda_{\mathrm{opt}}$ is given in \cref{app:cfg_lambda}.
UCFG recomputes $\lambda_{\mathrm{opt}}$ and $\lambda^*$ for each sample at every sampling step, capped by $\lambda_{\max}$.

UCFG and UG can be applied sequentially at each sampling step.
UCFG first returns the guided mean and variance via \cref{eq:cfg_mean_velocity,eq:cfg_var_approx,eq:lambda_clamp}.
UG then computes its gradient from this local guided variance and adds the correction in \cref{eq:uncertainty_classifier_guidance}.

\section{Experiments}
\label{sec:experiments}

We design experiments to evaluate whether the uncertainty estimated by UA-Flow
\emph{(i) can be used as a sample-level reliability signal for generated samples},
\emph{(ii) represents how weakly each generated sample is supported by the dataset manifold},
and \emph{(iii) can be used as a control signal to improve generation via guided sampling}.
To assess (i), we filter out high-uncertainty generated images and compare the resulting FID and precision/recall against baseline uncertainty estimation methods (\cref{subsec:filtering_exp}).
To assess (ii), we measure the correlation of the estimated uncertainty with two manifold-based quantities, distance to the data manifold and rarity, and again compare against the baselines (\cref{subsec:corr_exp}).
To assess (iii), we conduct controlled ablations on uncertainty-aware guidance (UG) (\cref{subsec:ucg_exp}) and uncertainty-aware classifier-free guidance (UCFG) (\cref{subsec:ucfg_exp}) under matched sampling settings.

\subsection{Experimental Setup}
\label{subsec:exp_setup}

We evaluate our method on CIFAR-10~\citep{Krizhevsky10CIFAR10} and ImageNet~\citep{Deng09ImageNet} at resolutions $128\times128$ (\emph{ImageNet-128}) and $256\times256$ (\emph{ImageNet-256}).
We report the generative quality metrics, Fr\'echet Inception Distance (FID)~\citep{Heusel17FID} and precision/recall~\citep{Kynkaanniemi19PR}, together with the sample-level manifold metrics, distance to the data manifold and rarity~\citep{Han23Rarity}.
FID measures the overall distributional similarity between generated and real images, while precision and recall respectively capture sample fidelity and distributional coverage.
Distance to the data manifold is the feature-space distance from a generated sample to its $k$-nearest real neighbor (here, $k=3$), and rarity quantifies how atypical a generated sample is with respect to the real data.
Details not described in this subsection are provided in \cref{app:implementation_details}.

\paragraph{Training and Model Architectures.}
For CIFAR-10, we use an ADM-based~\citep{Dhariwal21Diffusion} unconditional flow matching model in pixel space.
For ImageNet, we adopt a DiT-based conditional latent flow matching model~\citep{Dao23Latent} trained on images resized to $128\times128$ and $256\times256$.
Specifically, we use DiT-B/2~\citep{Peebles23DiT} as the backbone and the pretrained autoencoder from Stable Diffusion~\citep{Rombach22StableDiffusion} to map RGB images to a latent tensor with downsampling ratio 8 and 4 channels.

For CIFAR-10 and ImageNet-128, we first train a vanilla flow matching model and then fine-tune it by adding a variance prediction head.
For ImageNet-256, we fine-tune a pretrained latent flow matching model released in prior work~\citep{Dao23Latent}.

\paragraph{Sampling Process.}
All image-generation experiments use the second-order Heun's method solver with 50 sampling steps.
UA-Flow uses Euler updates for uncertainty propagation, as described in \cref{subsec:uncertainty_propagation}.
For CIFAR-10, we adopt the EDM sampling schedule~\citep{Karras22EDM}, while for ImageNet we use uniformly spaced time steps.
UG is applied every two sampling steps, whereas CFG and UCFG are applied at every step.

\paragraph{Baselines.}
We compare against BayesDiff~\citep{Kou23Bayesdiff}, Aleatoric Uncertainty (AU)~\citep{DeVita25Aleatoric}, and Generative Uncertainty (GenUnc)~\citep{Jazbec25Genunc}.
As these methods were originally proposed for diffusion models, we adapt them to flow matching by defining uncertainty over the velocity field and propagating it through the flow dynamics.
Implementation details are provided in \cref{app:subsec:flow_baselines}.

\paragraph{UA-Flow + CLIP.}
We additionally introduce a variant that scores uncertainty in the CLIP embedding space.
Specifically, we draw multiple samples $\mathbf{x}_{1,i}$ from the predicted distribution $\mathcal{N}(\bar{\mathbf{x}}_1, \mathrm{Var}[\mathbf{x}_1])$ and compute the variance of their CLIP embeddings as the sample-level uncertainty score, analogous to GenUnc.

\subsection{Filtering Images with High-Uncertainty}
\label{subsec:filtering_exp}

\begin{figure}[t]
	\centering
	\includegraphics[width=0.85\textwidth]{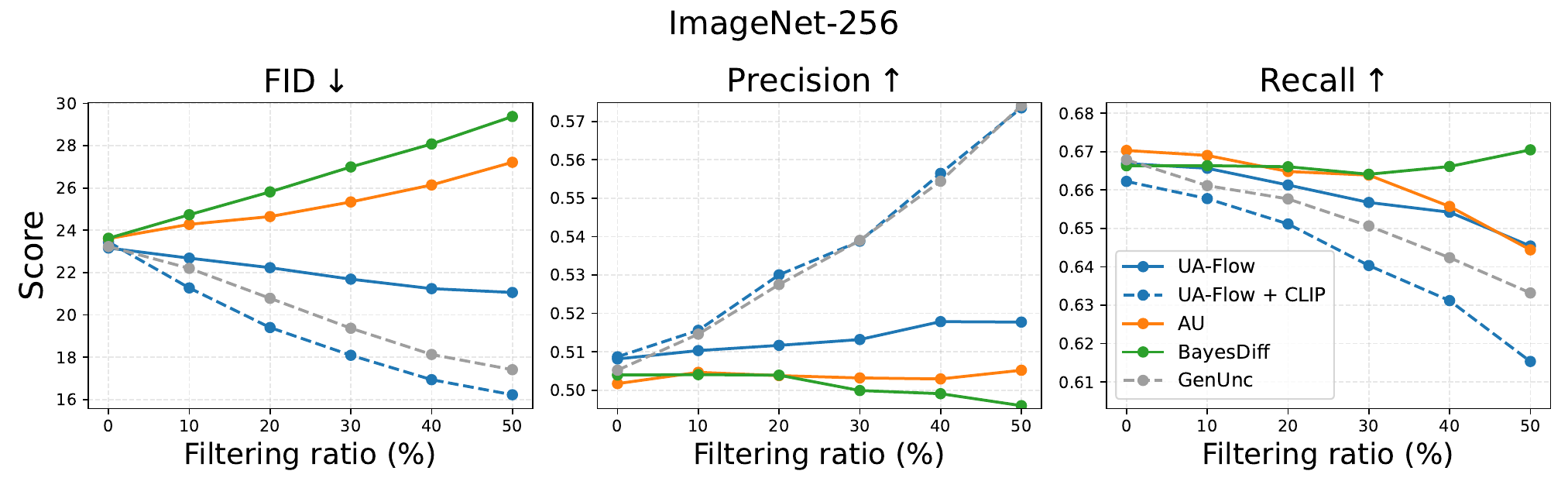}
	\caption{
		\textbf{Generative quality metrics vs.\ filtering ratio on ImageNet-256.}
		At each filtering level, high-uncertainty samples are removed and 50k samples are drawn from the remainder for evaluation.
		Compared to the element-wise baselines AU~\citep{DeVita25Aleatoric} and BayesDiff~\citep{Kou23Bayesdiff}, UA-Flow achieves lower FID and higher precision after filtering.
		UA-Flow + CLIP further matches the CLIP-based scalar reference GenUnc~\citep{Jazbec25Genunc}.
	}
	\label{fig:filtering_uncertainty_imagenet256}
\end{figure}

We evaluate whether the predicted uncertainty provides a sample-level reliability signal by progressively filtering out high-uncertainty generated samples and tracking changes in FID and precision/recall.

\begin{figure}[t]
	\centering
	\begin{subfigure}[t]{0.32\textwidth}
		\centering
		\includegraphics[width=\textwidth]{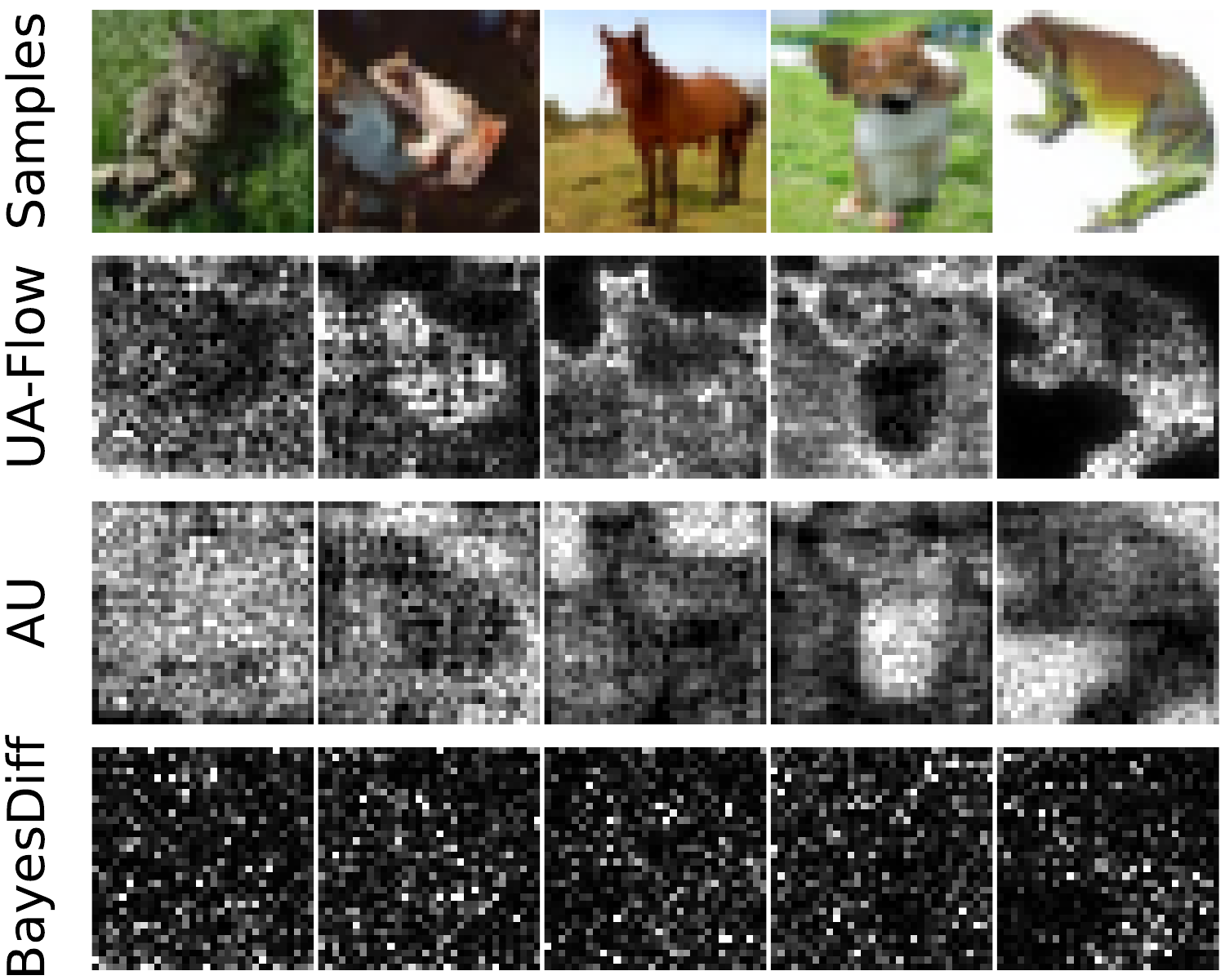}
		\caption{CIFAR-10 (pixel space).}
	\end{subfigure}
	\hfill
	\begin{subfigure}[t]{0.32\textwidth}
		\centering
		\includegraphics[width=\textwidth]{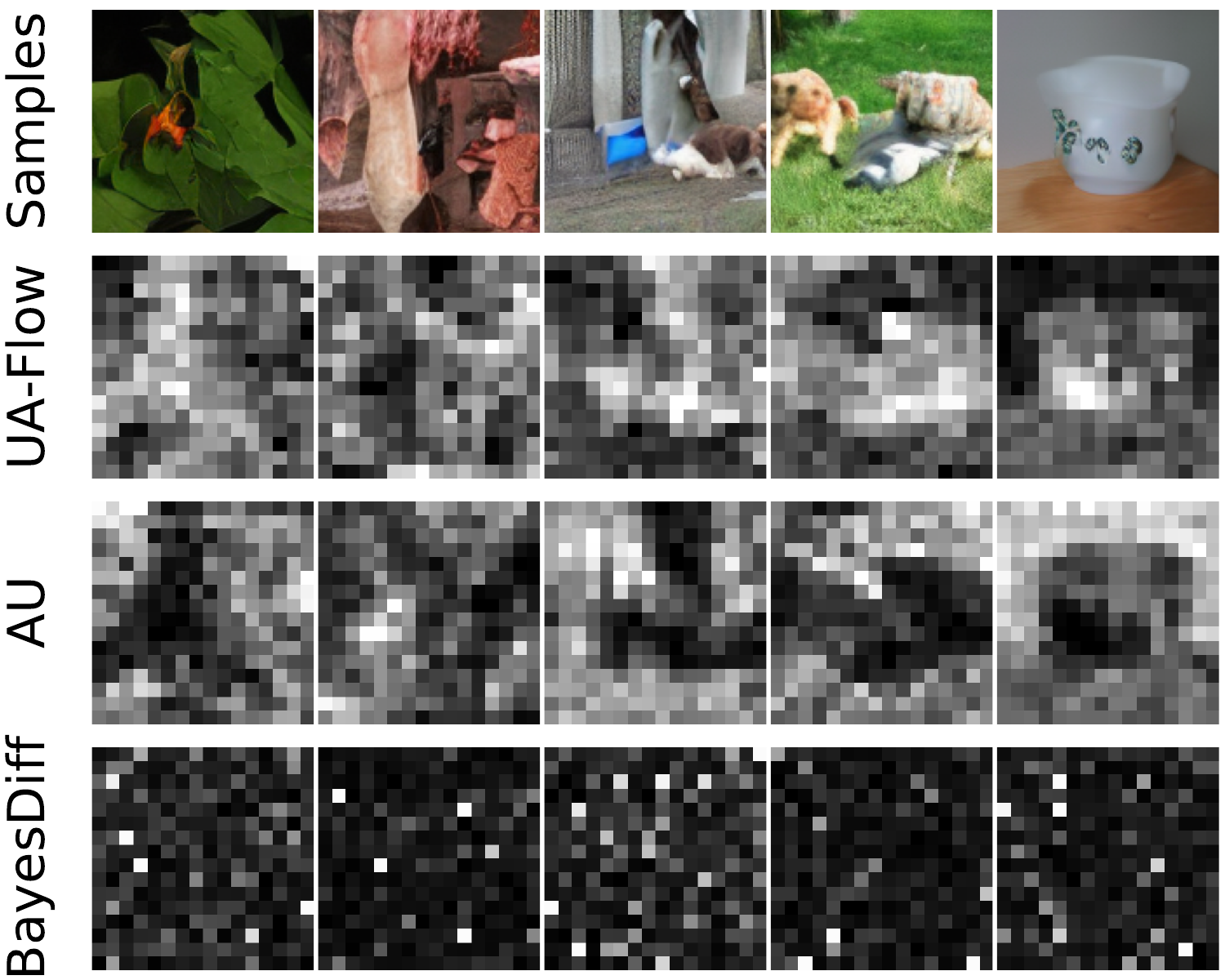}
		\caption{ImageNet-128 (latent space).}
	\end{subfigure}
	\hfill
	\begin{subfigure}[t]{0.32\textwidth}
		\centering
		\includegraphics[width=\textwidth]{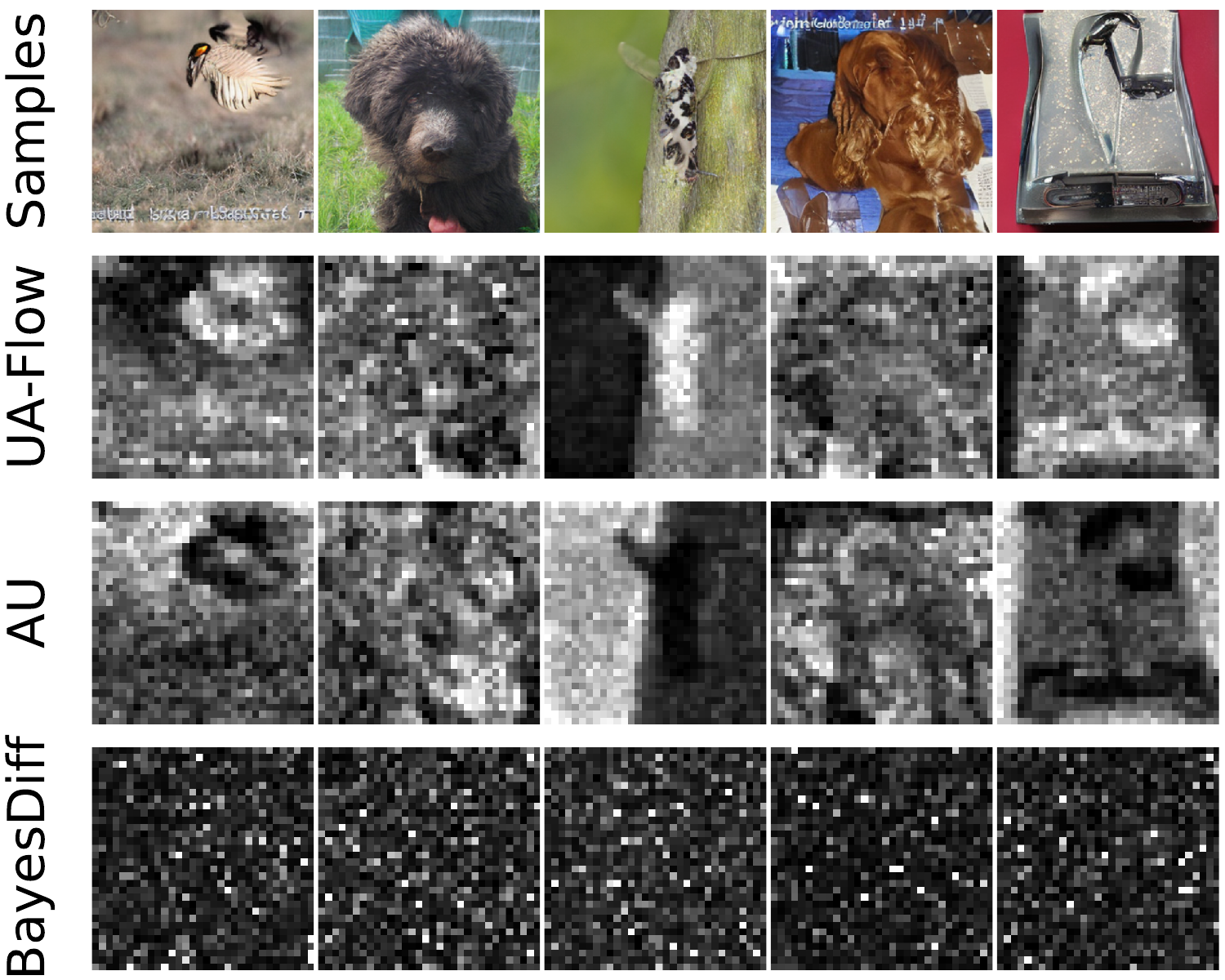}
		\caption{ImageNet-256 (latent space).}
	\end{subfigure}
	\caption{
		\textbf{Uncertainty maps from UA-Flow, AU, and BayesDiff on generated samples.}
		Brighter values indicate higher uncertainty; maps are normalized per image.
		UA-Flow localizes high-uncertainty regions, whereas AU often produces inverted patterns and BayesDiff produces noisy maps.
	}
	\label{fig:filtering_uncertainty_grids}
\end{figure}

\begin{table}[t]
	\centering
	\renewcommand{\arraystretch}{0.95}
	\caption{
		\textbf{Computational cost (TFLOPS) for sampling and uncertainty quantification per image.}
		\textit{Vanilla} reports sampling cost only.
	}
	\label{tab:flops_comparison}
	\begin{tabular}{lccc}
		\toprule
		Method & CIFAR-10 & ImageNet-128 & ImageNet-256 \\
		\midrule
		Vanilla   & 4.044 & 1.112 & 4.601 \\
		\midrule
		AU        & \textbf{5.014} & \textbf{1.379} & \textbf{5.706} \\
		BayesDiff & 9.201 & 2.479 & 10.26 \\
		GenUnc    & 24.32 & 7.655 & 31.39 \\
		\midrule
		\textbf{UA-Flow}        & 5.506 & 1.380 & 5.761 \\
		\textbf{UA-Flow + CLIP} & 5.550 & 2.200 & 8.916 \\
		\bottomrule
	\end{tabular}
\end{table}
\paragraph{Uncertainty aggregation for filtering.}
Filtering requires a single scalar uncertainty score per sample.
GenUnc and UA-Flow + CLIP directly output a scalar in the CLIP~\citep{Radford21CLIP} embedding space, whereas UA-Flow, BayesDiff, and AU produce element-wise uncertainty maps.
For these element-wise methods, we score each sample by the mean of the top $10\%$ highest-uncertainty elements of its uncertainty map, suppressing the influence of large low-uncertainty background regions.
This top-$10\%$ mean is the empirical Conditional Value-at-Risk (CVaR) at level $\alpha=0.1$, $\mathrm{CVaR}_{0.1}(u)=\E[u \mid u \ge Q_{0.9}(u)]$, a standard tail-risk objective~\citep{Rockafellar00CVaR}.
For UA-Flow and BayesDiff, the uncertainty map is the propagated endpoint variance $\mathrm{Var}[\mathbf{x}_1]$; AU aggregates velocity variance over late-stage sampling steps (\cref{app:subsec:flow_baselines}).
Following BayesDiff's protocol, UA-Flow and BayesDiff update uncertainty every four sampling steps, and we use a single Monte Carlo probe ($S=1$) for the covariance estimator in \cref{eq:cov_approx_jvp}. \cref{app:probe_s} shows that more probes do not noticeably improve filtering.

\paragraph{Filtering procedure.}
Without any guidance, we generate $100\mathrm{k}$ images per dataset (class-conditional uncertainty for ImageNet, unconditional for CIFAR-10), rank them by the sample-level uncertainty score, and progressively remove the top $10\%$ up to $50\%$.
At each ratio, metrics are computed on $50\mathrm{k}$ randomly drawn images against $50\mathrm{k}$ references.

\paragraph{Results.}
UA-Flow filtering induces a precision-recall trade-off (\cref{fig:filtering_uncertainty_imagenet256,fig:filtering_uncertainty_cifar10,fig:filtering_uncertainty_imagenet128}): increasing the filter ratio raises precision and lowers recall.
On ImageNet, FID improves after filtering; on CIFAR-10, where the unfiltered FID is already low, FID increases with the filter ratio.

Compared to element-wise baselines (AU, BayesDiff), UA-Flow achieves lower FID and higher precision after filtering on ImageNet-128/256.
With AU, recall consistently decreases, while precision either decreases or does not increase significantly.
BayesDiff shows a mild precision-recall trade-off on CIFAR-10 that does not carry over to ImageNet.
Together, these filtering results suggest that neither AU nor BayesDiff provides a consistent signal of sample fidelity across the tested datasets under flow matching.
Qualitatively, \cref{fig:filtering_uncertainty_grids} shows that UA-Flow's uncertainty maps localize high-uncertainty regions whereas AU produces largely inverted patterns and BayesDiff produces noisy maps that fail to localize.
Additional latent-space comparisons and an analysis of BayesDiff's noisy maps are provided in \cref{app:additional_results:filtering,app:additional_results:uncertainty_evolution}.

AU has the lowest computational cost, but UA-Flow exceeds it by only 10\% on CIFAR-10 and by less than 1\% on ImageNet-128/256, while remaining 1.7--1.8$\times$ cheaper than BayesDiff (\cref{tab:flops_comparison}).

GenUnc attains a lower FID than UA-Flow under its default aggregation, but aggregating UA-Flow's uncertainty in the same CLIP embedding space (UA-Flow + CLIP) recovers a comparable precision-recall trade-off, outperforming GenUnc in FID on ImageNet-128/256 and trailing in FID on CIFAR-10, at 3.5--4.4$\times$ less compute (\cref{tab:flops_comparison}).

\subsection{Correlation of Uncertainty with Manifold Support}
\label{subsec:corr_exp}

Filtering shows that uncertainty separates reliable from unreliable samples, but not why it does so.
We therefore test whether higher predicted uncertainty is associated with weaker support from the data manifold.
We quantify manifold support by the distance to the data manifold and by the rarity score, and measure how strongly the predicted uncertainty correlates with each.
When computing the rarity score, we exclude generated samples and patches classified as existing out-of-data manifold.

Without any guidance, we generate $5{,}000$ images per dataset and evaluate them against $5{,}000$ reference images.
As in \cref{subsec:filtering_exp}, uncertainty is updated every four sampling steps.
We use a single Monte Carlo probe ($S=1$) for the covariance estimator in \cref{eq:cov_approx_jvp}.
We report the Spearman rank correlation $\rho$ between the uncertainty score and each manifold metric, measured at two different levels: sample-wise and patch-wise.
Here, \emph{sample-wise} refers to one score per image, and \emph{patch-wise} refers to one score per image patch.

\paragraph{Sample-wise correlation.}
Both metrics are computed for each generated image as a whole and correlated with a single scalar uncertainty per image.
We use the CLIP-space scores, UA-Flow + CLIP and GenUnc, for this comparison, since \cref{subsec:filtering_exp} shows that the variance of CLIP embeddings tracks sample-level reliability more closely than uncertainty aggregated in pixel or latent space.

\paragraph{Patch-wise correlation.}
A sample-wise score collapses an image to a single scalar, whereas UA-Flow resolves uncertainty spatially across the generated sample.
To test whether this spatial structure correlates with local manifold support, we compute the same two metrics locally.
We divide generated and reference samples into non-overlapping patches.
We use $2\times2$ latent-space patches on ImageNet-128/256 and $4\times4$ pixel-space patches on CIFAR-10.
The uncertainty score of a patch is the sum of its element-wise uncertainty map over the patch elements, so UA-Flow, AU, and BayesDiff are all compared at this granularity.

\begin{table}[t]
	\centering
	\renewcommand{\arraystretch}{0.9}
	\caption{
		\textbf{Spearman rank correlation $\rho$ between predicted uncertainty and manifold support.}
		Sample-wise methods produce a single CLIP-space scalar per image, whereas patch-wise methods produce element-wise uncertainty maps and are correlated at the patch level.
		Sample-wise, UA-Flow + CLIP is strongest on ImageNet-128 and ImageNet-256, while GenUnc leads on CIFAR-10.
		Patch-wise, UA-Flow attains the strongest correlation with both the distance to the data manifold (Dist. to manifold) and the rarity score on every dataset.
	}
	\label{tab:corr_manifold}
	\begin{tabular}{llccccc}
		\toprule
		& & \multicolumn{2}{c}{Sample-wise} & \multicolumn{3}{c}{Patch-wise} \\
		\cmidrule(lr){3-4} \cmidrule(lr){5-7}
		Dataset & Metric & GenUnc & \textbf{UA-Flow + CLIP} & AU & BayesDiff & \textbf{UA-Flow} \\
		\midrule
		\multirow{2}{*}{CIFAR-10}
		& Dist.\ to manifold & $\mathbf{0.3462}$ & $0.2421$ & $-0.5654$ & $0.5064$ & $\mathbf{0.6114}$ \\
		& Rarity             & $\mathbf{0.3184}$ & $0.2183$ & $-0.5603$ & $0.4803$ & $\mathbf{0.6066}$ \\
		\midrule
		\multirow{2}{*}{ImageNet-128}
		& Dist.\ to manifold & $0.3579$ & $\mathbf{0.3672}$ & $-0.3187$ & $0.1340$ & $\mathbf{0.4609}$ \\
		& Rarity             & $0.3501$ & $\mathbf{0.3901}$ & $-0.2925$ & $0.1078$ & $\mathbf{0.4037}$ \\
		\midrule
		\multirow{2}{*}{ImageNet-256}
		& Dist.\ to manifold & $0.4190$ & $\mathbf{0.4426}$ & $-0.4719$ & $0.1820$ & $\mathbf{0.5404}$ \\
		& Rarity             & $0.3995$ & $\mathbf{0.4165}$ & $-0.4154$ & $0.1422$ & $\mathbf{0.4771}$ \\
		\bottomrule
	\end{tabular}
\end{table}
\paragraph{Results.}
In the sample-wise setting, UA-Flow + CLIP correlates more strongly with both the distance to the data manifold and the rarity score on ImageNet-128/256, whereas GenUnc leads on CIFAR-10.
To identify the source of this gap, we discard the variance propagation of \cref{subsec:uncertainty_propagation} on CIFAR-10 and instead draw a velocity from the learned mean and variance at every sampling step, taking the variance of the resulting samples directly as GenUnc does.
The correlations then rise to $0.3524$ and $0.3282$ for the distance to the data manifold and the rarity score, above the corresponding GenUnc values of $0.3462$ and $0.3184$.
This comparison suggests that UA-Flow's learned velocity uncertainty carries information about manifold support that is partly lost through the variance propagation approximations.
Under direct sampling, UA-Flow's higher endpoint correlations also suggest an advantage over the velocity uncertainty obtained by the last-layer Laplace approximation (LLLA)~\citep{Daxberger21Laplace} used in GenUnc.

Apart from this case, UA-Flow-based methods show higher correlations than all baselines.
The positive correlations indicate that higher uncertainty is associated with greater distance from the data manifold and higher rarity.
This supports our hypothesis that the uncertainty of UA-Flow carries information about how weakly each generated sample is supported by the data manifold.
Moreover, the higher patch-wise correlations show that UA-Flow captures this information spatially.

\subsection{Uncertainty-Aware Guidance}
\label{subsec:ucg_exp}

We evaluate UG by sweeping its scale $w \in [0, 50]$ under fixed CFG scales $\lambda \in \{0, 0.25, 0.5\}$.
Note that UG and CFG are disabled when $w=0$ and $\lambda=0$, respectively.
``CFG + UG'' first forms the standard CFG mean and its fixed-$\lambda$ variance using \cref{eq:cfg_mean_velocity,eq:cfg_var_approx}, then applies the uncertainty-aware correction in \cref{eq:uncertainty_classifier_guidance}; it does not use adaptive UCFG.
Thus, \cref{tab:cg_all} fixes $\lambda\in\{0,0.25,0.5\}$ and sweeps $w$ to isolate UG under matched CFG strength.

\begin{figure}[t]
	\centering
	\includegraphics[width=0.9\textwidth]{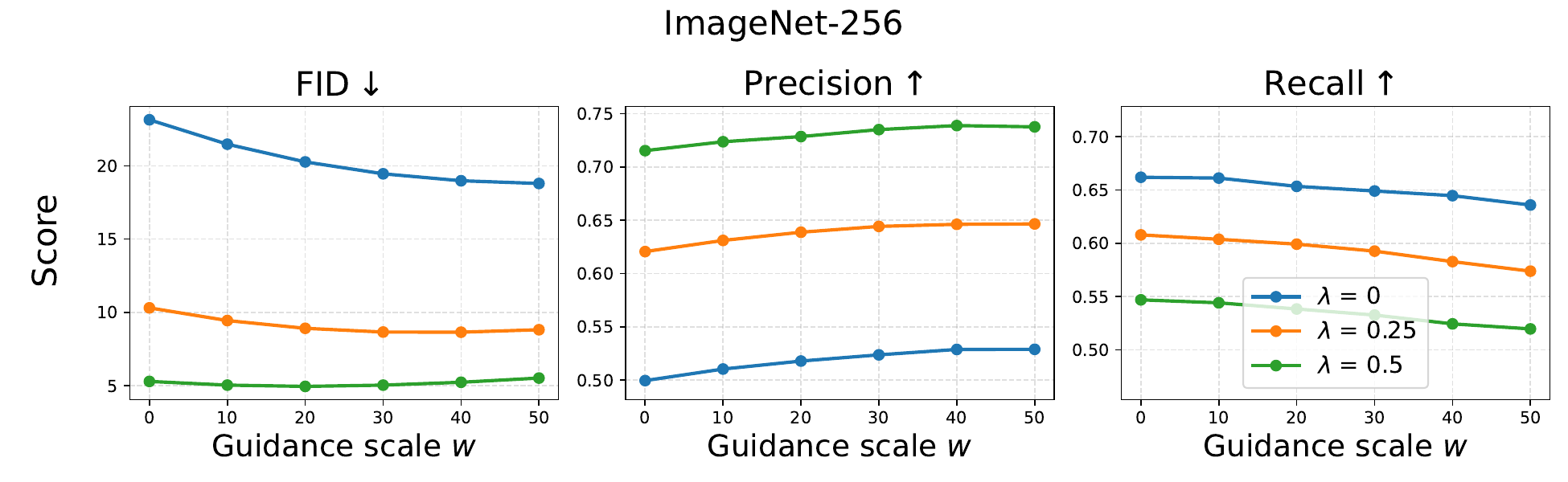}
	\caption{
		\textbf{FID, precision, and recall as the UG scale $w$ varies on ImageNet-256 under fixed CFG scales $\lambda\in\{0,0.25,0.5\}$.}
		Increasing $w$ generally improves precision while reducing recall.
		At nonzero CFG scales, FID reaches a minimum at an intermediate UG strength and then increases.
	}
	\label{fig:guidance_metrics_imagenet256}
\end{figure}

\begin{figure}[t]
	\centering
	\begin{subfigure}[t]{0.48\textwidth}
		\centering
		\begin{overpic}[width=\linewidth]{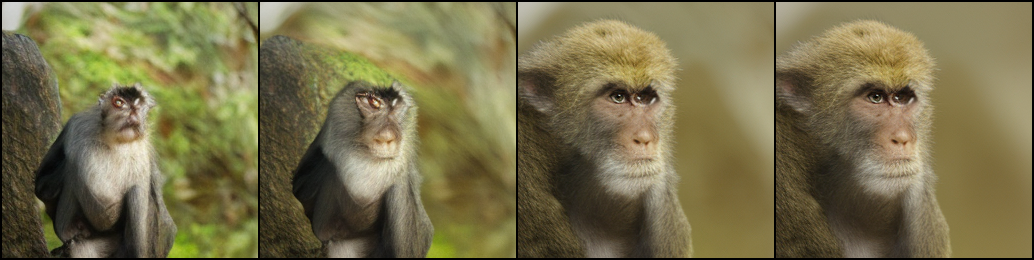}
		\end{overpic}
	\end{subfigure}
	\hfill
	\begin{subfigure}[t]{0.48\textwidth}
		\centering
		\begin{overpic}[width=\linewidth]{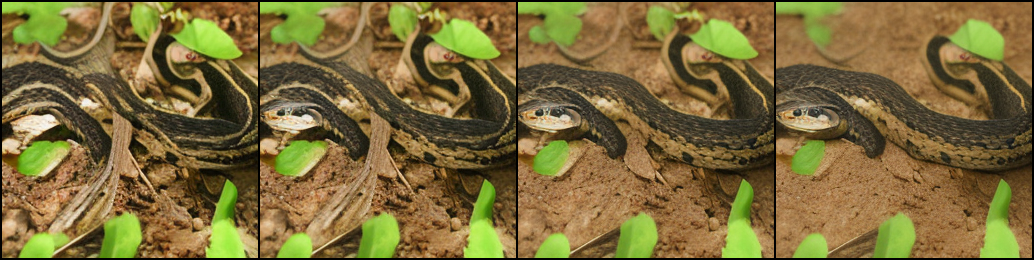}
		\end{overpic}
	\end{subfigure}

	\medskip
	\begin{subfigure}[t]{0.48\textwidth}
		\centering
		\includegraphics[width=\linewidth]{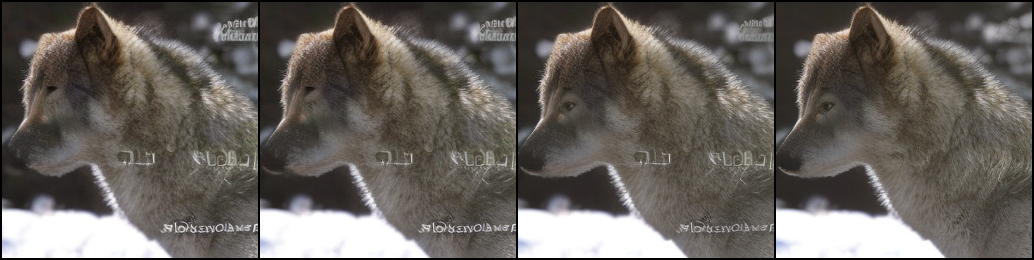}
	\end{subfigure}
	\hfill
	\begin{subfigure}[t]{0.48\textwidth}
		\centering
		\includegraphics[width=\linewidth]{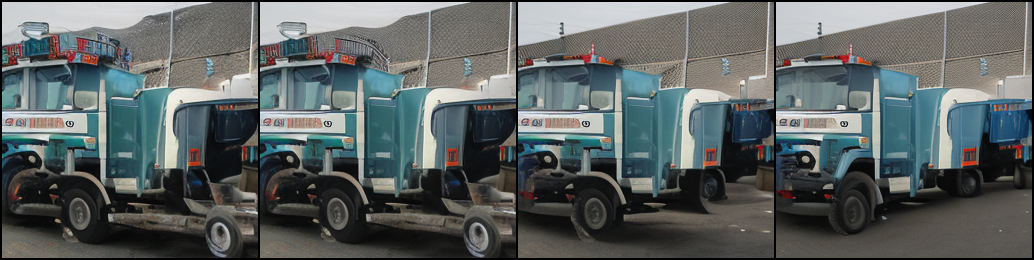}
	\end{subfigure}
	\caption{
		\textbf{ImageNet-256 samples under different UG scales $w$ at a fixed CFG scale $\lambda=0.5$.}
		Each panel uses a fixed class label and random seed; columns sweep $w\in\{0,10,30,50\}$ from left to right. UG trades diversity for fidelity as the guidance strength increases.
	}
	\label{fig:imagenet256_guidance_scales}
\end{figure}

\begin{table}[t]
	\centering
	\caption{
		\textbf{FID, precision and recall under uncertainty-aware guidance (UG) across datasets.}
		For each CFG scale $\lambda$, we report the best-performing UG setting (lowest FID).
		AU denotes uncertainty-aware guidance using aleatoric uncertainty~\citep{DeVita25Aleatoric} and is included as a reference baseline.
		UG consistently improves precision compared to vanilla sampling under matched CFG settings, and it also yields better FID on ImageNet-128 and ImageNet-256.
	}
	\label{tab:cg_all}
	
	\begin{subtable}[b]{0.26\columnwidth}
		\centering
		\caption{CIFAR-10}
		\resizebox{\textwidth}{!}{%
			\begin{tabular}{lcccc}
				\toprule
				Setting & $w$ & FID$\downarrow$ & Prec.$\uparrow$ & Rec.$\uparrow$ \\
				\midrule
				\textcolor{gray}{AU}  & \textcolor{gray}{--}  & \textcolor{gray}{2.18} & \textcolor{gray}{0.6549} & \textcolor{gray}{0.6328} \\
				\midrule
				Vanilla   & --  & \textbf{2.13} & 0.6570 & \textbf{0.6289} \\
				UG only & 10  & 2.43 & \textbf{0.6585} & 0.6245 \\
				\bottomrule
		\end{tabular}
		}
	\end{subtable}
	\hfill
	\begin{subtable}[b]{0.36\columnwidth}
		\centering
		\caption{ImageNet-128}
		\resizebox{\textwidth}{!}{%
			\begin{tabular}{lccccc}
				\toprule
				Setting & $\lambda$ & $w$ & FID$\downarrow$ & Prec.$\uparrow$ & Rec.$\uparrow$ \\
				\midrule
				\textcolor{gray}{AU}  & \textcolor{gray}{--}   & \textcolor{gray}{--}  & \textcolor{gray}{27.21} & \textcolor{gray}{0.4500} & \textcolor{gray}{0.6618} \\
				\midrule
				Vanilla    & \multirow{2}{*}{0.0}  & 0  & 27.23 & 0.4525 & \textbf{0.6697} \\
				UG only  &                       & 30 & \textbf{19.00} & \textbf{0.4925} & 0.6412 \\
				\midrule
				CFG only   & \multirow{2}{*}{0.25} & 0  & 14.76 & 0.5442 & \textbf{0.6297} \\
				CFG + UG &                       & 20 & \textbf{10.71} & \textbf{0.5798} & 0.6120 \\
				\midrule
				CFG only   & \multirow{2}{*}{0.5}  & 0  & 8.29 & 0.6251 & \textbf{0.5858} \\
				CFG + UG &                       & 20 & \textbf{6.95} & \textbf{0.6452} & 0.5633 \\
				\bottomrule
		\end{tabular}
		}
	\end{subtable}
	\hfill
	\begin{subtable}[b]{0.36\columnwidth}
		\centering
		\caption{ImageNet-256}
		\resizebox{\textwidth}{!}{%
			\begin{tabular}{lccccc}
				\toprule
				Setting & $\lambda$ & $w$ & FID$\downarrow$ & Prec.$\uparrow$ & Rec.$\uparrow$ \\
				\midrule
				\textcolor{gray}{AU}  & \textcolor{gray}{--}   & \textcolor{gray}{--}  & \textcolor{gray}{23.01} & \textcolor{gray}{0.4983} & \textcolor{gray}{0.6722} \\
				\midrule
				Vanilla    & \multirow{2}{*}{0.0}  & 0  & 23.14 & 0.4997 & \textbf{0.6619} \\
				UG only  &                       & 50 & \textbf{18.79} & \textbf{0.5290} & 0.6358 \\
				\midrule
				CFG only   & \multirow{2}{*}{0.25} & 0  & 10.31 & 0.6207 & \textbf{0.6078} \\
				CFG + UG &                       & 40 & \textbf{8.65} & \textbf{0.6463} & 0.5828 \\
				\midrule
				CFG only   & \multirow{2}{*}{0.5}  & 0  & 5.34 & 0.7132 & \textbf{0.5495} \\
				CFG + UG &                       & 20 & \textbf{5.00} & \textbf{0.7281} & 0.5393 \\
				\bottomrule
		\end{tabular}
		}
	\end{subtable}
\end{table}

\paragraph{Results.}
Increasing $w$ induces a precision-recall trade-off: precision generally increases while recall decreases, as shown for ImageNet-256 in \cref{fig:guidance_metrics_imagenet256}.
At nonzero CFG scales, FID decreases up to an optimal UG strength and then increases.
On CIFAR-10, where the baseline FID is already low, this trade-off may instead manifest as a slight increase in FID, consistent with the filtering behavior observed in \cref{subsec:filtering_exp}.

\cref{tab:cg_all} summarizes generation quality with and without UG under matched CFG scales, reporting the lowest-FID UG configuration for each CFG setting.
We additionally include uncertainty-aware guidance based on AU as a reference baseline.
AU leads to only marginal changes in generation metrics under flow matching, consistent with \cref{subsec:filtering_exp} where its uncertainty estimates correlate weakly with sample-level fidelity.

\Cref{fig:imagenet256_guidance_scales} shows that samples become more class-consistent and have less complex backgrounds as the UG scale increases.
These qualitative changes are consistent with the measured fidelity-diversity trade-off when steering toward low-uncertainty regions.
Overall, guidance based on a pseudo-likelihood derived from predicted velocity uncertainty improves sample fidelity in the tested settings.

\subsection{Uncertainty-Aware Classifier-Free Guidance}
\label{subsec:ucfg_exp}

\begin{figure}[t]
	\centering
	\includegraphics[width=0.9\textwidth]{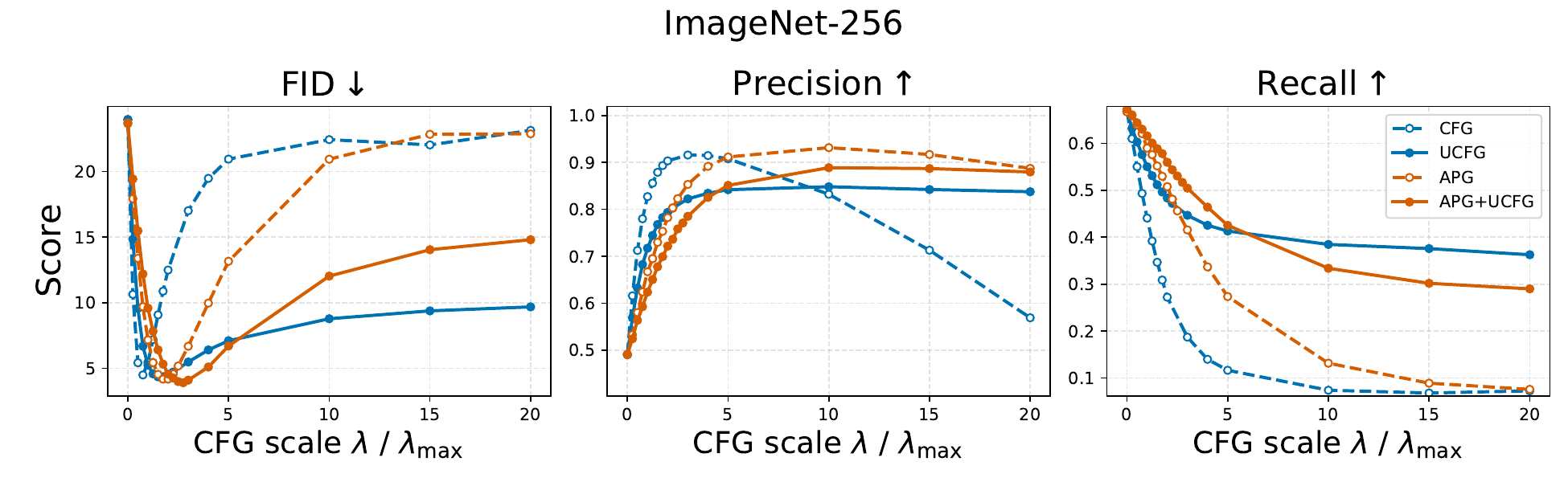}
	\caption{
		\textbf{FID, precision, and recall as a function of the fixed CFG scale $\lambda$ or the maximum UCFG scale $\lambda_{\max}$ on ImageNet-256.}
		CFG degrades sharply at large $\lambda$, while UCFG remains more stable as $\lambda_{\max}$ increases.
	}
	\label{fig:cfg_metrics_imagenet256}
\end{figure}

\begin{table}[t]
	\centering
	\renewcommand{\arraystretch}{0.9}
	\caption{\textbf{FID, precision and recall under CFG variants on ImageNet.}
		Best-performing scale (lowest FID) per method, averaged over 3 seeds.}
	\label{tab:cfg_scale}
	\begin{subtable}[t]{0.95\linewidth}
		\centering
		\caption{ImageNet-128}
		\begin{tabular}{lcccc}
			\toprule
			& $\lambda$/$\lambda_{\max}$ & FID$\downarrow$ & Precision$\uparrow$ & Recall$\uparrow$ \\
				\midrule
				CFG          & 1.0  & $5.26_{\pm.03}$ & $\mathbf{0.7287_{\pm.0019}}$ & $0.5005_{\pm.0033}$ \\
				UCFG         & 1.75 & $\mathbf{4.82_{\pm.04}}$ & $0.7092_{\pm.0005}$ & $\mathbf{0.5307_{\pm.0014}}$ \\
				\midrule
				APG          & 2.0  & $5.06_{\pm.02}$ & $\mathbf{0.7269_{\pm.0012}}$ & $0.5033_{\pm.0018}$ \\
				APG+UCFG     & 2.25 & $\mathbf{4.78_{\pm.01}}$ & $0.6976_{\pm.0036}$ & $\mathbf{0.5422_{\pm.0040}}$ \\
			\bottomrule
		\end{tabular}
	\end{subtable}
	\\[4pt]
	\begin{subtable}[t]{0.95\linewidth}
		\centering
		\caption{ImageNet-256}
		\begin{tabular}{lcccc}
			\toprule
			& $\lambda$/$\lambda_{\max}$ & FID$\downarrow$ & Precision$\uparrow$ & Recall$\uparrow$ \\
				\midrule
				CFG          & 0.75 & $4.48_{\pm.01}$ & $\mathbf{0.7809_{\pm.0019}}$ & $0.4984_{\pm.0022}$ \\
				UCFG         & 1.5  & $\mathbf{4.30_{\pm.01}}$ & $0.7660_{\pm.0013}$ & $\mathbf{0.5176_{\pm.0025}}$ \\
				\midrule
				APG          & 1.75 & $4.18_{\pm.01}$ & $0.7592_{\pm.0052}$ & $\mathbf{0.5246_{\pm.0048}}$ \\
				APG+UCFG     & 2.75 & $\mathbf{3.98_{\pm.07}}$ & $\mathbf{0.7740_{\pm.0029}}$ & $0.5117_{\pm.0050}$ \\
			\bottomrule
		\end{tabular}
	\end{subtable}
\end{table}

\begin{figure}[t]
	\centering
	\begin{subfigure}[t]{0.48\textwidth}
		\centering
		\includegraphics[width=\linewidth]{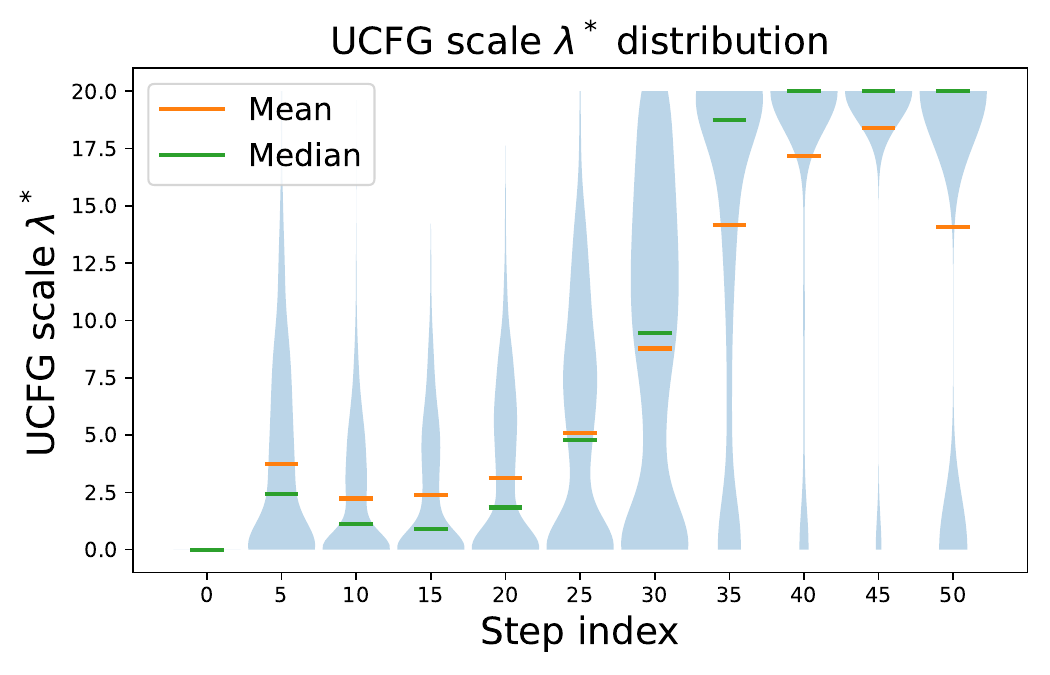}
		\caption{UCFG.}
		\label{fig:cfg_scale_violin_imagenet256}
	\end{subfigure}
	\hfill
	\begin{subfigure}[t]{0.48\textwidth}
		\centering
		\includegraphics[width=\linewidth]{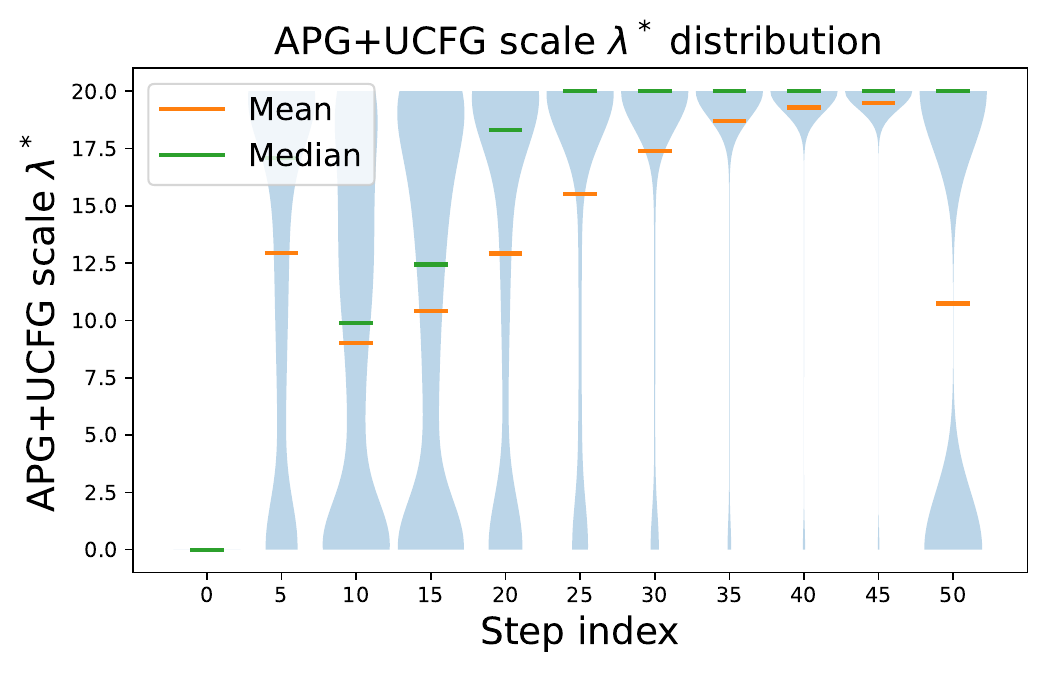}
		\caption{APG+UCFG.}
		\label{fig:cfg_scale_violin_adaptive_apg_imagenet256}
	\end{subfigure}
	\caption{
		\textbf{Distributions of the adaptive guidance scale $\lambda^*$ across sampling steps on ImageNet-256.}
		Violin plots are computed from 1{,}000 samples for (a) UCFG and (b) APG+UCFG.
		In both cases, $\lambda^*$ tends to be smaller in early steps and larger in later steps.
	}
	\label{fig:cfg_scale_violins_imagenet256}
\end{figure}

\begin{figure}[t]
	\centering
	\begin{subfigure}[t]{0.48\textwidth}
		\raggedleft
		\begin{overpic}[width=0.93\linewidth]{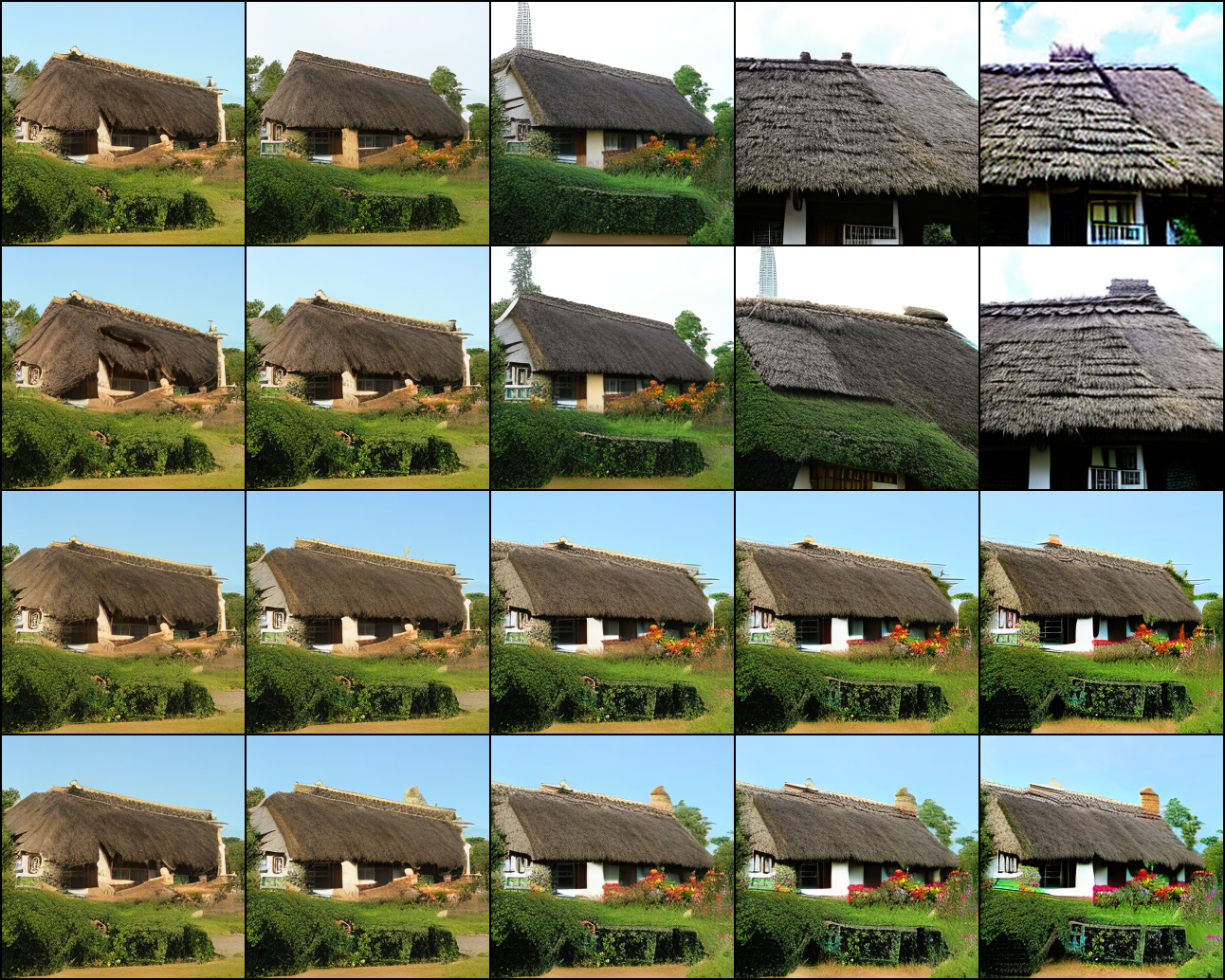}
			\put(-5,66){\rotatebox{90}{\scriptsize CFG}}
			\put(-5,46){\rotatebox{90}{\scriptsize APG}}
			\put(-5,25){\rotatebox{90}{\scriptsize UCFG}}
			\put(-8,4){\rotatebox{90}{\scriptsize\shortstack{APG+\\UCFG}}}
		\end{overpic}
	\end{subfigure}
	\hfill
	\begin{subfigure}[t]{0.48\textwidth}
		\raggedleft
		\begin{overpic}[width=0.93\linewidth]{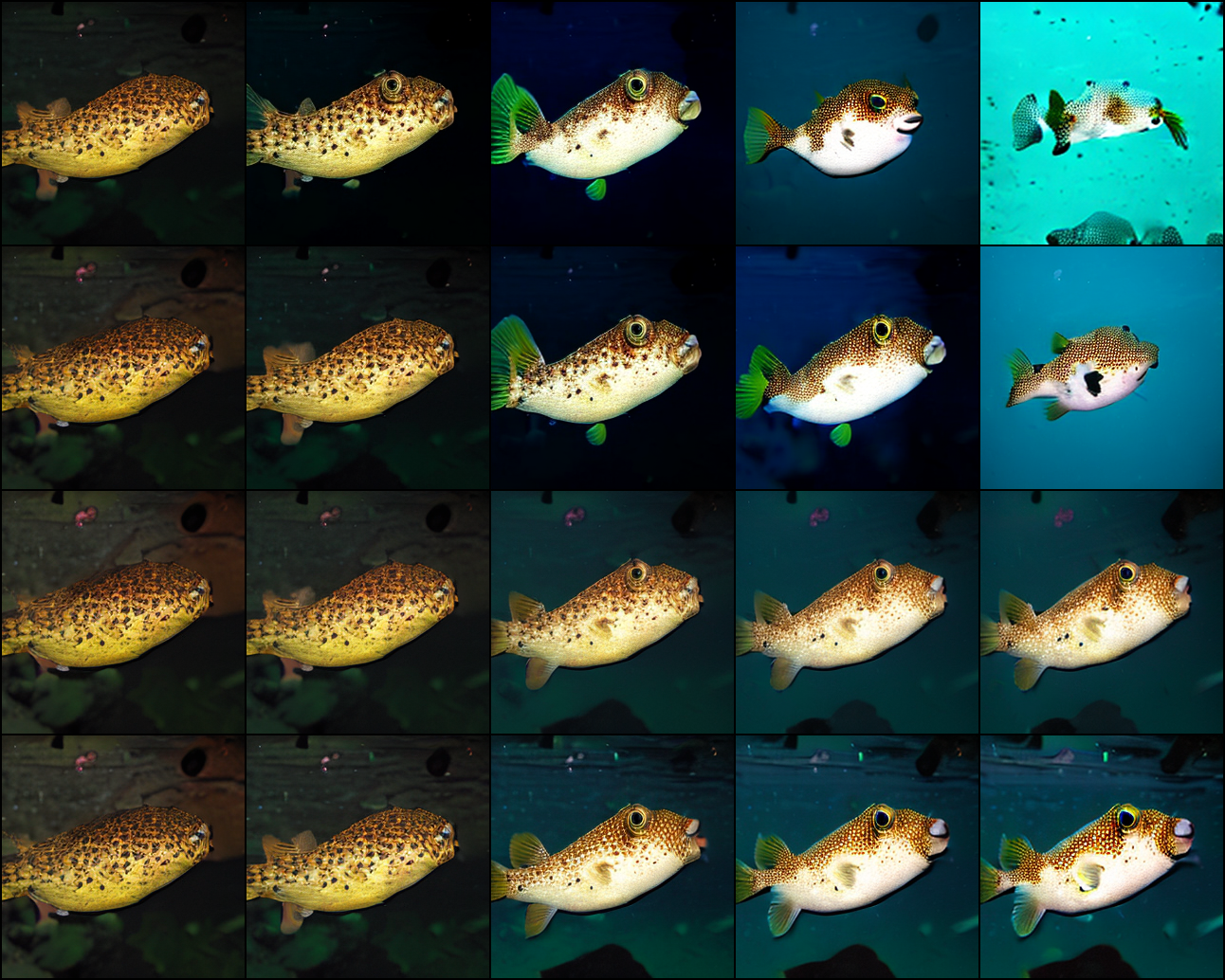}
			\put(-5,66){\rotatebox{90}{\scriptsize CFG}}
			\put(-5,46){\rotatebox{90}{\scriptsize APG}}
			\put(-5,25){\rotatebox{90}{\scriptsize UCFG}}
			\put(-8,4){\rotatebox{90}{\scriptsize\shortstack{APG+\\UCFG}}}
		\end{overpic}
	\end{subfigure}
	\caption{
		\textbf{ImageNet-256 samples under increasing classifier-free guidance scales.}
		Rows show CFG, APG, UCFG, and APG+UCFG from top to bottom; columns sweep the fixed scale $\lambda$ or adaptive-scale cap $\lambda_{\max}$ over $\{1,2,5,10,20\}$ from left to right.
		Compared to CFG and APG at large fixed scales $\lambda$, UCFG and APG+UCFG preserves structure more robustly under large $\lambda_{\max}$.
	}
	\label{fig:imagenet256_ucfg_scales}
\end{figure}

To observe the effect of UCFG, we compare standard CFG~\citep{Ho22CFG,Zheng23Guided} and APG~\citep{Sadat24APG} with their uncertainty-aware counterparts, UCFG and APG+UCFG, respectively.
We sweep the fixed guidance scale $\lambda$ for CFG and APG and the cap $\lambda_{\max}$ for UCFG and APG+UCFG.
The sweep covers $\{0, 0.25, \ldots, 2.0, 3, 4, 5, 10, 15, 20\}$ on ImageNet-128 and ImageNet-256.
Here, $\lambda_{\max}$ bounds the adaptive scale $\lambda^*$ selected from predicted uncertainty.

To reduce oversaturation and artifacts, APG applies guidance along the component of the CFG update orthogonal to the conditional velocity.
APG+UCFG adapts APG's guidance scale to minimize predicted uncertainty using the same mechanism as UCFG.
Implementation details for APG and APG+UCFG are provided in \cref{app:subsec:apg}.

\paragraph{Results.}
\Cref{fig:cfg_metrics_imagenet256} compares the ImageNet-256 metrics as the guidance parameters $\lambda$ and $\lambda_{\max}$ increase, while \cref{tab:cfg_scale} reports the best settings on both ImageNet resolutions.
As the fixed guidance scale $\lambda$ increases, recall degrades sharply for both CFG and APG, and precision of CFG also declines at large $\lambda$, leading to a steep increase in FID.
In contrast, recall decreases more gradually for UCFG and APG+UCFG as $\lambda_{\max}$ increases, resulting in only mild FID degradation at large $\lambda_{\max}$.

\Cref{fig:cfg_scale_violins_imagenet256} shows violin plots of $\lambda^*$ for UCFG and APG+UCFG, collected every five steps from 1{,}000 generated samples.
We observe that $\lambda^*$ is typically smaller in early sampling steps and becomes larger in later steps.
This pattern suggests that adaptive scaling limits over-guidance while the sample is still coarse, which may help explain the smaller loss of recall at large $\lambda_{\max}$.
Qualitatively, large $\lambda$ in CFG and APG produces oversaturated samples or mode collapse, while UCFG and APG+UCFG preserve global structure with less saturation under large $\lambda_{\max}$ (\cref{fig:imagenet256_ucfg_scales}).

Finally, \cref{tab:cfg_scale} reports the best-performing configurations (lowest FID) for each method.
UCFG and APG+UCFG achieve lower FID than their respective counterparts, CFG and APG.
APG+UCFG achieves the lowest FID on both ImageNet resolutions, while UCFG and APG also improve over CFG by retaining higher recall at their optimal scales.

\subsection{Additional Analyses}
\label{subsec:additional_analyses}


\subsubsection{Ablations on design choices.}
We validate the design choices behind our main configurations (\cref{app:additional_results:uncertainty_ablations}).
Filtering quality changes little across the tested numbers of Hutchinson probes $S$, covariance approximations, variance propagation orders, uncertainty update intervals, and top-$k\%$ aggregation ratios.
These results support defaults that reduce computation or simplify the approximation.
UG also changes little across the tested functional forms of $f(\sigma^2)$ and guidance intervals, supporting application at every two sampling steps.
Without the bias-correction term $U_t$ in \cref{eq:cufm}, filtering and UG retain most of their gains, but UCFG no longer surpasses fixed CFG.
This ablation indicates that $U_t$ is needed for UCFG to outperform fixed CFG in the tested setting.

\subsubsection{Bias and Batch-Size Sensitivity of the Correction Estimator.}
\label{subsec:correction_estimator_bias}
\begin{table}[h]
	\centering
	\renewcommand{\arraystretch}{0.9}
	\caption{
		\textbf{Batch-size sensitivity of the correction estimator on ImageNet-256 latents.}
		Average bias-correction $U_t$ measured over $t \in [0, 0.1]$, where the correction is largest, together with the fraction of the reference value it recovers.
		The reference replaces the mini-batch estimate with a $65{,}536$-candidate estimate.
	}
	\label{tab:correction_bias}
	\begin{tabular}{ccc}
		\toprule
		Batch size $B$ & Average bias-correction $U_t$ & Recovered \\
		\midrule
		Reference & $-0.579$ & --- \\
		\midrule
		2       & $-0.100$ & 17.21\% \\
		16      & $-0.273$ & 47.21\% \\
		64      & $-0.349$ & 60.35\% \\
		256     & $-0.407$ & 70.21\% \\
		1{,}024 & $-0.454$ & \textbf{78.32\%} \\
		\bottomrule
	\end{tabular}
\end{table}
The correction term $U_t$ in \cref{eq:cufm} is formed from the reweighted mini-batch estimate $\hat{u}_t(\mathbf{x}_t)$ of \cref{eq:u_hat}.
Squaring that estimate introduces a bias since the surrogate $\hat{u}_t^2$ is biased for $u_t^2$ even when $\hat{u}_t$ is a consistent proxy for $u_t$.
The bias therefore is governed by the mini-batch estimator variance and decreases with the batch size $B$ (\cref{app:loss_deriv}).

\paragraph{Procedure.} We quantify this on ImageNet-256 latent features by averaging $U_t$ across a range of batch sizes $B$.
As a reference, we replace the mini-batch estimate with one formed from 65,536 candidates, evaluated on each of 1,024 queries.
Every tested batch size averages 16 independently sampled candidate pools per query.

\paragraph{Results.} The correction concentrates in the first 10\% of forward generation. Its reference magnitude is 0.579 over $t \in [0, 0.1]$, falls to 0.0146 over $t \in [0.1, 0.2]$, and is at most $3.60 \times 10^{-7}$ thereafter.
Over $t \in [0.1, 0.2]$, it is only 2.32\% of the predicted velocity uncertainty, and the error incurred at $B=1{,}024$ is 2.16\% of the same quantity, so batch size has little absolute effect once $t>0.1$.
This observation is consistent with \citet{Bertrand26Closed}, finding that the empirical marginal velocity becomes close to the conditional velocity over time in high dimensional space.

\cref{tab:correction_bias} shows that the correction is material over $t \in [0, 0.1]$ and that its estimate improves monotonically with the candidate batch size, recovering 17.2\% of the reference at $B = 2$ and 78.3\% at $B = 1{,}024$.
The average predicted velocity uncertainty over this interval is 0.310 with the correction, compared with 0.652 without it.
The variance correction also improves the best UCFG configuration from an FID of 4.67 to 4.30, outperforming fixed CFG at 4.48 (\cref{app:bias_correction}).
Larger batches therefore sharpen the estimate exactly where the correction is large, while batch size leaves it essentially unaffected over the rest of the trajectory.

\subsubsection{Calibration Analysis}
\begin{table}[t]
	\centering
	\renewcommand{\arraystretch}{0.9}
	\caption{
		\textbf{Calibration of UA-Flow on ImageNet-256 validation images}.
		Scores before/after post-hoc isotonic calibration at three noise levels $t$.
		Details and baselines in \cref{app:calibration_analysis}.
	}
	\label{tab:calibration_main}
	\begin{tabular}{c|cc|cc}
		\toprule
		& \multicolumn{2}{c|}{Before} & \multicolumn{2}{c}{After} \\
		$t$ & ECE$\downarrow$ & Brier$\downarrow$ & ECE$\downarrow$ & Brier$\downarrow$ \\
		\midrule
		0.5 & 0.1328 & 0.4771 & 0.0066 & 0.0515 \\
		0.7 & 0.0992 & 0.3281 & 0.0038 & 0.0492 \\
		0.9 & \textbf{0.0277} & \textbf{0.1009} & \textbf{0.0008} & \textbf{0.0481} \\
		\bottomrule
	\end{tabular}
\end{table}
Generative models learn a probability distribution rather than predict a target, so there is no fixed ground truth against which to calibrate UA-Flow's predicted variance.
We therefore use an indirect protocol on ImageNet-256 validation latents (\cref{app:calibration_analysis}): real latents are noised to time $t$ and recovered by the ODE, treating the clean latent as ground truth for the Expected Calibration Error (ECE) and Brier score of the predicted endpoint distribution.
Before calibration, both metrics improve monotonically with $t$ (\cref{tab:calibration_main}).
Post-hoc calibration further reduces both metrics at each tested noise level.
This reduction suggests that isotonic calibration can improve UA-Flow's predicted endpoint variance under this indirect reconstruction protocol.

\subsubsection{Beyond Image Generation.}
\begin{table}[h]
	\centering
	\renewcommand{\arraystretch}{0.9}
	\caption{
		\textbf{Push-T early termination.}
		Rollouts whose chunk uncertainty exceeds $\tau$ are excluded.
		\emph{Excl.\ \%} is the excluded fraction of $1{,}000$ rollouts, and \emph{Success \%} (maximum reward $>1.0$) and \emph{Mean MaxR} are measured over those retained.
		Full table in \cref{tab:pusht}.
	}
	\label{tab:pusht_main}
	\begin{tabular}{lccc}
		\toprule
		Threshold $\tau$ & Excl.\ \% & Success \%$\uparrow$ & Mean MaxR$\uparrow$ \\
		\midrule
		no removal & 0.0\% & 62.7\% & 0.9514 \\
		\midrule
		0.0010     & 87.8\% & \textbf{91.0}\% & \textbf{0.9702} \\
		0.0025     & 62.9\% & 77.6\%          & 0.9597          \\
		0.0050     & 29.2\% & 66.8\%          & 0.9536          \\
		0.0075     & 14.6\% & 63.6\%          & 0.9536          \\
		0.0100     &  7.9\% & 63.3\%          & 0.9521          \\
		\bottomrule
	\end{tabular}
\end{table}
As preliminary evidence that UA-Flow extends beyond images, we evaluate it on three non-image domains: a 2D checkerboard density (\cref{app:non_image:checkerboard}), a conditional time-series forecasting task (\cref{app:non_image:timeseries}), and a Push-T robot policy benchmark (\cref{app:non_image:pusht}).
In all three, the uncertainty scores distinguish retained samples by coverage, prediction error, or rollout success, respectively.
On the time-series task, discarding the top $10\%$ most uncertain predictions reduces MSE by $3.8\times$ (\cref{tab:timeseries_filtering}).
On Push-T, aborting rollouts when chunk-level uncertainty exceeds a threshold $\tau=0.001$ excludes $87.8\%$ of rollouts and raises the retained success rate from $62.7\%$ to $91.0\%$, with the mean maximum reward rising from $0.9514$ to $0.9702$ (\cref{tab:pusht_main}).
These results suggest that UA-Flow can provide reliability signals beyond image generation.
We treat these as preliminary evidence and leave thorough domain-specific studies to future work.

\section{Conclusion}
\label{sec:conclusion}
We propose \emph{uncertainty-aware flow matching} (UA-Flow), which predicts element-wise heteroscedastic velocity uncertainty and propagates it through deterministic flow dynamics.
This yields per-sample and spatially localized uncertainty for filtering and enables uncertainty-aware guidance (UG) and step-wise adaptive classifier-free guidance (UCFG).
Across CIFAR-10 and ImageNet, UA-Flow's uncertainty scores correlate with sample fidelity more closely than element-wise baselines; UA-Flow + CLIP is competitive with the scalar CLIP-based baseline at lower computational cost (\cref{tab:flops_comparison}).
Moreover, our results show that UA-Flow's uncertainty reflects the degree to which each generated sample is supported by the data manifold.
UG induces a precision-recall trade-off by steering toward low-uncertainty trajectories, while UCFG mitigates failures of large fixed guidance via adaptive scaling and remains robust under strong guidance.
A key limitation of our formulation is its reliance on simplifying assumptions made for tractability and efficiency, most notably bias-correction term, diagonal variance and first-order Taylor variance propagation. Relaxing these assumptions is left to future work.

\subsubsection*{Broader Impact Statement}
This work advances uncertainty quantification for sampling-based generative models by explicitly modeling and propagating uncertainty in flow matching dynamics.
By providing per-sample and spatially localized uncertainty estimates, the proposed approach can help practitioners assess the reliability of generated outputs and make more informed decisions when deploying generative models.

Potential positive impacts include improved robustness and safety in downstream applications that rely on generative models.
In particular, uncertainty-aware guidance may reduce failure cases caused by overconfident or excessively guided generation.

At the same time, as with other advances in generative modeling, improved generation quality and controllability may amplify existing societal risks associated with synthetic data, including misuse, misinformation, or overreliance on automatically generated content.
We emphasize that uncertainty estimates should be used as a complementary reliability signal rather than a guarantee of correctness.

\subsubsection*{Acknowledgments}
This work is supported by Hyundai Motor Company.


\bibliography{root}
\bibliographystyle{tmlr}

\newpage
\appendix

\etocdepthtag.toc{appendix}
\etocsettocstyle{\section*{\color{appendixtocblue}Appendix Contents}}{}
\etocsetstyle{section}
  {}
  {\noindent\normalsize\bfseries\color{appendixtocblue}\etocnumber\enspace}
  {\color{appendixtocblue}\etocname\hfill\etocpage\par}
  {}
\etocsetstyle{subsection}
  {}
  {\noindent\hspace*{1.5em}\small\bfseries\color{appendixtocblue}\etocnumber\enspace}
  {\color{appendixtocblue}\etocname\hfill\etocpage\par}
  {}
\etocsetnexttocdepth{subsection}
\begingroup
\hypersetup{hidelinks}
\tableofcontents
\endgroup
\newpage

\section{Derivation of the Uncertainty-Aware Flow Matching Loss}
\label{app:loss_deriv}
This appendix provides a detailed derivation of the conditional uncertainty-aware flow matching loss $\mathcal{L}_{\mathrm{CUFM}}$ by rewriting the unconditional Gaussian negative log-likelihood loss $\mathcal{L}_{\mathrm{UFM}}$ in terms of conditional flow matching.
The key idea is to express expectations under the marginal distribution $p_t(\mathbf{x})$ using the conditional distribution $p_t(\mathbf{x} \mid \mathbf{x}_1)$, which enables tractable training despite the inaccessibility of the marginal velocity.
For notational simplicity, time is uniformly sampled, i.e. $t \sim U([0,1])$, though any alternative time-sampling distribution could be used.

We begin by expanding the uncertainty-aware flow matching loss $\mathcal{L}_{\mathrm{UFM}}$ and decomposing it into four expectation terms, which will later be rewritten under the conditional distribution.
\begin{equation}
	\begin{aligned}
		&\mathcal{L}_{\mathrm{UFM}}(\theta)
		= \E_{t,\,p_t(\mathbf{x}_t)}\Big[\,
		\frac{\big(\bar{u}^{\theta}_t(\mathbf{x}_t) - u_t(\mathbf{x}_t)\big)^2}{2(\sigma_t^\theta(\mathbf{x}_t))^2} + \log(\sigma_t^\theta(\mathbf{x}_t))
		\Big] \\
		=& \underbrace{\E_{t,\,p_t(\mathbf{x}_t)}\Big[\,
			\frac{(\bar{u}^{\theta}_t(\mathbf{x}_t))^2}{2(\sigma_t^\theta(\mathbf{x}_t))^2}
			\Big]}_{(A)}
		-2\underbrace{\E_{t,\,p_t(\mathbf{x}_t)}\Big[\,
			\frac{\bar{u}^{\theta}_t(\mathbf{x}_t)u_t(\mathbf{x}_t)}{2(\sigma_t^\theta(\mathbf{x}_t))^2}
			\Big]}_{(B)} \\
		+&\underbrace{\E_{t,\,p_t(\mathbf{x}_t)}\Big[\, 
			\frac{(u_t(\mathbf{x}_t)\big)^2}{2(\sigma_t^\theta(\mathbf{x}_t))^2}
			\Big]}_{(C)}
		+\underbrace{ \E_{t,\,p_t(\mathbf{x}_t)}\Big[\,
			\log(\sigma_t^\theta(\mathbf{x}_t))
			\Big]}_{(D)} 
	\end{aligned}
\end{equation}

Term (A) depends only on the predicted mean and variance and can be rewritten by expressing the marginal distribution $p_t(\mathbf{x}_t)$ as an integral over the conditional distribution $p_t(\mathbf{x}_t \mid \mathbf{x}_1)$ and the data distribution $p_1(\mathbf{x}_1)$.
\begin{equation}
	\begin{aligned}
	(A) &= \int \frac{(\bar{u}^{\theta}_t(\mathbf{x}_t))^2}{2(\sigma_t^\theta(\mathbf{x}_t))^2} p_t(\mathbf{x}_t) d\mathbf{x}_tdt \\
	&=  \int \frac{(\bar{u}^{\theta}_t(\mathbf{x}_t))^2}{2(\sigma_t^\theta(\mathbf{x}_t))^2} p_t( \mathbf{x}_t \mid \mathbf{x}_1)p_1(\mathbf{x}_1) d\mathbf{x}_1d\mathbf{x}_tdt \\
	&= \mathbb{E}_{t, p_1(\mathbf{x}_1), p_t( \mathbf{x}_t \mid \mathbf{x}_1)}\Big[\,\frac{(\bar{u}^{\theta}_t(\mathbf{x}_t))^2}{2(\sigma_t^\theta(\mathbf{x}_t))^2}
	 \Big]
	\end{aligned}
\end{equation}

Term (B) involves the cross term between the predicted mean and the true marginal velocity.
Since the marginal velocity $u_t(\mathbf{x})$ is intractable, we rewrite it using the law of total expectation under the conditional flow matching formulation.
\begin{equation}
	\begin{aligned}
		(B) &= \int \frac{\bar{u}^{\theta}_t(\mathbf{x}_t)u_t(\mathbf{x}_t)}{2(\sigma_t^\theta(\mathbf{x}_t))^2} p_t(\mathbf{x}_t) d\mathbf{x}_tdt \\
		&= \int \frac{\bar{u}^{\theta}_t(\mathbf{x}_t)}{2(\sigma_t^\theta(\mathbf{x}_t))^2}
		\Big( \int \frac{u_t(\mathbf{x}_t \mid \mathbf{x}_1)p_t(\mathbf{x}_t \mid \mathbf{x}_1)p_1(\mathbf{x}_1)}{p_t(\mathbf{x}_t)} d\mathbf{x}_1 \Big)
		p_t(\mathbf{x}_t) d\mathbf{x}_tdt \\
		&= \int \frac{\bar{u}^{\theta}_t(\mathbf{x}_t)u_t(\mathbf{x}_t \mid \mathbf{x}_1)}{2(\sigma_t^\theta(\mathbf{x}_t))^2} p_t(\mathbf{x}_t \mid \mathbf{x}_1)p_1(\mathbf{x}_1) d\mathbf{x}_1d\mathbf{x}_tdt \\
		&= \mathbb{E}_{t, p_1(\mathbf{x}_1), p_t( \mathbf{x}_t \mid \mathbf{x}_1)}\Big[\,
		\frac{\bar{u}^{\theta}_t(\mathbf{x}_t)u_t(\mathbf{x}_t \mid \mathbf{x}_1)}{2(\sigma_t^\theta(\mathbf{x}_t))^2} \Big]
	\end{aligned}
\end{equation}

By the same change of measure, terms (C) and (D) can be rewritten as
\begin{equation}
	(C) = \E_{t, p_1(\mathbf{x}_1), p_t( \mathbf{x}_t \mid \mathbf{x}_1)}\Big[\,  \frac{(u_t(\mathbf{x}_t))^2}{2(\sigma_t^\theta(\mathbf{x}_t))^2} \Big]
\end{equation}

\begin{equation}
(D) = \E_{t, p_1(\mathbf{x}_1), p_t( \mathbf{x}_t \mid \mathbf{x}_1)}\Big[\,  \log(\sigma_t^\theta(\mathbf{x}_t)) \Big]
\end{equation}

Therefore, $\mathcal{L}_{UFM}(\theta)$ can be rewritten as:
\begin{equation}
\begin{aligned}
	 &\mathcal{L}_{\mathrm{UFM}}(\theta)
	=  \E_{t, p_1(\mathbf{x}_1), p_t( \mathbf{x}_t \mid \mathbf{x}_1)}\Big[\,
	\frac{(\bar{u}^{\theta}_t(\mathbf{x}_t))^2 - 2\bar{u}^{\theta}_t(\mathbf{x}_t)u_t(\mathbf{x}_t \mid \mathbf{x}_1) + (u_t(\mathbf{x}_t))^2}{2(\sigma_t^\theta(\mathbf{x}_t))^2} + \log(\sigma_t^\theta(\mathbf{x}_t))
	\Big] \\
	&= \E_{t, p_1(\mathbf{x}_1), p_t( \mathbf{x}_t \mid \mathbf{x}_1)}\Big[\,
	\frac{(\bar{u}^{\theta}_t(\mathbf{x}_t) - u_t(\mathbf{x}_t \mid \mathbf{x}_1))^2  + (u_t(\mathbf{x}_t))^2 - (u_t(\mathbf{x}_t \mid \mathbf{x}_1))^2}{2(\sigma_t^\theta(\mathbf{x}_t))^2} + \log(\sigma_t^\theta(\mathbf{x}_t))
	\Big]
	\end{aligned}
\end{equation}

However, we cannot evaluate the true marginal velocity $u_t(\mathbf{x}_t)$ in closed form.
Using the identity in \cref{eq:u_expansion}, we can rewrite $u_t(\mathbf{x}_t)$ as a ratio of expectations over $\mathbf{x}_1 \sim p_1$:
\begin{equation}
	\label{eq:u_expansion}
	\begin{aligned}
		u_t(\mathbf{x}_t) 
		&= \int \frac{u_t(\mathbf{x}_t \mid \mathbf{x}_1)p_t(\mathbf{x}_t \mid \mathbf{x}_1)p_1(\mathbf{x}_1)}{p_t(\mathbf{x}_t)}d\mathbf{x}_1 \\
		&= \frac{\int u_t(\mathbf{x}_t \mid \mathbf{x}_1)p_t(\mathbf{x}_t \mid \mathbf{x}_1)p_1(\mathbf{x}_1) d\mathbf{x}_1}{\int p_t(\mathbf{x}_t \mid \mathbf{x}_1)p_1(\mathbf{x}_1) d\mathbf{x}_1} \\
		&=\frac{\mathbb{E}_{p_1(\mathbf{x}_1)}[u_t(\mathbf{x}_t \mid \mathbf{x}_1)p_t(\mathbf{x}_t \mid \mathbf{x}_1)]}{\mathbb{E}_{p_1(\mathbf{x}_1)}[p_t(\mathbf{x}_t \mid \mathbf{x}_1)]}.
	\end{aligned}
\end{equation}
This suggests a self-normalized importance-sampling estimator based on a mini-batch
$\{\mathbf{x}_{1,b}\}_{b=1}^B \sim p_1$:
\begin{equation}
	\label{eq:u_hat_app}
	\hat{u}_t(\mathbf{x}_t)
	=
	\frac{\sum_{b=1}^B u_t(\mathbf{x}_t \mid \mathbf{x}_{1,b})\, p_t(\mathbf{x}_t \mid \mathbf{x}_{1,b})}
	{\sum_{b=1}^B p_t(\mathbf{x}_t \mid \mathbf{x}_{1,b})}.
\end{equation}
Since $u_t(\mathbf{x}_t)$ is intractable, we approximate $(u_t(\mathbf{x}_t))^2$ by $(\hat{u}_t(\mathbf{x}_t))^2$ in our objective.
This ratio-of-expectations naturally motivates an importance-weighted approximation, providing a tractable proxy for the unconditional target. Substituting $\hat{u}_t(\mathbf{x}_t)^2$ yields the correction term
$U_t(\mathbf{x}_t,\mathbf{x}_1):= \hat{u}_t(\mathbf{x}_t)^2 - u_t(\mathbf{x}_t \mid \mathbf{x}_1)^2$
and results in the conditional objective $\mathcal{L}_{\mathrm{CUFM}}(\theta)$ in \cref{eq:cufm}.

\paragraph{Remark on Jensen bias in approximating $(u_t(\mathbf{x}_t))^2$ and why we still keep $U_t(\mathbf{x}_t,\mathbf{x}_1)$.}
However, $\hat u_t(\mathbf{x}_t)^2$ is not an unbiased estimator of $u_t(\mathbf{x}_t)^2$ even when $\hat u_t(\mathbf{x}_t)$ is a consistent proxy for $u_t(\mathbf{x}_t)$.
This follows from the identity $\E[\hat u^2]-(\E[\hat u])^2=\Var(\hat u)$, i.e., squaring introduces a Jensen gap proportional to the estimator variance.
Accordingly, the bias is controlled by the (mini-batch) estimator variance and typically decreases as the mini-batch size $B$ increases.

Despite this limitation, introducing the correction term
$U_t(\mathbf{x}_t,\mathbf{x}_1) := \hat u_t(\mathbf{x}_t)^2 - u_t(\mathbf{x}_t \mid \mathbf{x}_1)^2$
still yields a closer surrogate to the original unconditional objective than omitting $U_t$ altogether.
Indeed, letting $p_{t\vert1}(\mathbf{x}_1 \mid \mathbf{x}_t) = \frac{p_t(\mathbf{x}_t \mid \mathbf{x}_1)\,p_1(\mathbf{x}_1)}{p_t(\mathbf{x}_t)}$ denote the induced posterior in \cref{eq:u_expansion}, we have
$u_t(\mathbf{x}_t)=\E_{p_{1\vert t}(\mathbf{x}_1 \mid \mathbf{x}_t)}[u_t(\mathbf{x}_t\mid \mathbf{x}_1)]$ and thus
\[
\E_{p_{1\vert t}(\mathbf{x}_1 \mid \mathbf{x}_t)}\!\left[u_t(\mathbf{x}_t)^2 - u_t(\mathbf{x}_t\mid \mathbf{x}_1)^2\right]
= -\,\Var_{p_{1\vert t}(\mathbf{x}_1 \mid \mathbf{x}_t)}\!\left(u_t(\mathbf{x}_t\mid \mathbf{x}_1)\right)\ \le\ 0.
\]
Therefore, dropping $U_t$ implicitly sets this negative term to zero, incurring a systematic bias that does not vanish with $B$.
In contrast, our proxy retains this variance-related correction up to the residual bias in $\hat u_t(\mathbf{x}_t)^2$, which diminishes as the mini-batch estimator variance decreases (e.g., as $B$ increases).

\section{Details on Variance Propagation and Covariance Approximations}
\label{app:variance_propagation}
This appendix provides detailed derivations for the variance evolution equations and tractable approximations of the covariance between the state and the velocity field. 

\subsection{Derivations of the Mean and Variance Propagation Rules}

\paragraph{Mean propagation (\cref{eq:mean_ode}).}
Using the Gaussian velocity model $u_t^\theta(\mathbf{x})=\bar{u}_t^\theta(\mathbf{x}_t)+\sigma_t^\theta(\mathbf{x}_t)\odot\epsilon$ with $\epsilon\sim\mathcal{N}(0,I)$, we have
$\mathbb{E}[u_t^\theta(\mathbf{x}_t)]=\mathbb{E}[\bar{u}_t^\theta(\mathbf{x}_t)]$.
Applying a first-order Taylor expansion of $\bar{u}^\theta_t(\mathbf{x}_t)$ around $\bar{\mathbf{x}}_t:=\mathbb{E}[\mathbf{x}_t]$ yields
\begin{equation}
\label{eq:mean_ode_deriv}
	\dd{\bar{\mathbf{x}}_t}{t} 
	= \E\big[u_t^{\theta}(\mathbf{x}_t)\big] 
	= \E\big[ \bar{u}^{\theta}_t(\mathbf{x}_t) \big]
	\approx \E\big[ \bar{u}^{\theta}_t(\bar{\mathbf{x}}_t) + J^\theta_t(\bar{\mathbf{x}}_t)(\mathbf{x}_t - \bar{\mathbf{x}}_t) \big] 
	= \bar{u}^{\theta}_t(\bar{\mathbf{x}}_t).
\end{equation}

\paragraph{Variance propagation (\cref{eq:variance_propagation}).}
Under Euler integration between times $t$ and $t + \Delta t$, the flow dynamics become
\begin{equation} \label{eq:euler_flow}
	\mathbf{x}_{t + \Delta t} = \mathbf{x}_t + u^\theta_t(\mathbf{x}_t)\, \Delta t.
\end{equation}
Applying the element-wise variance identity $\Var(X+aY)=\Var(X)+a^2\Var(Y)+2a\,\Cov(X,Y)$ to \cref{eq:euler_flow} yields \cref{eq:variance_propagation}, where $Y=u_t^\theta(\mathbf{x}_t)$.

Next, we justify the approximation $\Var(u_t^\theta(\mathbf{x}_t)) \approx (\sigma_t^\theta(\bar{\mathbf{x}}_t))^2$ used in the main text.
By the law of total variance and $u^\theta_t(\mathbf{x}_t) = \bar{u}^\theta_t(\mathbf{x}_t) + \sigma^\theta_t(\mathbf{x}_t) \odot \epsilon$ with $\epsilon\sim\mathcal{N}(0,I)$, we have
\begin{equation}
	\label{eq:variance_approx}
	\begin{aligned}
		\Var(u^\theta_t(\mathbf{x}_t)) 
		=& \E[\Var(u^\theta_t(\mathbf{x}_t) \mid \mathbf{x}_t)] + 
		\Var (\E[u^\theta_t(\mathbf{x}_t) \mid \mathbf{x}_t] ) \\
		=& \E_{\mathbf{x}_t}\big[ (\sigma^\theta_t(\mathbf{x}_t))^2 \big] + \Var(\bar{u}^\theta_t(\mathbf{x}_t)) \\
		\approx& \E_{\mathbf{x}_t}\Big[ (\sigma^\theta_t(\bar{\mathbf{x}}_t))^2 
		+ \pp{(\sigma^\theta_t)^2}{\mathbf{x}}\Big|_{\bar{\mathbf{x}}_t}(\mathbf{x}_t - \bar{\mathbf{x}}_t) \Big] +
		\Var(\bar{u}^\theta_t(\mathbf{x}_t)) \\
		=& (\sigma^\theta_t(\bar{\mathbf{x}}_t))^2 + \Var(\bar{u}^\theta_t(\mathbf{x}_t))
	\end{aligned}
\end{equation}
where the last step follows by dropping higher-order terms and using $\E[\mathbf{x}_t-\bar{\mathbf{x}}_t]=0$.
The remaining term $\Var(\bar{u}^\theta_t(\mathbf{x}_t))$ captures variance induced by the spread of $\mathbf{x}_t$.
In principle, it can be estimated by Monte Carlo sampling:
draw $\mathbf{x}_{t,i}\sim\mathcal{N}(\bar{\mathbf{x}}_t,\Var[\mathbf{x}_t])$ and compute the empirical variance of $\bar{u}^\theta_t(\mathbf{x}_{t,i})$.

In the main text, we focus on the predicted heteroscedastic uncertainty and avoid this additional Monte Carlo overhead.
Thus, we neglect $\Var(\bar{u}^\theta_t(\mathbf{x}_t))$ and use the approximation
\begin{equation*}
\Var(u^\theta_t(\mathbf{x}_t)) \approx (\sigma^\theta_t(\bar{\mathbf{x}}_t))^2.
\end{equation*}
\cref{app:additional_results:uncertainty_ablations:variance} further shows that explicitly including
$\Var(\bar{u}^\theta_t(\mathbf{x}_t))$ has negligible empirical effect on the resulting uncertainty estimates.

\paragraph{Taylor covariance approximation (\cref{eq:cov_approx_taylor}).}
We write $u^\theta_t(\mathbf{x}_t) = \bar{u}^{\theta}_t(\mathbf{x}_t) + \sigma^\theta_t(\mathbf{x}_t) \odot \epsilon$ with $\epsilon \sim \mathcal{N}(0, I)$.
The noise term does not contribute to the covariance because $\epsilon$ is independent of $\mathbf{x}_t$ and has zero mean.
Thus,
\begin{equation} \label{eq:cov_expansion}
	\begin{aligned}
		\Cov(\mathbf{x}_t, u^{\theta}_t(\mathbf{x}_t)) 
		&= \E[\mathbf{x}_t \odot u^{\theta}_t(\mathbf{x}_t)] - \E[\mathbf{x}_t] \odot \E[u^{\theta}_t(\mathbf{x}_t)] \\
		&= \E_{\mathbf{x}_t}[\mathbf{x}_t \odot \bar{u}^{\theta}_t(\mathbf{x}_t)] - \bar{\mathbf{x}}_t \odot \E_{\mathbf{x}_t}[\bar{u}^{\theta}_t(\mathbf{x}_t)] \\
		&\approx \E_{\mathbf{x}_t}[\mathbf{x}_t \odot \bar{u}^{\theta}_t(\mathbf{x}_t)] - \bar{\mathbf{x}}_t \odot \bar{u}^{\theta}_t(\bar{\mathbf{x}}_t).
	\end{aligned}
\end{equation}

Applying a first-order Taylor expansion of $\bar{u}^{\theta}_t(\mathbf{x}_t)$ around $\bar{\mathbf{x}}_t$ gives
\begin{equation}
	\begin{aligned} \label{eq:cov_lhs_approx}
		\E_{\mathbf{x}_t}\big[
		\mathbf{x}_t \odot \bar{u}^{\theta}_t(\mathbf{x}_t) \big]
		&\approx \E_{\mathbf{x}_t}\Big[ 
		\mathbf{x}_t \odot \big(\bar{u}^{\theta}_t(\bar{\mathbf{x}}_t) + J^{\theta}_t(\bar{\mathbf{x}}_t)\, (\mathbf{x}_t - \bar{\mathbf{x}}_t) \big)
		\Big] \\
		&= \bar{\mathbf{x}}_t \odot \bar{u}^{\theta}_t(\bar{\mathbf{x}}_t) +
		\E_{\mathbf{x}_t}\Big[
		(\mathbf{x}_t - \bar{\mathbf{x}}_t) \odot \big(J^{\theta}_t(\bar{\mathbf{x}}_t)\, (\mathbf{x}_t - \bar{\mathbf{x}}_t)\big)
		\Big].
	\end{aligned}
\end{equation}

To keep the propagation tractable in high dimensions, we approximate $\Cov(\mathbf{x}_t)\in\mathbb{R}^{n\times n}$ as diagonal (i.e., we neglect off-diagonal entries).
Then the $i$-th element of the last expectation becomes
\begin{equation} \label{eq:cov_to_diag_jacobian1}
	\begin{aligned}
		\mathbb{E}\Big[
		(\mathbf{x}_t - \bar{\mathbf{x}}_t)_i \,\big(J^{\theta}_t(\bar{\mathbf{x}}_t)\, (\mathbf{x}_t - \bar{\mathbf{x}}_t)\big)_i 
		\Big]
		&=
		\sum_{j=1}^n
		(J^{\theta}_t(\bar{\mathbf{x}}_t))_{ij} \, \mathbb{E}\Big[
		(\mathbf{x}_t - \bar{\mathbf{x}}_t)_i \, (\mathbf{x}_t - \bar{\mathbf{x}}_t)_j
		\Big] \\
		&=  \sum_{j=1}^n (J^{\theta}_t(\bar{\mathbf{x}}_t))_{ij} \, (\Cov(\mathbf{x}_t))_{ij}
		= (J^{\theta}_t(\bar{\mathbf{x}}_t))_{ii} \, \Var[\mathbf{x}_t]_i.
	\end{aligned}
\end{equation}

Therefore,
\begin{equation} \label{eq:cov_to_diag_jacobian2}
	\E_{\mathbf{x}_t}\big[
	\mathbf{x}_t \odot \bar{u}^{\theta}_t(\mathbf{x}_t) \big] 
	\approx \bar{\mathbf{x}}_t \odot \bar{u}^{\theta}_t(\bar{\mathbf{x}}_t) +
	\diag(J^{\theta}_t(\bar{\mathbf{x}}_t)) \odot \Var[\mathbf{x}_t].
\end{equation}

Combining \cref{eq:cov_expansion,eq:cov_lhs_approx,eq:cov_to_diag_jacobian1,eq:cov_to_diag_jacobian2} yields \cref{eq:cov_approx_taylor}.

\paragraph{Jacobian vector product approximation (\cref{eq:cov_approx_jvp}).}
Let $\mathbf{r}\in\mathbb{R}^n$ be a Rademacher vector with independent entries sampled uniformly from $\{-1,+1\}$, so that $\E[r_i r_j]=\delta_{ij}$.
Define $\boldsymbol{\sigma}_t^x := \sqrt{\Var[\mathbf{x}_t]}$ and $\mathbf{v} := \boldsymbol{\sigma}_t^x \odot \mathbf{r}$.
Then, for each coordinate $i$,
\[
\mathbb{E}\big[v_i (J_t^\theta(\bar{\mathbf{x}}_t)\mathbf{v})_i\big] 
= \sum_{j=1}^n (J_t^\theta(\bar{\mathbf{x}}_t))_{ij}\, \mathbb{E}[v_i v_j]
= (J_t^\theta(\bar{\mathbf{x}}_t))_{ii}\, (\boldsymbol{\sigma}_t^x)_i^2,
\]
since $\mathbb{E}[v_i v_j] = (\boldsymbol{\sigma}_t^x)_i(\boldsymbol{\sigma}_t^x)_j\,\mathbb{E}[r_i r_j]$ and $\mathbb{E}[r_i r_j]=\delta_{ij}$.
Stacking all coordinates gives
\begin{equation}
	\mathbb{E}\big[ \mathbf{v} \odot (J^{\theta}_t(\bar{\mathbf{x}}_t)\mathbf{v})  \big]
	= \diag(J^{\theta}_t(\bar{\mathbf{x}}_t)) \odot \Var[\mathbf{x}_t].
\end{equation}
Therefore, \cref{eq:cov_approx_jvp} provides an unbiased Monte Carlo estimator of $\diag(J^{\theta}_t(\bar{\mathbf{x}}_t)) \odot \Var[\mathbf{x}_t]$ using only Jacobian--vector products.

\subsection{Approximations of $\Cov(\mathbf{x}_t, u^\theta_t(\mathbf{x}_t))$}
\label{app:cov_approximation}
We approximate $\Cov(\mathbf{x}_t, u^{\theta}_t(\mathbf{x}_t))$ in three tractable ways:
\begin{equation}
	\label{eq:cov_approximation}
	\Cov(\mathbf{x}_t, u^{\theta}_t(\mathbf{x}_t)) \approx
	\begin{cases}
		0 & (\text{Option 1}) \\[2mm]
		\frac{1}{S}\sum_{i=1}^S(\boldsymbol{\sigma}^x_t \odot \mathbf{r}_i) \odot 
		\Big(J^{\theta}_t(\bar{\mathbf{x}}_t)(\boldsymbol{\sigma}^x_t \odot \mathbf{r}_i)\Big) & (\text{Option 2}) \\[2mm]
		\frac{1}{S}\sum_{i=1}^S \mathbf{x}_{t,i} \odot \bar{u}^{\theta}_t(\mathbf{x}_{t,i}) 
		- \bar{\mathbf{x}}_t \odot \Big(\frac{1}{S}\sum_{i=1}^S \bar{u}^{\theta}_t(\mathbf{x}_{t,i})\Big) 
		& (\text{Option 3})\\
	\end{cases}
\end{equation}
where $\mathbf{r}_i$ are Rademacher vectors and $\mathbf{x}_{t,i} \sim \mathcal{N}(\bar{\mathbf{x}}_t, \diag(\Var[\mathbf{x}_t]))$.
Here $S$ denotes the number of samples.

\textbf{Option 1.} Ignore the covariance term. This is the cheapest choice computationally, but it discards the interaction between state and flow.

\textbf{Option 2.} This is the estimator deployed in our implementation. 
It estimates $\diag(J^{\theta}_t(\bar{\mathbf{x}}_t))\odot \Var[\mathbf{x}_t]$ using $S$ Rademacher probes and  Jacobian-vector products; see Algorithm~\ref{alg:jacobian_diag_sigma2} for details.

\textbf{Option 3.} A Monte Carlo alternative, similar to BayesDiff~\citep{Kou23Bayesdiff}, draws $\mathbf{x}_{t,i} \sim \mathcal{N}(\bar{\mathbf{x}}_t, \Var[\mathbf{x}_t])$ for $i=1,\cdots,S$ and estimates the covariance directly from sample moments.

\section{Variance Propagation under Heun's 2nd-Order Sampling}
\label{app:heun_variance_propagation}
This appendix extends the variance propagation rule of \cref{app:variance_propagation} from the Euler discretization to Heun's 2nd-order integrator.
We follow the same conventions as in the main text: all operations are element-wise, $\Var[\cdot]$ and $\Cov(\cdot, \cdot)$ denote the element-wise (diagonal) variance and covariance, and the Gaussian velocity model is $u^\theta_t(\mathbf{x}) = \bar{u}^\theta_t(\mathbf{x}) + \sigma^\theta_t(\mathbf{x}) \odot \epsilon$ with $\epsilon \sim \mathcal{N}(0, I)$.

\paragraph{Heun's 2nd-order update.}
Given a step size $\Delta t$, Heun's 2nd-order (Heun2) method uses one Euler predictor and a trapezoidal corrector:
\begin{equation}
	\label{eq:heun_update}
	\begin{aligned}
		\mathbf{k}_1 \;&=\; u^\theta_t(\mathbf{x}_t) \;=\; \bar{u}^\theta_t(\mathbf{x}_t) + \sigma^\theta_t(\mathbf{x}_t) \odot \epsilon_1, \\
		\tilde{\mathbf{x}}_{t + \Delta t} \;&=\; \mathbf{x}_t + \mathbf{k}_1\, \Delta t, \\
		\mathbf{k}_2 \;&=\; u^\theta_{t + \Delta t}(\tilde{\mathbf{x}}_{t + \Delta t})
		\;=\; \bar{u}^\theta_{t + \Delta t}(\tilde{\mathbf{x}}_{t + \Delta t}) + \sigma^\theta_{t + \Delta t}(\tilde{\mathbf{x}}_{t + \Delta t}) \odot \epsilon_2, \\
		\mathbf{x}_{t + \Delta t} \;&=\; \mathbf{x}_t + \tfrac{\Delta t}{2}\big(\mathbf{k}_1 + \mathbf{k}_2\big),
	\end{aligned}
\end{equation}
with independent $\epsilon_1, \epsilon_2 \sim \mathcal{N}(0, I)$ sampled at the two stages.

For notational brevity we define the Euler-predicted mean state
\begin{equation}
	\label{eq:heun_mean_predictor}
	\bar{\tilde{\mathbf{x}}}_{t + \Delta t} \;:=\; \E[\tilde{\mathbf{x}}_{t + \Delta t}]
	\;\approx\; \bar{\mathbf{x}}_t + \bar{u}^\theta_t(\bar{\mathbf{x}}_t)\, \Delta t,
\end{equation}
which follows from the same linearization as in \cref{eq:mean_ode_deriv}, and abbreviate
\begin{equation*}
	\sigma_1 := \sigma^\theta_t(\bar{\mathbf{x}}_t), \quad
	\sigma_2 := \sigma^\theta_{t + \Delta t}(\bar{\tilde{\mathbf{x}}}_{t + \Delta t}), \quad
	J_1 := J^\theta_t(\bar{\mathbf{x}}_t), \quad
	J_2 := J^\theta_{t + \Delta t}(\bar{\tilde{\mathbf{x}}}_{t + \Delta t}).
\end{equation*}

\paragraph{Mean dynamics.}
Applying a first-order Taylor expansion of $\bar{u}^\theta_t$ around $\bar{\mathbf{x}}_t$ (as in \cref{eq:mean_ode_deriv}) to both stages of \cref{eq:heun_update} yields
\begin{equation}
	\label{eq:heun_mean}
	\bar{\mathbf{x}}_{t + \Delta t}
	\;\approx\; \bar{\mathbf{x}}_t + \tfrac{\Delta t}{2} \Big(\bar{u}^\theta_t(\bar{\mathbf{x}}_t) + \bar{u}^\theta_{t + \Delta t}(\bar{\tilde{\mathbf{x}}}_{t + \Delta t})\Big),
\end{equation}
which is exactly the deterministic Heun step applied to the mean state.

\paragraph{Variance decomposition.}
Applying the element-wise variance identity to $\mathbf{x}_{t + \Delta t} = \mathbf{x}_t + \tfrac{\Delta t}{2}(\mathbf{k}_1 + \mathbf{k}_2)$ gives the exact decomposition
\begin{equation}
	\label{eq:heun_variance_decomposition}
	\begin{aligned}
		\Var[\mathbf{x}_{t + \Delta t}]
		\;=\; &\Var[\mathbf{x}_t] + \tfrac{(\Delta t)^2}{4} \Big(\Var[\mathbf{k}_1] + \Var[\mathbf{k}_2] + 2\, \Cov(\mathbf{k}_1, \mathbf{k}_2)\Big) \\
		&+ \Delta t \, \Big(\Cov(\mathbf{x}_t, \mathbf{k}_1) + \Cov(\mathbf{x}_t, \mathbf{k}_2)\Big).
	\end{aligned}
\end{equation}
Compared with the Euler rule (\cref{eq:variance_propagation}), \cref{eq:heun_variance_decomposition} introduces two additional variance terms ($\Var[\mathbf{k}_2]$ and $\Cov(\mathbf{k}_1, \mathbf{k}_2)$) and one additional state--velocity coupling term ($\Cov(\mathbf{x}_t, \mathbf{k}_2)$), all arising from the corrector stage.
We now derive tractable approximations for each term.

\paragraph{Stage variances.}
\emph{$\Var[\mathbf{k}_1]$.} By the law of total variance and a first-order Taylor expansion of $\sigma^\theta_t$ around $\bar{\mathbf{x}}_t$ (identical to \cref{eq:variance_approx}), and neglecting $\Var(\bar{u}^\theta_t(\mathbf{x}_t))$,
\begin{equation}
	\label{eq:heun_var_k1}
	\Var[\mathbf{k}_1] \;\approx\; \sigma_1^2.
\end{equation}

\emph{$\Var[\mathbf{k}_2]$.} Applying the law of total variance to $\mathbf{k}_2 \mid \tilde{\mathbf{x}}_{t + \Delta t}$ yields
\begin{equation*}
	\Var[\mathbf{k}_2]
	\;=\; \E_{\tilde{\mathbf{x}}}\big[\sigma^\theta_{t + \Delta t}(\tilde{\mathbf{x}})^2\big] + \Var\big(\bar{u}^\theta_{t + \Delta t}(\tilde{\mathbf{x}})\big).
\end{equation*}
A first-order Taylor expansion of $\sigma^\theta_{t + \Delta t}$ around $\bar{\tilde{\mathbf{x}}}_{t + \Delta t}$, combined with the same neglect-of-Var-of-mean step, gives
\begin{equation}
	\label{eq:heun_var_k2}
	\Var[\mathbf{k}_2] \;\approx\; \sigma_2^2.
\end{equation}
The only change relative to \cref{eq:heun_var_k1} is that $\sigma_2^2$ is evaluated at the Euler-predicted mean $\bar{\tilde{\mathbf{x}}}_{t + \Delta t}$ and at time $t + \Delta t$.

\paragraph{State--velocity covariances.}
\emph{$\Cov(\mathbf{x}_t, \mathbf{k}_1)$.} This term is identical to the Euler case in \cref{eq:cov_approx_taylor,eq:cov_approx_jvp}: it approximates $\diag(J_1) \odot \Var[\mathbf{x}_t]$ via Hutchinson's estimator.
With $\boldsymbol{\sigma}^x_t := \sqrt{\Var[\mathbf{x}_t]}$ and $S$ independent Rademacher probes $\mathbf{r}_i \in \mathbb{R}^n$ whose entries are i.i.d. uniform on $\{-1, +1\}$, defining the scaled probes $\mathbf{v}_i := \boldsymbol{\sigma}^x_t \odot \mathbf{r}_i$,
\begin{equation}
	\label{eq:heun_cov_x_k1}
	\Cov(\mathbf{x}_t, \mathbf{k}_1) \;\approx\; \frac{1}{S}\sum_{i=1}^S \mathbf{v}_i \odot (J_1 \mathbf{v}_i).
\end{equation}

\emph{$\Cov(\mathbf{x}_t, \mathbf{k}_2)$.}
Expand $\bar{u}^\theta_{t + \Delta t}$ around $\bar{\tilde{\mathbf{x}}}_{t + \Delta t}$ and write the predictor perturbation as
\begin{equation}
	\label{eq:heun_predictor_perturbation}
	\tilde{\mathbf{x}}_{t + \Delta t} - \bar{\tilde{\mathbf{x}}}_{t + \Delta t}
	\;\approx\; (I + J_1\, \Delta t)(\mathbf{x}_t - \bar{\mathbf{x}}_t) + \sigma_1 \odot \epsilon_1\, \Delta t.
\end{equation}
Since $\epsilon_1, \epsilon_2$ are zero-mean and independent of $\mathbf{x}_t$, the noise terms do not contribute to $\Cov(\mathbf{x}_t, \mathbf{k}_2)$.
Applying the same diagonal-$\Cov(\mathbf{x}_t)$ approximation as in \cref{eq:cov_to_diag_jacobian1,eq:cov_to_diag_jacobian2},
\begin{equation}
	\label{eq:heun_cov_x_k2}
	\Cov(\mathbf{x}_t, \mathbf{k}_2)
	\;\approx\; \diag\!\big(J_2(I + J_1\, \Delta t)\big) \odot \Var[\mathbf{x}_t].
\end{equation}

\emph{$\Cov(\mathbf{k}_1, \mathbf{k}_2)$.}
Using the first-order expansions
\begin{equation}
	\label{eq:heun_k1_expansion}
	\mathbf{k}_1 - \bar{\mathbf{k}}_1 \;\approx\; J_1(\mathbf{x}_t - \bar{\mathbf{x}}_t) + \sigma_1 \odot \epsilon_1,
\end{equation}
\begin{equation}
	\label{eq:heun_k2_expansion}
	\mathbf{k}_2 - \bar{\mathbf{k}}_2 \;\approx\; J_2\big((I + J_1\, \Delta t)(\mathbf{x}_t - \bar{\mathbf{x}}_t) + \sigma_1 \odot \epsilon_1\, \Delta t\big) + \sigma_2 \odot \epsilon_2,
\end{equation}
and the mutual independence of $\mathbf{x}_t, \epsilon_1, \epsilon_2$, the element-wise cross-covariance at coordinate $i$ splits into a state-driven and a noise-driven part.
Under the diagonal-$\Cov(\mathbf{x}_t)$ approximation, a direct computation (cf. \cref{eq:cov_to_diag_jacobian1}) gives
\begin{equation}
	\label{eq:heun_cov_k1_k2}
	\Cov(\mathbf{k}_1, \mathbf{k}_2)_i
	\;\approx\; \underbrace{\sum_{j = 1}^n (J_1)_{ij} \big(J_2(I + J_1\, \Delta t)\big)_{ij}\, \Var[\mathbf{x}_t]_j}_{\text{state-driven}}
	\;+\; \underbrace{\Delta t\, (J_2)_{ii}\, (\sigma_1)_i^2}_{\text{noise-driven}},
\end{equation}
or equivalently in matrix form,
\begin{equation}
	\label{eq:heun_cov_k1_k2_matrix}
	\Cov(\mathbf{k}_1, \mathbf{k}_2)
	\;\approx\; \diag\!\big(J_1\, \diag(\Var[\mathbf{x}_t])\, (J_2(I + J_1\, \Delta t))^\top\big)
	\;+\; \Delta t\, \diag(J_2) \odot \sigma_1^2.
\end{equation}
Unlike \cref{eq:heun_cov_x_k2}, the state-driven part of \cref{eq:heun_cov_k1_k2} is \emph{not} of the form $\diag(M) \odot \Var[\mathbf{x}_t]$ for a single matrix $M$, because the two velocity stages are linearly mapped by different Jacobians. A Hutchinson estimator that respects this structure is derived below.

\paragraph{Hutchinson--JVP estimators.}
\label{app:heun_variance_propagation:hutchinson}
Explicitly forming $J_1$ or $J_2$ is intractable in high dimensions.
We therefore estimate the diagonal Jacobian products appearing in \cref{eq:heun_cov_x_k1,eq:heun_cov_x_k2,eq:heun_cov_k1_k2_matrix} via Rademacher probes and Jacobian--vector products, following the same strategy as \cref{eq:cov_approx_jvp}.

\emph{Setup.}
Let $\boldsymbol{\sigma}^x_t := \sqrt{\Var[\mathbf{x}_t]}$, and draw two independent Rademacher vectors $\mathbf{r}, \mathbf{r}' \in \mathbb{R}^n$ whose entries are i.i.d. uniform on $\{-1, +1\}$, so that
\begin{equation}
	\label{eq:rademacher_identity}
	\E[r_i r_j] \;=\; \E[r'_i r'_j] \;=\; \delta_{ij}, \qquad \E[r_i r'_j] \;=\; 0.
\end{equation}
Define the scaled probes
\begin{equation}
	\label{eq:rademacher_scaled}
	\mathbf{v} \;:=\; \boldsymbol{\sigma}^x_t \odot \mathbf{r}, \qquad \mathbf{w} \;:=\; \sigma_1 \odot \mathbf{r}',
\end{equation}
which satisfy $\E[v_i v_j] = \Var[\mathbf{x}_t]_i\, \delta_{ij}$ and $\E[w_i w_j] = (\sigma_1)_i^2\, \delta_{ij}$.

\emph{Derivation of the $\Cov(\mathbf{x}_t, \mathbf{k}_2)$ estimator.}
Let $M := J_2(I + J_1\, \Delta t)$. At coordinate $i$,
\begin{equation}
	\label{eq:heun_x_k2_hutch_deriv}
	\E\big[v_i\, (M\mathbf{v})_i\big]
	\;=\; \sum_{j = 1}^n M_{ij}\, \E[v_i v_j]
	\;=\; M_{ii}\, \Var[\mathbf{x}_t]_i.
\end{equation}
Stacking all coordinates yields
\begin{equation}
	\label{eq:heun_x_k2_hutch_expectation}
	\E_\mathbf{r}\!\big[\mathbf{v} \odot (M \mathbf{v})\big]
	\;=\; \diag(M) \odot \Var[\mathbf{x}_t]
	\;=\; \diag\!\big(J_2(I + J_1\, \Delta t)\big) \odot \Var[\mathbf{x}_t],
\end{equation}
which matches \cref{eq:heun_cov_x_k2}. The product $M\mathbf{v} = J_2(\mathbf{v} + J_1\, \mathbf{v}\, \Delta t)$ is computed via two Jacobian--vector products, avoiding explicit formation of $J_1, J_2$, or $M$. Using $S$ i.i.d. probes $\{\mathbf{r}_i\}$ gives the unbiased Monte Carlo estimator
\begin{equation}
	\label{eq:heun_cov_x_k2_jvp}
	\Cov(\mathbf{x}_t, \mathbf{k}_2)
	\;\approx\; \tfrac{1}{S}\sum_{i=1}^S \mathbf{v}_i \odot \Big(J_2(\mathbf{v}_i + J_1\, \mathbf{v}_i\, \Delta t)\Big).
\end{equation}
Setting $J_2 = I$ and $\Delta t = 0$ recovers the Euler-case estimator \cref{eq:cov_approx_jvp} for $\Cov(\mathbf{x}_t, \mathbf{k}_1)$,
\begin{equation}
	\label{eq:heun_cov_x_k1_jvp}
	\Cov(\mathbf{x}_t, \mathbf{k}_1) \;\approx\; \tfrac{1}{S}\sum_{i=1}^S \mathbf{v}_i \odot (J_1 \mathbf{v}_i).
\end{equation}

\emph{Derivation of the $\Cov(\mathbf{k}_1, \mathbf{k}_2)$ estimator.}
We treat the state-driven and noise-driven parts of \cref{eq:heun_cov_k1_k2} separately.

\emph{State-driven part.} Let $A := J_1$ and $B := J_2(I + J_1\, \Delta t)$. Using \cref{eq:rademacher_identity},
\begin{equation}
	\label{eq:heun_k1_k2_state_hutch_deriv}
	\E\big[(A\mathbf{v})_i\, (B\mathbf{v})_i\big]
	\;=\; \sum_{j, k = 1}^n A_{ij}\, B_{ik}\, \E[v_j v_k]
	\;=\; \sum_{j=1}^n A_{ij}\, B_{ij}\, \Var[\mathbf{x}_t]_j,
\end{equation}
which coincides with the state-driven part of \cref{eq:heun_cov_k1_k2} coordinate by coordinate.
Thus, stacking over $i$,
\begin{equation}
	\label{eq:heun_k1_k2_state_hutch_expectation}
	\E_\mathbf{r}\!\big[(J_1\mathbf{v}) \odot (J_2(I + J_1\, \Delta t)\mathbf{v})\big]
	\;=\; \diag\!\big(J_1\, \diag(\Var[\mathbf{x}_t])\, (J_2(I + J_1\, \Delta t))^\top\big).
\end{equation}

\emph{Noise-driven part.} Using $\E[w_i w_j] = (\sigma_1)_i^2\, \delta_{ij}$,
\begin{equation}
	\label{eq:heun_k1_k2_noise_hutch_deriv}
	\E\big[w_i\, (J_2 \mathbf{w})_i\big]
	\;=\; \sum_{j=1}^n (J_2)_{ij}\, \E[w_i w_j]
	\;=\; (J_2)_{ii}\, (\sigma_1)_i^2.
\end{equation}
Stacking over $i$ gives
\begin{equation}
	\label{eq:heun_k1_k2_noise_hutch_expectation}
	\E_{\mathbf{r}'}\!\big[\mathbf{w} \odot (J_2 \mathbf{w})\big] \;=\; \diag(J_2) \odot \sigma_1^2,
\end{equation}
which matches the noise-driven term of \cref{eq:heun_cov_k1_k2}.

\emph{Combined estimator.} Drawing $S$ independent probe pairs $\{(\mathbf{r}_i, \mathbf{r}'_i)\}_{i=1}^S$ and combining \cref{eq:heun_k1_k2_state_hutch_expectation,eq:heun_k1_k2_noise_hutch_expectation} yields the unbiased Monte Carlo estimator
\begin{equation}
	\label{eq:heun_cov_k1_k2_jvp}
	\Cov(\mathbf{k}_1, \mathbf{k}_2) \;\approx\; \tfrac{1}{S}\sum_{i=1}^S (J_1 \mathbf{v}_i) \odot \Big(J_2(\mathbf{v}_i + J_1\, \mathbf{v}_i\, \Delta t)\Big)
	\;+\; \tfrac{\Delta t}{S}\sum_{i=1}^S \mathbf{w}_i \odot (J_2 \mathbf{w}_i).
\end{equation}

\emph{Computational cost and probe sharing.}
The three estimators \cref{eq:heun_cov_x_k1_jvp,eq:heun_cov_x_k2_jvp,eq:heun_cov_k1_k2_jvp} can share the probes $\{\mathbf{r}_i\}$ and cache the intermediate products $\{J_1 \mathbf{v}_i,\, \mathbf{v}_i, + \Delta t J_1 \mathbf{v}_i \}$ across terms.
Per sampling step, the total overhead is one JVPs with $J_1$ (for $J_1 \mathbf{v}_i$) and two JVPs with $J_2$ (for $J_2 (\mathbf{v}_i, + \Delta t J_1 \mathbf{v}_i)$, and $J_2 \mathbf{w}_i$).

\paragraph{Final Heun2 variance propagation rule.}
Substituting \cref{eq:heun_var_k1,eq:heun_var_k2,eq:heun_cov_x_k1_jvp,eq:heun_cov_x_k2_jvp,eq:heun_cov_k1_k2_jvp} into \cref{eq:heun_variance_decomposition} yields the Heun2 variance update directly in Hutchinson/JVP form:
\begin{equation}
	\label{eq:heun_variance_propagation}
	\begin{aligned}
		\Var[\mathbf{x}_{t + \Delta t}]
		\;\approx\; & \Var[\mathbf{x}_t]
		\;+\; \tfrac{(\Delta t)^2}{4}\big(\sigma_1^2 + \sigma_2^2\big) \\
		&+ \tfrac{\Delta t}{S}\sum_{i=1}^S \mathbf{v}_i \odot (J_1 \mathbf{v}_i) \\
		&+ \tfrac{\Delta t}{S}\sum_{i=1}^S \mathbf{v}_i \odot \Big(J_2(\mathbf{v}_i + \Delta t\, J_1 \mathbf{v}_i)\Big) \\
		&+ \tfrac{(\Delta t)^2}{2S}\sum_{i=1}^S (J_1 \mathbf{v}_i) \odot \Big(J_2(\mathbf{v}_i + \Delta t\, J_1 \mathbf{v}_i)\Big) \\
		&+ \tfrac{(\Delta t)^3}{2S}\sum_{i=1}^S \mathbf{w}_i \odot (J_2 \mathbf{w}_i).
	\end{aligned}
\end{equation}
The first line is the trapezoidal injected-noise term, while the remaining three probe averages estimate $\Delta t\, \Cov(\mathbf{x}_t, \mathbf{k}_1)$, $\Delta t\, \Cov(\mathbf{x}_t, \mathbf{k}_2)$, and $\tfrac{(\Delta t)^2}{2}\Cov(\mathbf{k}_1, \mathbf{k}_2)$, respectively.
Expanding the compact JVP products recover the matrix-form expressions in \cref{eq:heun_cov_x_k2,eq:heun_cov_k1_k2_matrix}, including the $\mathcal{O}(\Delta t)$, $\mathcal{O}(\Delta t^2)$, and $\mathcal{O}(\Delta t^3)$ contributions discussed above.

\section{Derivations of $\lambda^*$}
\label{app:cfg_lambda}

Recall that UCFG chooses a scalar CFG scale $\lambda \ge 0$ by minimizing the total predicted variance of the
extrapolated velocity. Using the notation in \cref{subsec:uncertainty_aware_guidance}, let
$\sigma^\theta_{t,y}(\bar{\mathbf{x}}_t) \in \mathbb{R}^n$ and
$\sigma^\theta_{t,\varnothing}(\bar{\mathbf{x}}_t) \in \mathbb{R}^n$
denote the element-wise standard deviations of the conditional and unconditional velocities, respectively.
As described in \cref{eq:cfg_var_approx}, the element-wise standard deviation of the extrapolated
velocity is
\begin{equation}
	\tilde{\sigma}(\lambda)
	= (1+\lambda)\sigma^\theta_{t,y}(\bar{\mathbf{x}}_t) - \lambda\sigma^\theta_{t,\varnothing}(\bar{\mathbf{x}}_t)
	= \sigma^\theta_{t,y}(\bar{\mathbf{x}}_t) + \lambda(\sigma^\theta_{t,y}(\bar{\mathbf{x}}_t)-\sigma^\theta_{t,\varnothing}(\bar{\mathbf{x}}_t)).
\end{equation}
Therefore, \cref{eq:lambda_opt_argmin} can be written as the following least-squares problem:
\begin{equation}
	\lambda_{\mathrm{opt}}
	= \arg\min_{\lambda \ge 0} \sum_{i=1}^n \tilde{\sigma}_i(\lambda)^2
	= \arg\min_{\lambda \ge 0} \Vert \sigma^\theta_{t,y}(\bar{\mathbf{x}}_t) + \lambda \mathbf{d} \Vert_2^2,
	\qquad
	\mathbf{d} := \sigma^\theta_{t,y}(\bar{\mathbf{x}}_t) - \sigma^\theta_{t,\varnothing}(\bar{\mathbf{x}}_t).
\end{equation}
Expanding the objective yields a convex quadratic in $\lambda$:
\begin{equation}
	J(\lambda)
	= \Vert \sigma^\theta_{t,y}(\bar{\mathbf{x}}_t) + \lambda \mathbf{d} \Vert_2^2
	= \sigma^\theta_{t,y}(\bar{\mathbf{x}}_t)^\top \sigma^\theta_{t,y}(\bar{\mathbf{x}}_t)
	+ 2\lambda\, \mathbf{d}^\top \sigma^\theta_{t,y}(\bar{\mathbf{x}}_t)
	+ \lambda^2 \mathbf{d}^\top \mathbf{d}.
\end{equation}
When $\mathbf{d}^\top \mathbf{d} > 0$, the unconstrained minimizer is obtained by setting $\frac{dJ}{d\lambda}=0$:
\begin{equation}
	\frac{dJ}{d\lambda}
	= 2\,\mathbf{d}^\top \sigma^\theta_{t,y}(\bar{\mathbf{x}}_t)
	+ 2\lambda\, \mathbf{d}^\top \mathbf{d}
	= 0
	\quad\Longrightarrow\quad
	\lambda^\star
	= - \frac{\mathbf{d}^\top \sigma^\theta_{t,y}(\bar{\mathbf{x}}_t)}{\mathbf{d}^\top \mathbf{d}}
	= \frac{(\sigma^\theta_{t,\varnothing})^\top \sigma^\theta_{t,y} - \Vert \sigma^\theta_{t,y} \Vert_2^2}{\Vert \sigma^\theta_{t,y} - \sigma^\theta_{t,\varnothing} \Vert_2^2}.
\end{equation}
Imposing the constraint $\lambda \ge 0$ gives the closed-form solution:
\begin{equation}
	\lambda_{\mathrm{opt}} = \max(0, \lambda^\star).
\end{equation}
If $\mathbf{d}^\top \mathbf{d} = 0$ (i.e., $\boldsymbol{\sigma}_{y} = \boldsymbol{\sigma}_{\varnothing}$),
the objective $J(\lambda)$ is constant in $\lambda$, and we set $\lambda_{\mathrm{opt}} = 0$.
Finally, we apply the clamp in \cref{eq:lambda_clamp}:
\begin{equation*}
	\lambda^{\ast} = \min(\lambda_{\mathrm{opt}}, \lambda_{\max}).
\end{equation*}

\section{Implementation Details}
\label{app:implementation_details}

\subsection{Model \& Training}

\begin{table}[h]
	\centering
	\caption{\textbf{Hyper-parameters.}
		When two learning rates are reported as $(\eta_{\text{head}}, \eta_{\text{backbone}})$, they denote the head and backbone learning rates, respectively.}
	\label{tab:dit_hyperparams}
	\begin{tabular}{l||cc|cc|c}
		\toprule
		& \multicolumn{2}{c|}{CIFAR-10} & \multicolumn{2}{c|}{ImageNet-128}  & ImageNet-256 \\
		& pretraining & finetuning & pretraining & finetuning & finetuning \\
		\midrule
		Learning rate    & $1e\text{-}4$ & $(1e\text{-}5, 1e\text{-}6)$ & $2e\text{-}4$ & $ (2e\text{-}5, 2e\text{-}5)$ & $(1e\text{-}5, 1e\text{-}5)$ \\
		AdamW $(\beta_1, \beta_2)$ 
		& $(0.9, 0.95)$ & $(0.9, 0.999)$ & $(0.9, 0.999)$ & $(0.9, 0.999)$ & $(0.9, 0.999)$ \\
		Gradient warm-up step & -- & 10000 & -- & 1000 & 1000 \\
		Gradient clipping & -- & 1.0 & -- & 1.0 & 1.0 \\
		EMA decay rate   & 0.999 & 0.999 & 0.9999 & 0.9999 & 0.9999 \\
		Batch size       & 128 & 128 & 1024 & 1024 & 512\\
		Epochs           & 1000 & 100 & 900 & 100 & 90 \\
		GPUs             & 2 & 2 & 4 & 4 & 4 \\
		\bottomrule
	\end{tabular}
\end{table}

\begin{table}[t]
	\centering
	\setlength{\tabcolsep}{3.5pt}
	\caption{ADM configuration for CIFAR-10.}
	\label{tab:adm_cifar10}
	\begin{tabular}{@{}lc@{}}
		\toprule
		& CIFAR-10 \\
		\midrule
		\# of ResNet blocks per scale & 4 \\
		Base channels & 128 \\
		Channel multiplier per scale & (2, 2, 2) \\
		Attention resolutions & 2 \\
		Dropout & 0.3 \\
		\bottomrule
	\end{tabular}
\end{table}

\paragraph{Architectures.}
For CIFAR-10, we use an ADM-based unconditional flow matching model.
The architectural configuration follows standard ADM design choices and is detailed in \cref{tab:adm_cifar10}.
For ImageNet-128 and ImageNet-256, we adopt DiT-based conditional latent flow matching models, as described in \cref{sec:experiments}, using DiT-B/2 as the backbone and a pretrained autoencoder to map RGB images to a latent space.

\paragraph{Training procedure.}
We summarize the training hyper-parameters for all datasets in \cref{tab:dit_hyperparams}. All models are trained with EMA, using the decay rates reported in the table.
For CIFAR-10, we employ a skewed time-step sampling strategy used in EDM~\citep{Karras22EDM} during training.
For ImageNet-256, we train the model using reverse-time parameterization, consistent with the pretrained latent flow matching setup.

For datasets involving fine-tuning, we apply linear gradient warm-up for the specified number of steps and use gradient clipping with a maximum norm of 1.0.
Batch sizes and the number of GPUs used for each setting are reported in \cref{tab:dit_hyperparams}.
For the reweighted mini-batch estimator used to compute $\hat{u}_t$, the batch size is defined per GPU and aggregated across devices during training.

\paragraph{Hardware.}
All training, fine-tuning, and TFLOPs-based compute measurements are run on NVIDIA RTX 6000 Ada GPUs (48\,GB memory); the GPU counts reported in \cref{tab:dit_hyperparams} refer to this hardware.
A subset of inference runs (sampling and evaluation that does not contribute to the TFLOPs measurements) is additionally executed on NVIDIA RTX PRO 6000 Blackwell GPU.

\paragraph{Uncertainty modeling.}
For uncertainty-aware training, we add a variance prediction head to the pretrained flow matching backbone.
The variance head predicts $\log \sigma$ to improve numerical stability during training.
We use the $\beta$-NLL~\citep{Seitzer22Betanll} objective with $\beta = 1.0$ for all experiments.

For ImageNet-128 and ImageNet-256, classifier-free conditioning is enabled by applying label dropout with probability 0.1 during training.

\subsection{Flow Matching Versions of Baselines}
\label{app:subsec:flow_baselines}

Existing uncertainty quantification methods for sampling-based generative models are primarily developed for diffusion models.
To enable a fair comparison with UA-Flow, we adapt these baselines~\citep{DeVita25Aleatoric, Kou23Bayesdiff, Jazbec25Genunc} to the flow matching framework.
In this subsection, we describe how the diffusion-based formulations of these methods are converted into their flow matching counterparts.

\paragraph{Aleatoric Uncertainty (AU)~\citep{DeVita25Aleatoric}.}
In the original formulation, Aleatoric Uncertainty (AU) estimates sample uncertainty using the score function of a diffusion model.
At each sampling step $t$, the model first estimates the clean data sample $\hat{\mathbf{x}}_1$ from the predicted score $\epsilon_t^\theta(\mathbf{x}_t)$.
Multiple perturbed states $\hat{\mathbf{x}}_t^{\,i}$ are then generated by re-noising the estimated data, and the variance of the score predictions $\epsilon_t^\theta(\hat{\mathbf{x}}_t^{\,i})$ is used as a measure of sample uncertainty.

To adapt AU to flow matching, we replace the score function with the learned velocity field $u_t^\theta(\mathbf{x}_t)$.
Assuming an affine probability path, the clean data estimate $\hat{\mathbf{x}}_1$ can be recovered from the current state $\mathbf{x}_t$ and the predicted velocity as
\begin{equation}
	\hat{\mathbf{x}}_1 =
	\frac{1}{\dot{\alpha}_t \beta_t - \dot{\beta}_t \alpha_t}
	\left(
	-\dot{\beta}_t \mathbf{x}_t + \beta_t u_t^\theta(\mathbf{x}_t)
	\right).
\end{equation}
Starting from the estimated data $\hat{\mathbf{x}}_1$, we generate multiple perturbed states $\hat{\mathbf{x}}_t^{\,i}$ by applying the forward affine transformation.
We then compute the element-wise variance of the velocity predictions $u_t^\theta(\hat{\mathbf{x}}_t^{\,i})$, which serves as the aleatoric uncertainty estimate at time $t$.
We aggregate the element-wise velocity variances over the late-stage sampling steps to obtain a sample uncertainty map, which is reduced to a scalar score for filtering as described in \cref{subsec:filtering_exp}.

\paragraph{BayesDiff~\citep{Kou23Bayesdiff}.}
In the diffusion-based formulation, BayesDiff estimates the uncertainty of the score function at each sampling step using Laplace Last Layer Approximation (LLLA)~\citep{Daxberger21Laplace}.
The estimated score variance is then propagated through the diffusion dynamics to obtain uncertainty estimates for the generated samples.

To adapt BayesDiff to flow matching, we apply LLLA to the velocity field and estimate the variance of the predicted velocity at each time step.
This velocity variance is subsequently propagated through the flow dynamics following the same uncertainty propagation scheme as in the original BayesDiff formulation.

\paragraph{Generative Uncertainty (GenUnc)~\citep{Jazbec25Genunc}.}
In the diffusion-based formulation, Generative Uncertainty (GenUnc) estimates sample uncertainty by sampling multiple model weights through LLLA and generating multiple images from the same noise realization.
The resulting images are embedded into the CLIP~\citep{Radford21CLIP} feature space, and the entropy of the extracted features is used as a sample-level uncertainty score.

For flow matching, we follow the same procedure by sampling model weights, generating multiple samples from the same initial noise using the corresponding flow matching model, and computing the variance of the resulting CLIP features as the sample-level uncertainty score.

\subsection{APG and APG+UCFG for Flow Matching}
\label{app:subsec:apg}

We adapt adaptive projected guidance (APG)~\citep{Sadat24APG} to flow matching by operating directly on the conditional and unconditional mean velocities.
Using the notation in \cref{subsec:uncertainty_aware_guidance}, let
$\bar{u}^\theta_{t,y}(\bar{\mathbf{x}}_t)$ and
$\bar{u}^\theta_{t,\varnothing}(\bar{\mathbf{x}}_t)$ denote the conditional and unconditional mean velocities, respectively.
For brevity, we suppress their dependence on $\bar{\mathbf{x}}_t$ and define the CFG direction
\begin{equation}
	g_t
	=
	\bar{u}^\theta_{t,y}-\bar{u}^\theta_{t,\varnothing}.
\end{equation}

\paragraph{APG.}
Following APG, we first apply reverse momentum:
\begin{equation}
	m_t = g_t + \beta m_{t-1},
\end{equation}
where $\beta=-0.25$ in our experiments.
We then clip the norm of the momentum direction independently for each sample:
\begin{equation}
	\widetilde{g}_t
	=
	\min\left(1,\frac{r}{\Vert m_t\Vert_2+\epsilon}\right)m_t,
\end{equation}
where $r=2.5$ and $\epsilon=10^{-6}$.
The clipped direction is decomposed into components parallel and orthogonal to the conditional velocity:
\begin{equation}
	p_t
	=
	\frac{\widetilde{g}_t^\top\bar{u}^\theta_{t,y}}
	{\Vert\bar{u}^\theta_{t,y}\Vert_2^2+\epsilon},
	\qquad
	\widetilde{g}_{t,\parallel}
	=
	p_t\bar{u}^\theta_{t,y},
	\qquad
	\widetilde{g}_{t,\perp}
	=
	\widetilde{g}_t-\widetilde{g}_{t,\parallel}.
\end{equation}
The APG-guided mean velocity is
\begin{equation}
	\bar{u}^\theta_{t,\mathrm{APG}}
	=
	\bar{u}^\theta_{t,y}
	+
	\lambda
	\left(
	\widetilde{g}_{t,\perp}
	+
	\eta\widetilde{g}_{t,\parallel}
	\right).
\end{equation}
We set $\eta=0$, thereby removing the parallel component entirely.
APG modifies only the mean velocity and retains the conditional branch's predicted standard deviation.

\paragraph{APG+UCFG.}
APG+UCFG replaces the fixed scale $\lambda$ with a sample-wise adaptive scale selected to minimize the predicted uncertainty of the guided velocity.
Unlike APG, APG+UCFG does not use reverse momentum.
Instead, it clips the raw CFG direction using
\begin{equation}
	c_t
	=
	\min\left(1,\frac{r}{\Vert g_t\Vert_2+\epsilon}\right)
\end{equation}
and removes its component parallel to the conditional velocity.
Let
\begin{equation}
	q_t
	=
	\frac{(c_tg_t)^\top\bar{u}^\theta_{t,y}}
	{\Vert\bar{u}^\theta_{t,y}\Vert_2^2+\epsilon}.
\end{equation}
The resulting mean-guidance direction can be written as
\begin{equation}
	d_t^u
	=
	c_tg_t-q_t\bar{u}^\theta_{t,y}
	=
	(c_t-q_t)\bar{u}^\theta_{t,y}
	-
	c_t\bar{u}^\theta_{t,\varnothing}.
\end{equation}

Let $\sigma^\theta_{t,y}(\bar{\mathbf{x}}_t)$ and
$\sigma^\theta_{t,\varnothing}(\bar{\mathbf{x}}_t)$ denote the element-wise standard deviations of the conditional and unconditional velocities, respectively.
Applying the same clipping and projection coefficients to these standard deviations gives the uncertainty-guidance direction
\begin{equation}
	d_t^\sigma
	=
	(c_t-q_t)\sigma^\theta_{t,y}
	-
	c_t\sigma^\theta_{t,\varnothing}.
\end{equation}
Therefore, the predicted standard deviation under scale $\lambda$ is
\begin{equation}
	\widetilde{\sigma}_{t,\mathrm{APG}}(\lambda)
	=
	\sigma^\theta_{t,y}
	+
	\lambda d_t^\sigma.
\end{equation}
APG+UCFG then selects the sample-wise scale $\lambda^*$ using the same variance-minimization and clamping procedure as UCFG in \cref{app:cfg_lambda}, with $d_t^\sigma$ replacing the standard CFG uncertainty direction and $\lambda_{\max}$ serving as the upper bound.
Finally, APG+UCFG returns the guided mean velocity
$\bar{u}^\theta_{t,y}+\lambda^*d_t^u$ and the standard-deviation surrogate
$\sigma^\theta_{t,y}+\lambda^*d_t^\sigma$.

\section{Additional Results on Main Experiments}
\label{app:additional_results}
This section collects additional figures referenced by the main paper to complement the main experiments.

\subsection{Uncertainty-Based Filtering}
\begin{figure}[t]
	\centering
	\begin{subfigure}[t]{0.8\textwidth}
		\centering
		\includegraphics[width=\textwidth]{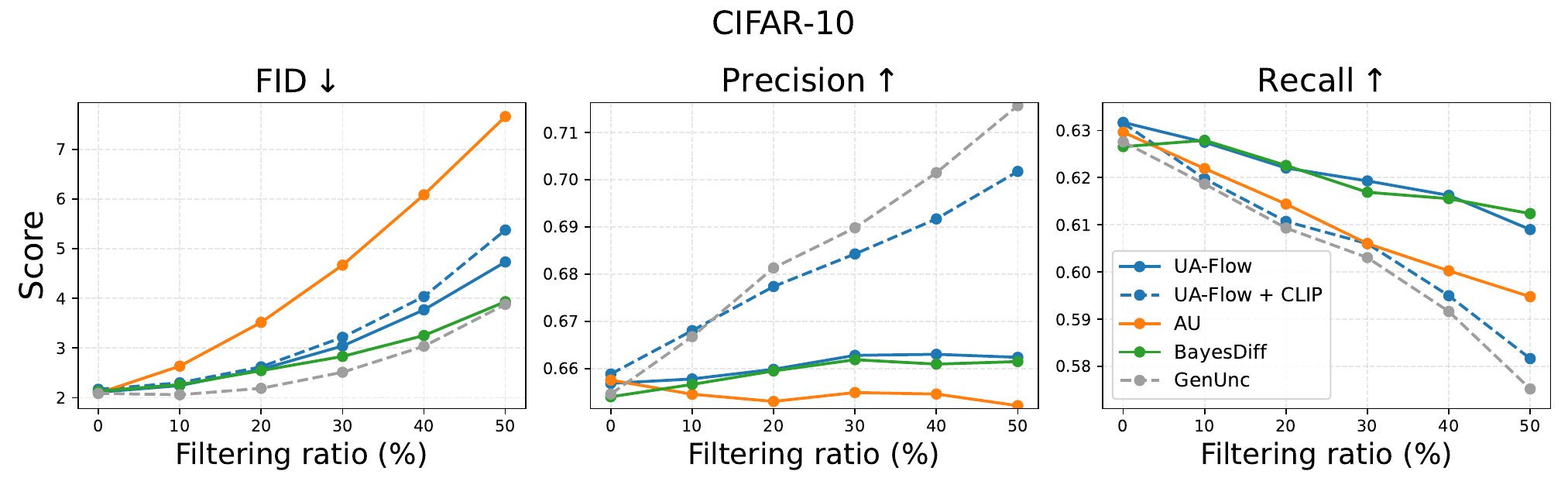}
		\caption{CIFAR-10.}
		\label{fig:filtering_uncertainty_cifar10}
	\end{subfigure}\\[0.8ex]
	\begin{subfigure}[t]{\textwidth}
		\centering
		\includegraphics[width=0.8\textwidth]{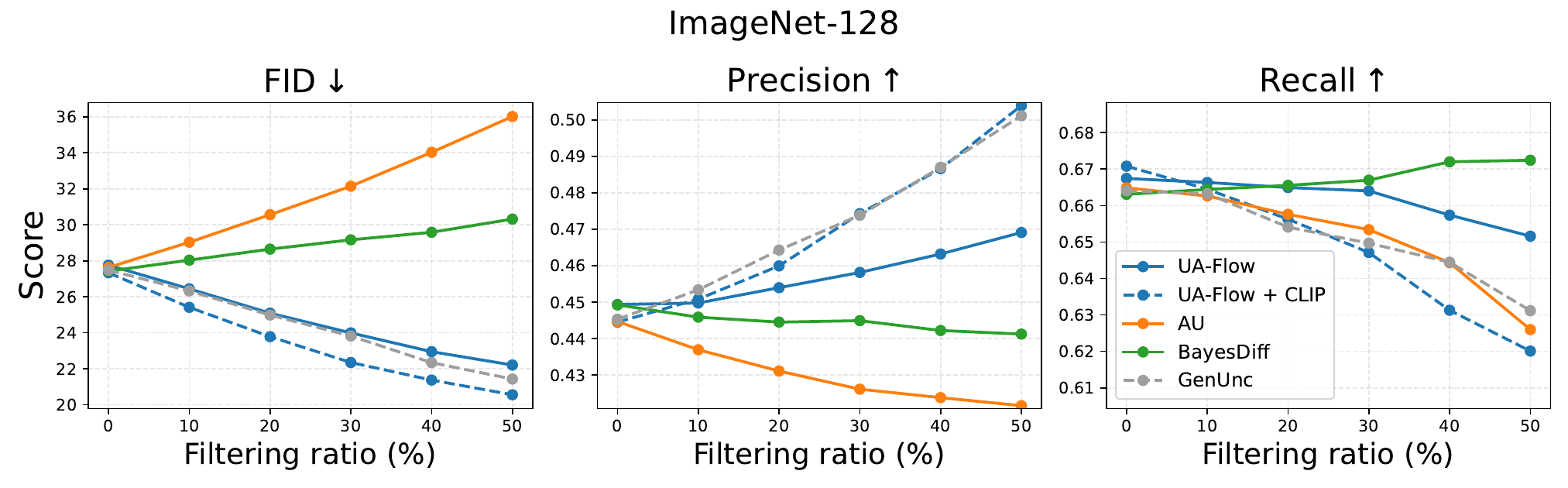}
		\caption{ImageNet-128.}
		\label{fig:filtering_uncertainty_imagenet128}
	\end{subfigure}
	\caption{
		\textbf{Filtering high-uncertainty samples across datasets.}
		Generative quality metrics as a function of the filtering ratio.
		On CIFAR-10, uncertainty-based filtering primarily induces a precision-recall trade-off that can negatively affect FID despite high overall sample quality.
		On ImageNet-128, the trends are consistent with ImageNet-256: UA-Flow achieves lower FID and higher precision after filtering compared to AU and BayesDiff; GenUnc is included as a reference baseline.
	}
	\label{fig:filtering_uncertainty}
\end{figure}

\label{app:additional_results:filtering}
We evaluate whether uncertainty provides a useful reliability signal by progressively filtering out high-uncertainty generated samples and tracking changes in FID and precision/recall.
\Cref{fig:filtering_uncertainty} summarizes the results on CIFAR-10 and ImageNet-128.
Across datasets, UA-Flow exhibits a consistent precision-recall trade-off under filtering: removing the most uncertain samples increases precision while reducing recall.

\paragraph{CIFAR-10.}
On CIFAR-10, the unfiltered sample quality is already high, so the precision gain from filtering is largely offset by the loss in recall, and FID tends to increase as the filtering ratio increases.
We emphasize that this FID trend does not indicate that our uncertainty estimates degrade generation quality: the filtering experiment is designed to validate uncertainty as a \emph{per-sample} reliability signal, not as a measure of the overall quality of the generated set.
The consistent improvement in precision (i.e., fidelity) under filtering confirms that samples flagged as high-uncertainty by UA-Flow are indeed less reliable.

\paragraph{ImageNet-128.}
On ImageNet-128, filtering more reliably improves fidelity: UA-Flow achieves lower FID and higher precision after filtering compared to AU and BayesDiff (with the expected decrease in recall).
We include GenUnc as a reference baseline; it uses a domain-specific scalar uncertainty estimated in a CLIP embedding space, which is different in nature from element-wise uncertainty predicted by UA-Flow.

\subsection{Uncertainty-Aware Guidance}
\label{app:additional_results:guidance}

\subsubsection{Uncertainty-Aware Guidance (UG)}

\begin{figure}[htbp]
	\centering
	\begin{subfigure}[t]{\textwidth}
		\centering
		\includegraphics[width=0.8\textwidth]{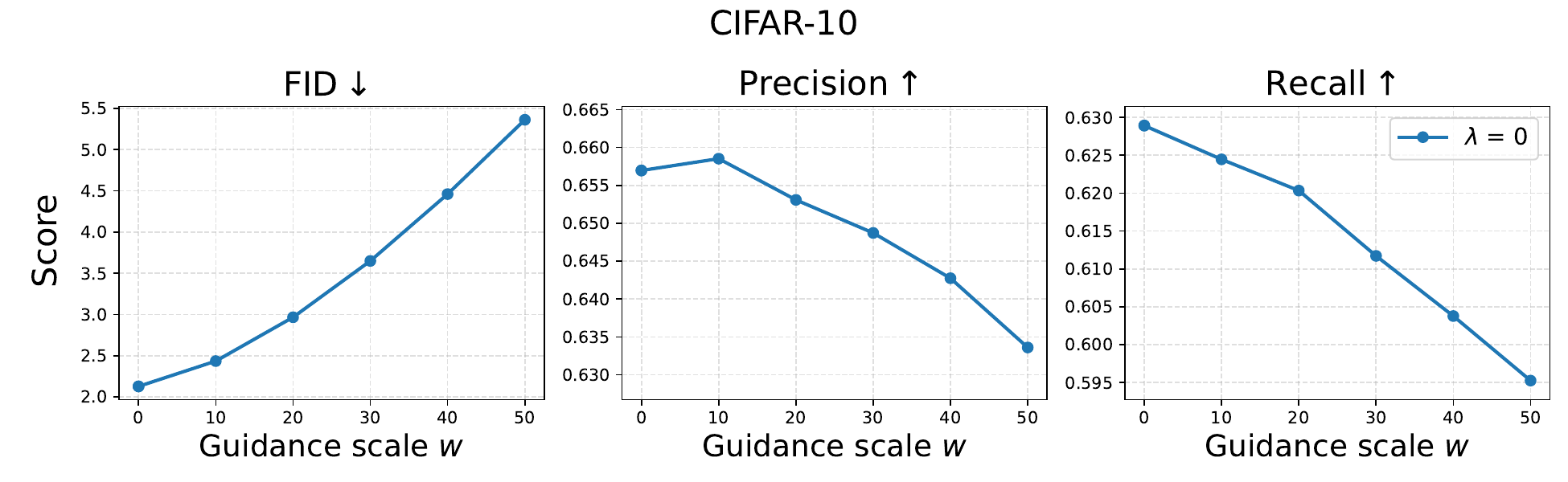}
		\caption{CIFAR-10.}
		\label{fig:guidance_metrics_cifar10}
	\end{subfigure}\\[0.8ex]
	\begin{subfigure}[t]{\textwidth}
		\centering
		\includegraphics[width=0.8\textwidth]{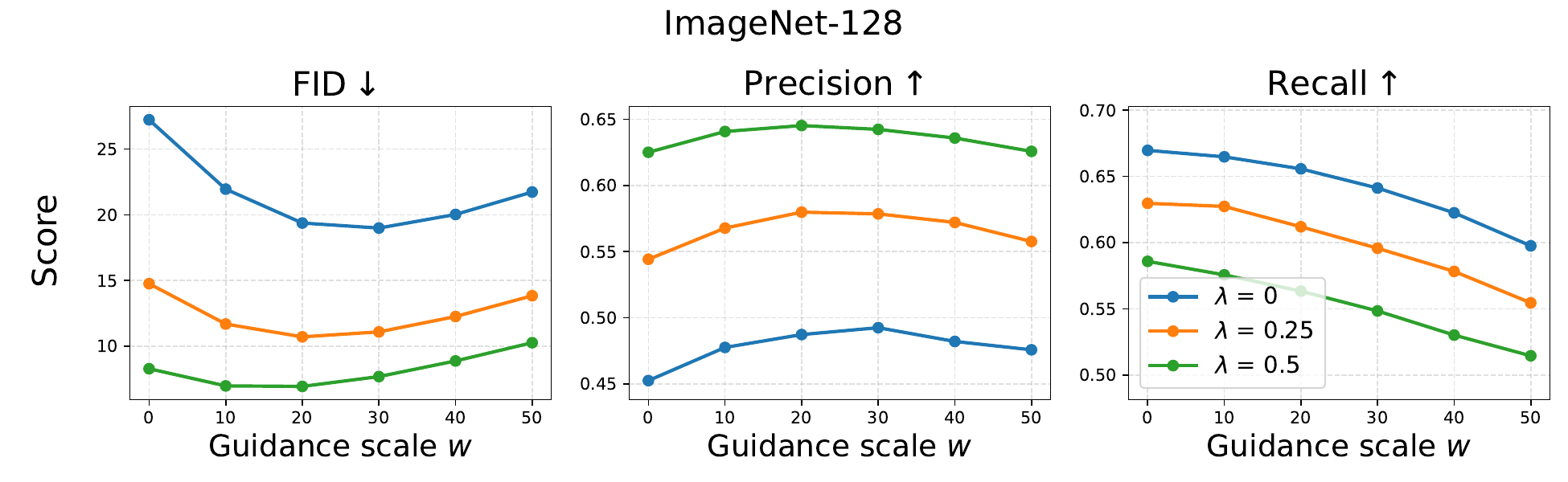}
		\caption{ImageNet-128.}
		\label{fig:guidance_metrics_imagenet128}
	\end{subfigure}
	\caption{
		\textbf{Generation quality metrics as the UG scale $w$ varies on CIFAR-10 and ImageNet-128 under fixed classifier-free guidance (CFG) scales.}
		Increasing $w$ induces a fidelity–diversity trade-off, improving precision while reducing recall.
		The FID response depends on the dataset and the fixed CFG scale.
	}
	\label{fig:guidance_metrics_fixed_cfg}
\end{figure}

\begin{table}[t]
	\centering
	\setlength{\tabcolsep}{3.5pt}
	\caption{
		\textbf{Multi-seed evaluation of UG on ImageNet-256 ($\lambda=0.5$).}
		Mean $\pm$ standard deviation over 3 random seeds.
		This corresponds to the $\lambda=0.5$ rows of \cref{tab:cg_all}(c); we report multi-seed statistics here because the FID gap at this operating point is small, and we wish to confirm that the improvement is statistically consistent.
	}
	\label{tab:ucg_multiseed}
	\begin{tabular}{@{}cccc@{}}
		\toprule
		Setting & FID$\downarrow$ & Prec.$\uparrow$ & Rec.$\uparrow$ \\
		\midrule
		CFG only ($w\!=\!0$) & $5.337_{\pm.042}$ & $\mathbf{0.7132_{\pm.0019}}$ & $\mathbf{0.5495_{\pm.0024}}$ \\
		CFG + UG ($w\!=\!20$) & $\mathbf{4.999_{\pm.047}}$ & $0.7281_{\pm.0005}$ & $0.5393_{\pm.0012}$ \\
		\bottomrule
	\end{tabular}
\end{table}

We sweep the UG scale $w$ under fixed classifier-free guidance (CFG) scales.
As shown in \cref{fig:guidance_metrics_fixed_cfg,fig:guidance_metrics_imagenet256}, increasing $w$ induces a consistent precision-recall trade-off across datasets: precision typically increases or peaks at an intermediate $w$, while recall decreases as guidance becomes stronger.
As a result, FID improves up to a dataset-dependent optimum, after which excessive guidance degrades performance.
On CIFAR-10, where the baseline FID is already low, the same trade-off can translate into a FID increase at larger $w$.

In \cref{tab:cg_all}(c), UG improves FID from 5.34 to 5.00 at $\lambda=0.5$ on ImageNet-256.
Because this margin is relatively small, we evaluate over 3 random seeds to confirm the improvement is not due to sampling randomness.
As shown in \cref{tab:ucg_multiseed}, the FID gap persists across seeds with non-overlapping standard deviations ($5.337 \pm 0.042$ vs.\ $4.999 \pm 0.047$), confirming that UG yields a consistent improvement at this operating point.

\subsubsection{Uncertainty-Aware Classifier-Free Guidance (UCFG)}

\begin{figure}[htbp]
	\centering
	\includegraphics[width=0.8\textwidth]{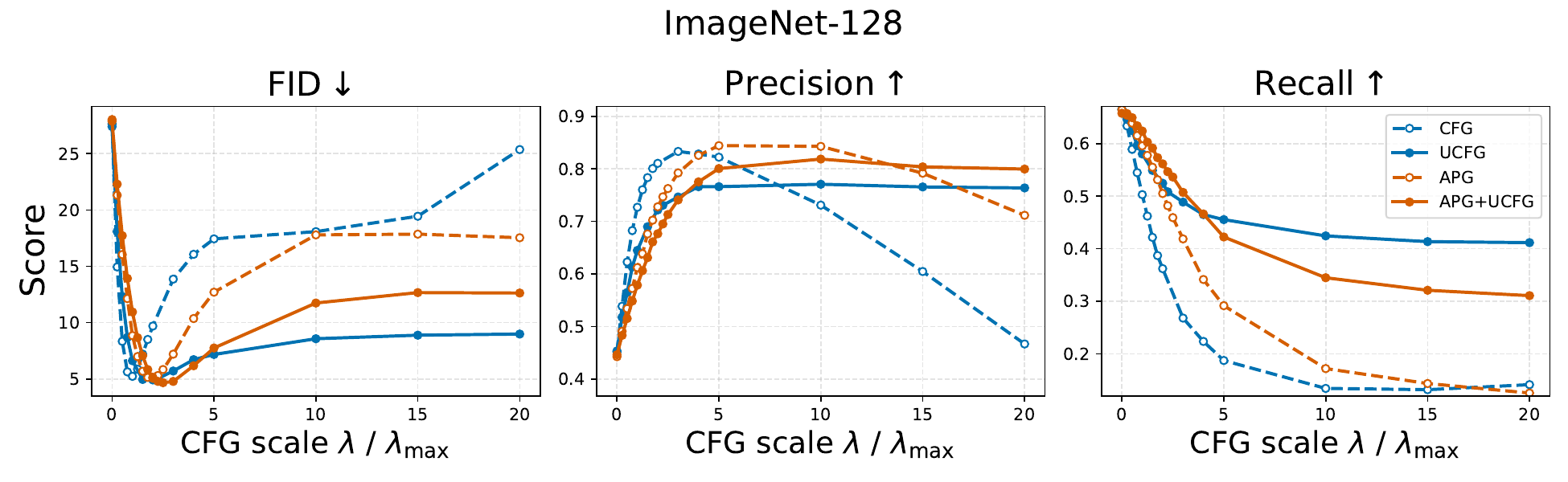}
	\caption{
		\textbf{FID, precision, and recall as a function of the fixed CFG scale $\lambda$ or the maximum UCFG scale $\lambda_{\max}$ on ImageNet-128.}
		CFG degrades sharply at large $\lambda$, while UCFG remains more stable as $\lambda_{\max}$ increases.
	}
	\label{fig:cfg_metrics_imagenet128}
\end{figure}

\begin{figure}[H]
	\centering
	\begin{subfigure}[t]{0.48\textwidth}
		\centering
		\includegraphics[width=\columnwidth]{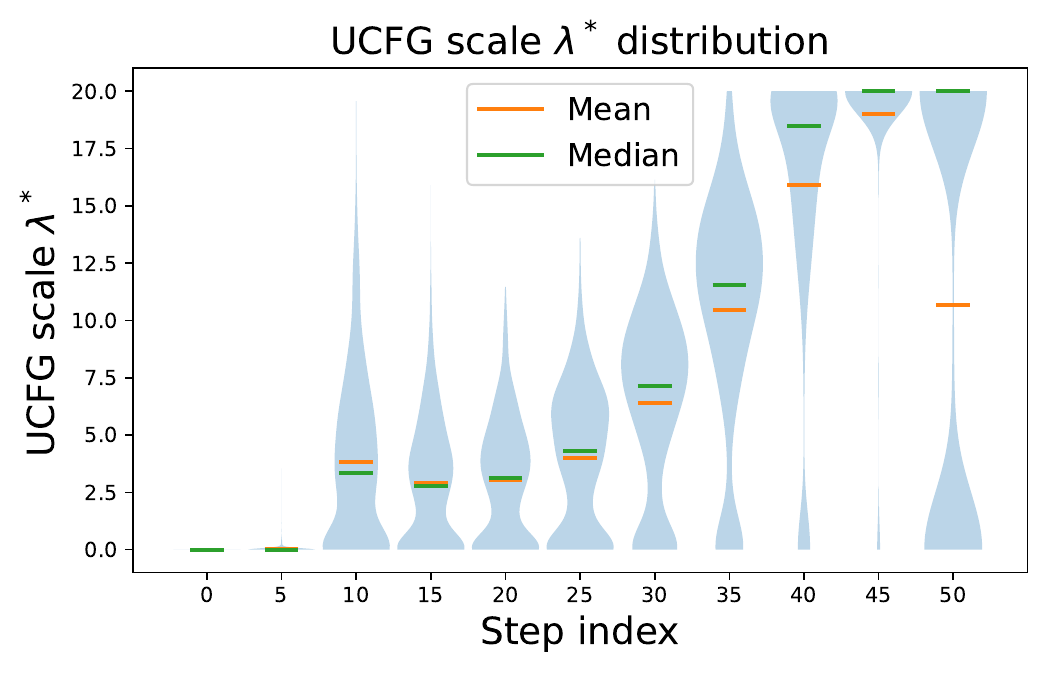}
		\caption{UCFG.}
		\label{fig:cfg_scale_violin_imagenet128}
	\end{subfigure}
	\hfill
	\begin{subfigure}[t]{0.48\textwidth}
		\centering
		\includegraphics[width=\columnwidth]{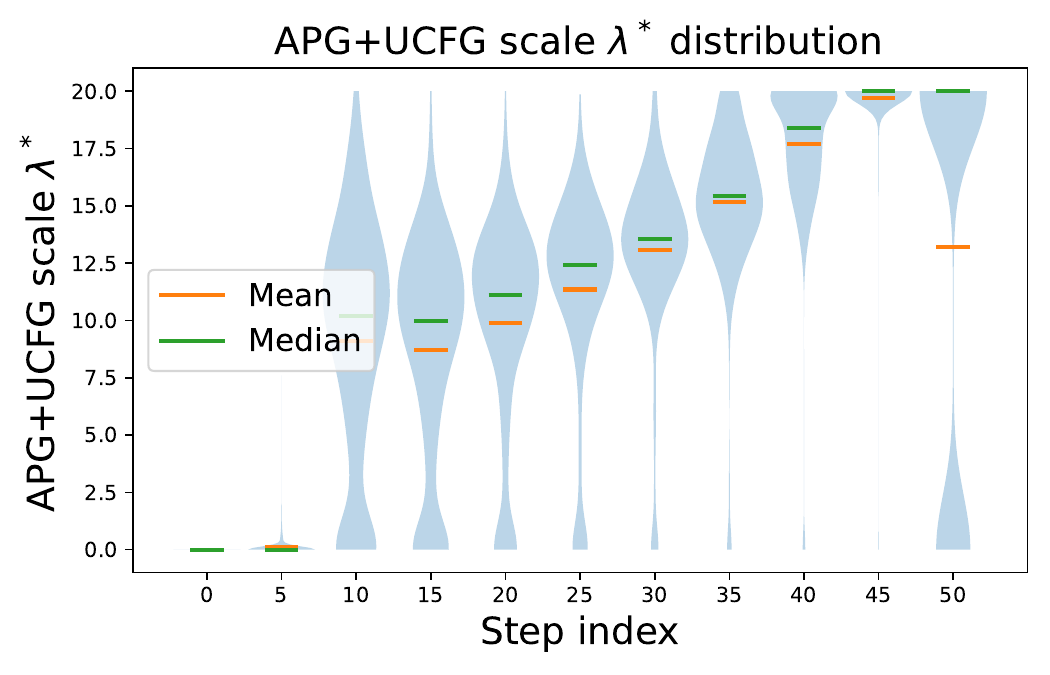}
		\caption{APG+UCFG.}
		\label{fig:cfg_scale_violin_adaptive_apg_imagenet128}
	\end{subfigure}
	\caption{
		\textbf{Distributions of the adaptive guidance scale $\lambda^*$ across sampling steps on ImageNet-128.}
		Violin plots are computed from 1{,}000 samples for (a) UCFG and (b) APG+UCFG.
		In both cases, $\lambda^*$ tends to be smaller in early steps and larger in later steps.
	}
	\label{fig:cfg_scale_violins_imagenet128}
\end{figure}

We compare standard CFG having scale $\lambda$ with UCFG, which uses a step-wise adaptive scale $\lambda^*$ clamped by $\lambda_{\max}$ on ImageNet-128.
In \cref{fig:cfg_metrics_imagenet128}, increasing the fixed CFG scale eventually degrades both precision and recall, leading to a sharp rise in FID at large $\lambda$.
UCFG shows smaller changes in precision/recall and less FID degradation as $\lambda_{\max}$ increases.
The violin plots in \cref{fig:cfg_scale_violins_imagenet128} show that $\lambda^*$ is typically smaller in earlier sampling steps and larger in later steps.
This pattern suggests that adaptive scaling limits over-guidance while the sample is still coarse and permits stronger guidance later in sampling.

\section{Supplementary Analyses}
\label{app:supplementary}



\subsection{Uncertainty Evolution in BayesDiff and UA-Flow}
\label{app:additional_results:uncertainty_evolution}

\begin{figure}[htbp]
	\centering
	\setlength{\tabcolsep}{1.5pt}
	\begin{tabular}{cccc}
		\multicolumn{4}{c}{\includegraphics[width=0.3\textwidth]{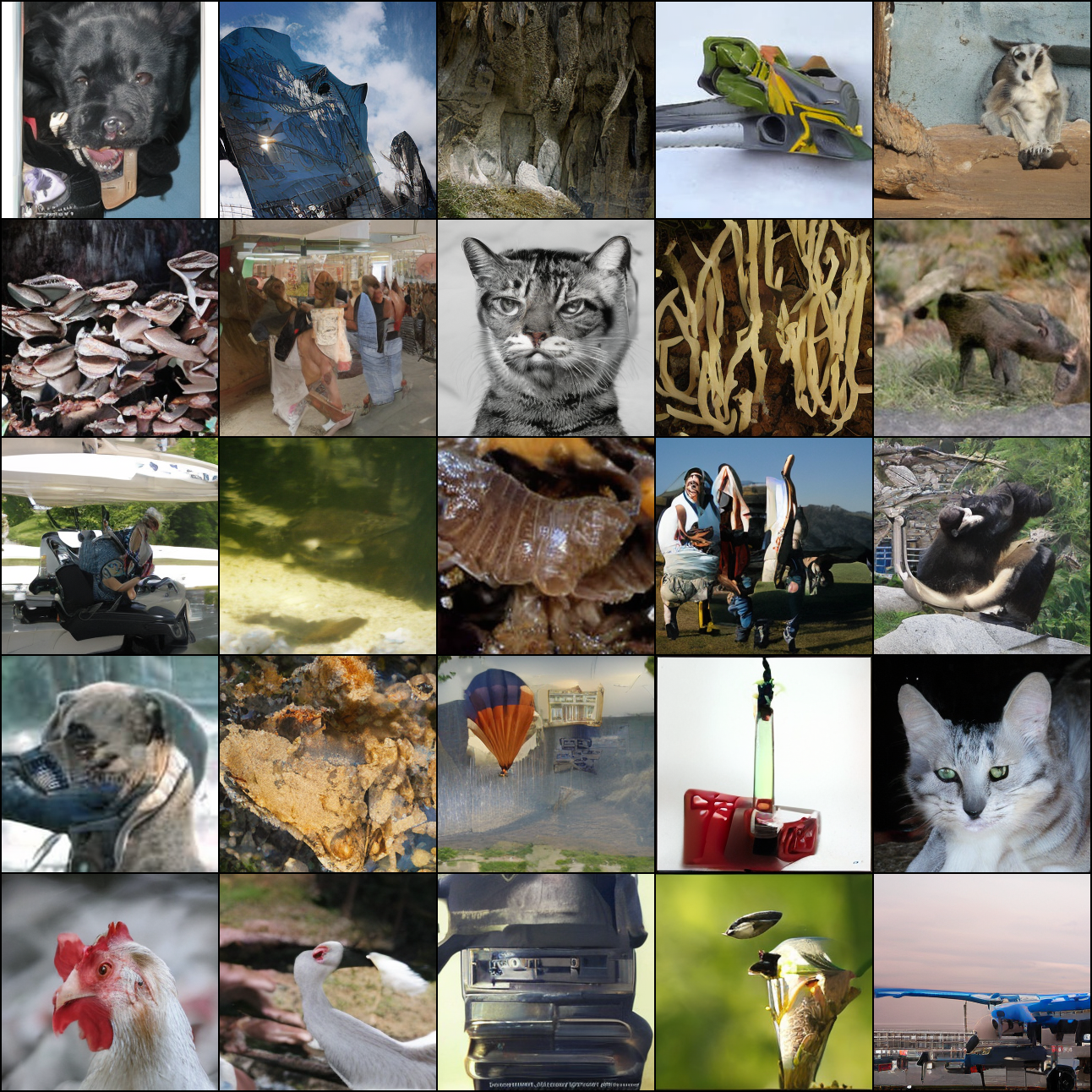}} \\[-0.8ex]
		\includegraphics[width=0.23\textwidth]{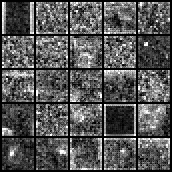} &
		\includegraphics[width=0.23\textwidth]{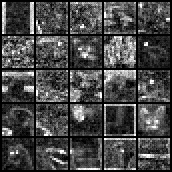} &
		\includegraphics[width=0.23\textwidth]{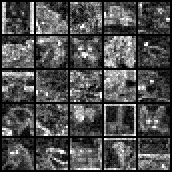} &
		\includegraphics[width=0.23\textwidth]{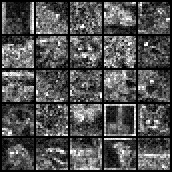} \\[-0.8ex]
		\includegraphics[width=0.23\textwidth]{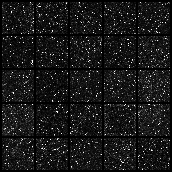} &
		\includegraphics[width=0.23\textwidth]{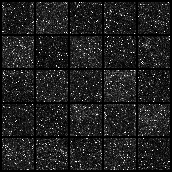} &
		\includegraphics[width=0.23\textwidth]{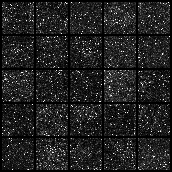} &
		\includegraphics[width=0.23\textwidth]{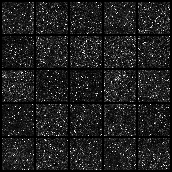} \\
	\end{tabular}
	\caption{
		\textbf{Qualitative uncertainty evolution for BayesDiff.}
		Images are sampled on ImageNet-256 without guidance (no classifier guidance or classifier-free guidance).
		\emph{Top}: the generated sample.
		\emph{Middle}: BayesDiff’s velocity uncertainty maps at intermediate sampling steps 12/24/36/48 (left to right).
		\emph{Bottom}: state uncertainty maps obtained by propagating the velocity uncertainty through the sampling dynamics using the same variance propagation rule as UA-Flow.
		While the velocity uncertainty exhibits spatial correlation, the propagated state uncertainty becomes largely noise-like and fails to preserve coherent spatial structure.
	}
	\label{fig:bayesdiff_uncertainty_evolution}
\end{figure}

\begin{figure}[htbp]
	\centering
	\setlength{\tabcolsep}{1.5pt}
	\begin{tabular}{cccc}
		\multicolumn{4}{c}{\includegraphics[width=0.3\textwidth]{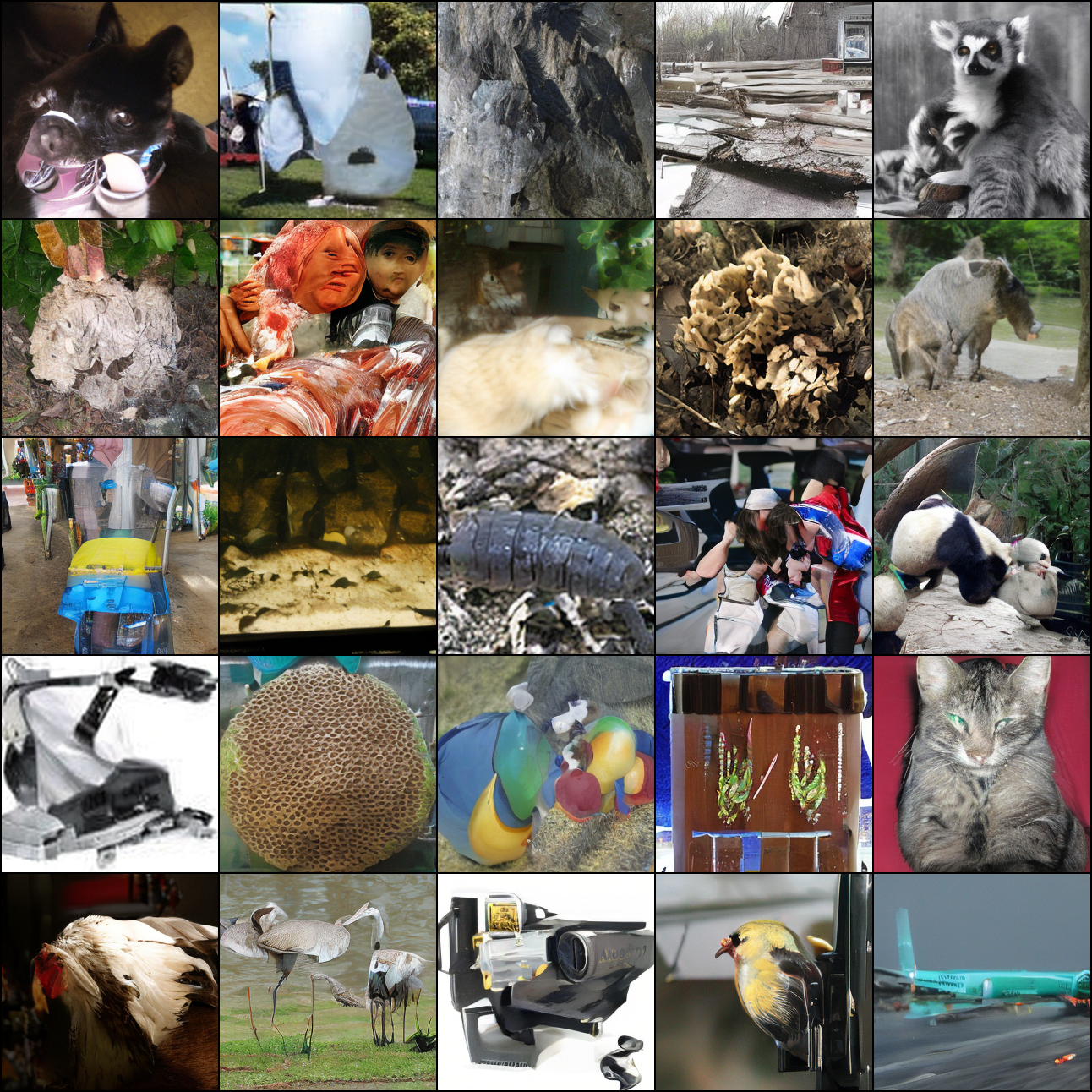}} \\[-0.8ex]
		\includegraphics[width=0.23\textwidth]{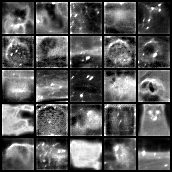} &
		\includegraphics[width=0.23\textwidth]{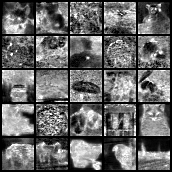} &
		\includegraphics[width=0.23\textwidth]{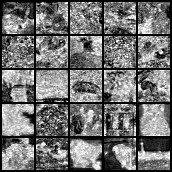} &
		\includegraphics[width=0.23\textwidth]{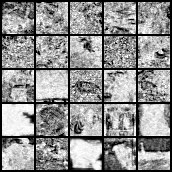} \\[-0.8ex]
		\includegraphics[width=0.23\textwidth]{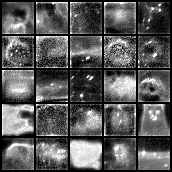} &
		\includegraphics[width=0.23\textwidth]{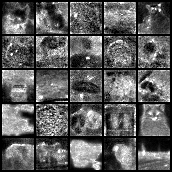} &
		\includegraphics[width=0.23\textwidth]{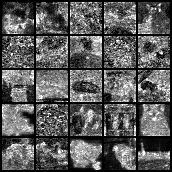} &
		\includegraphics[width=0.23\textwidth]{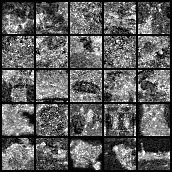} \\
	\end{tabular}
	\caption{
		\textbf{Qualitative uncertainty evolution for UA-Flow.}
		Images are sampled on ImageNet-256 without guidance (no classifier guidance or classifier-free guidance).
		\emph{Top}: the generated sample.
		\emph{Middle}: UA-Flow’s velocity uncertainty maps at intermediate sampling steps 12/24/36/48 (left to right).
		\emph{Bottom}: propagated state uncertainty maps computed from the velocity uncertainty via variance propagation.
		In contrast to BayesDiff, the state uncertainty remains spatially coherent and tracks structured regions throughout sampling.
	}
	\label{fig:uncflow_uncertainty_evolution}
\end{figure}

\begin{figure}[htbp]
	\centering
	\includegraphics[width=0.8\textwidth]{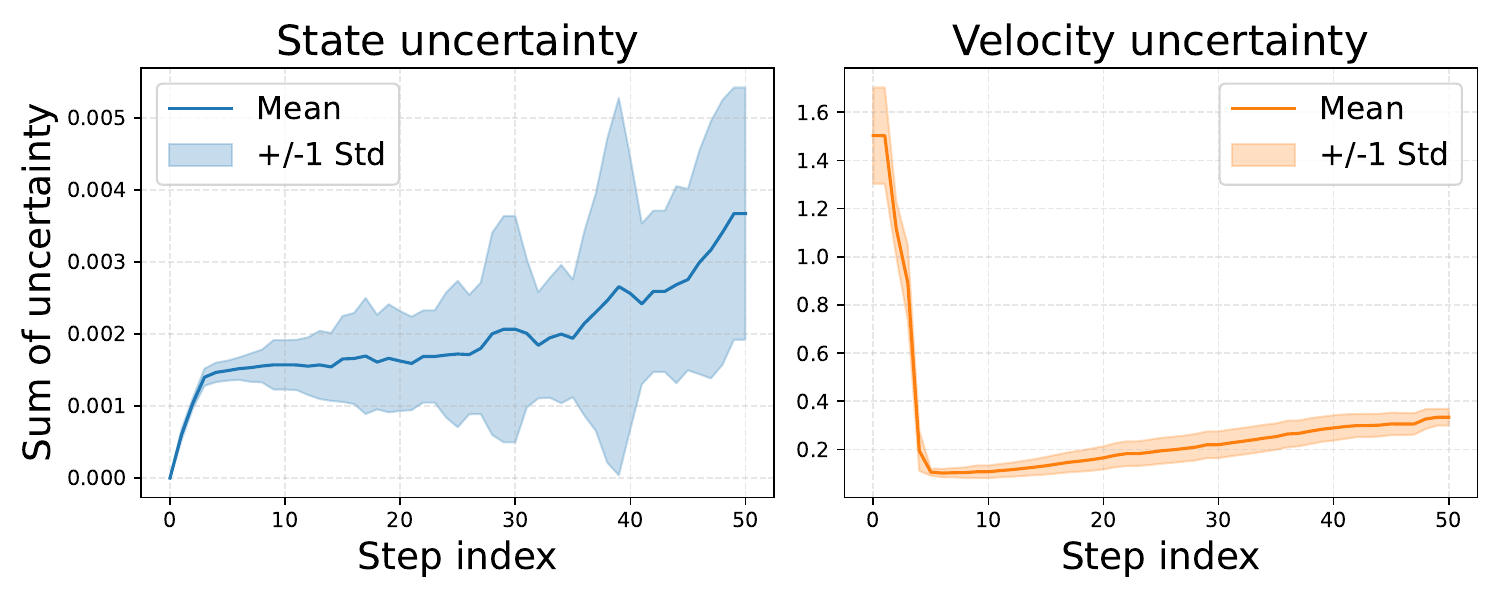}
	\caption{
		\textbf{Uncertainty scale over sampling steps for BayesDiff.}
		We quantify the magnitude of uncertainty at each sampling step by summing the uncertainty map over all elements (spatial locations and channels) for each sample.
		The plot reports the mean $\pm$ standard deviation of this scalar summary across the $5\times5$ sample grid shown in \cref{fig:bayesdiff_uncertainty_evolution}.
		The noisy state uncertainty maps in \cref{fig:bayesdiff_uncertainty_evolution} are consistent with the elevated uncertainty scale in the early sampling phase, suggesting that BayesDiff’s uncertainty is poorly calibrated across sampling time and can dominate variance propagation.
	}
	\label{fig:bayesdiff_uncertainty_components}
\end{figure}

\Cref{fig:bayesdiff_uncertainty_evolution,fig:uncflow_uncertainty_evolution} compare the evolution of uncertainty maps over the sampling trajectory for BayesDiff and UA-Flow.
Each figure shows the generated image (top), the estimated velocity uncertainty at intermediate steps (middle), and the propagated state uncertainty at the corresponding steps (bottom).
Notably, both methods produce velocity uncertainty maps with clear spatial correlation, indicating that uncertainty concentrates in specific regions rather than being uniformly distributed.
More specifically, velocity uncertainty obtained from UA-Flow is typically less noisy.

Despite using the same variance propagation rule (\cref{eq:variance_propagation}), the propagated state uncertainty behaves very differently.
For BayesDiff, the state uncertainty maps become largely unstructured and visually resemble noise, whereas UA-Flow yields state uncertainty maps that remain spatially coherent across steps.
This behavior is consistent with \cref{fig:bayesdiff_uncertainty_components}: BayesDiff exhibits a large uncertainty scale in the early sampling phase, and the corresponding early-step velocity uncertainty maps are also visibly noisy (e.g., step 12 in \cref{fig:bayesdiff_uncertainty_evolution}).
When such high-magnitude, noise-like velocity uncertainty is propagated, it can dominate the resulting state uncertainty and produce the unstructured, noise-like maps observed in \cref{fig:bayesdiff_uncertainty_evolution}.
In contrast, UA-Flow’s learned heteroscedastic velocity uncertainty is more spatially coherent, leading to propagated state uncertainty maps that remain structured.

\subsection{Qualitative verification of uncertainty-based filtering}
\label{app:qualitative_filtering_cfg}

\begin{figure}[htbp]
	\centering
	\includegraphics[width=0.45\textwidth]{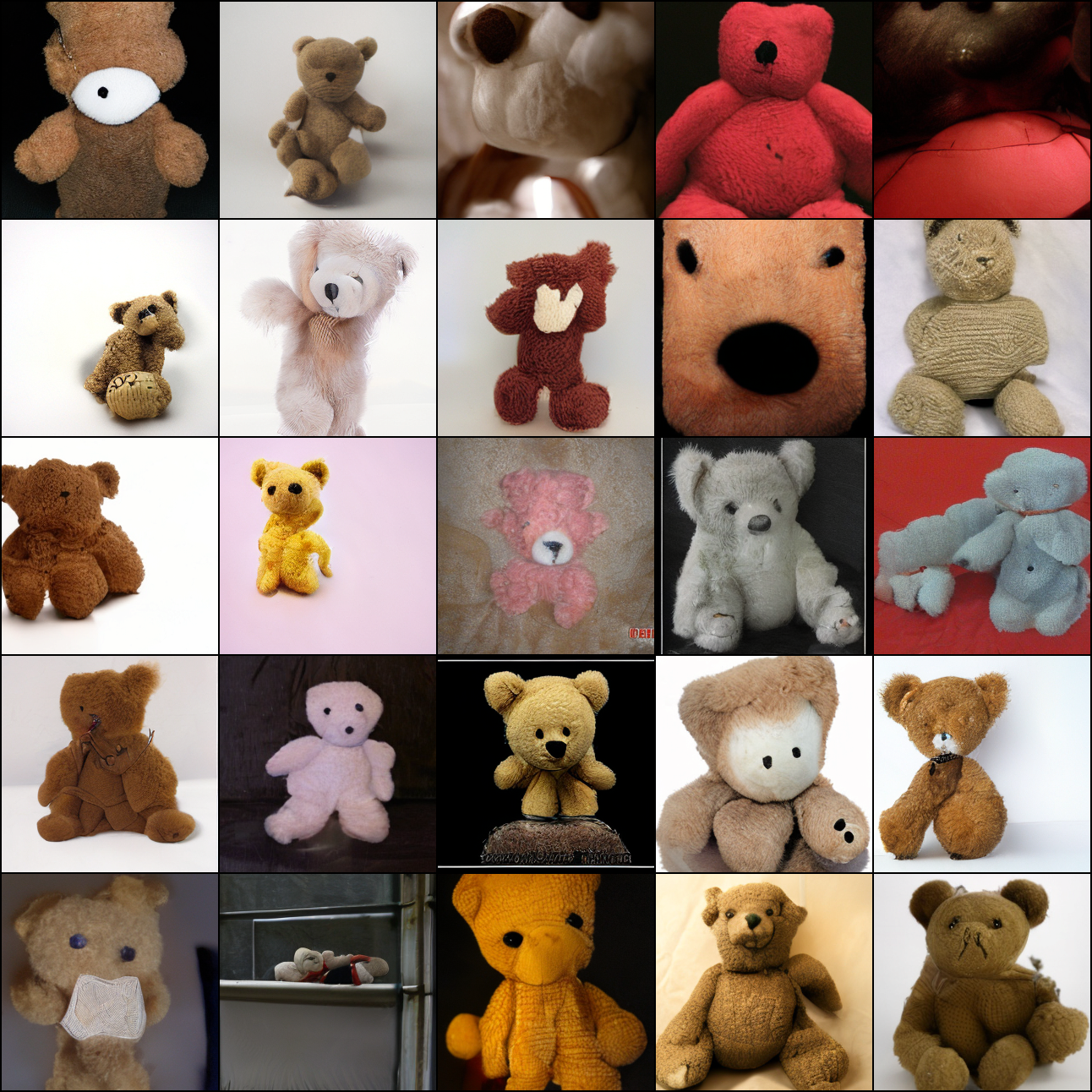}
	\hfill
	\includegraphics[width=0.45\textwidth]{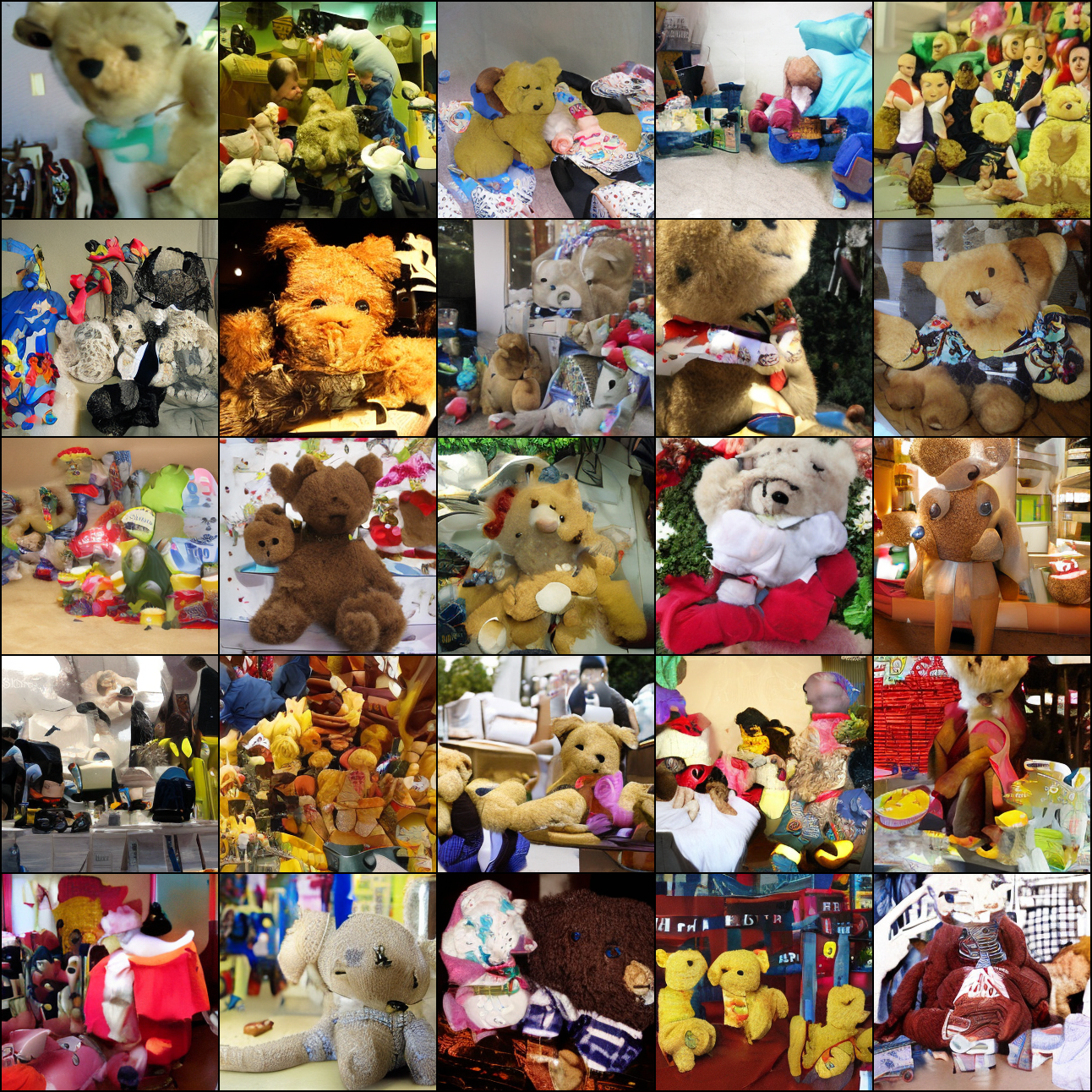}
	\caption{
		\textbf{Teddy bear (ImageNet-256, CFG $\lambda=0.5$): lowest vs.\ highest uncertainty samples.}
		We generate 1{,}000 samples and select the 25 lowest-uncertainty images (top) and the 25 highest-uncertainty images (bottom) using the same uncertainty aggregation and ranking procedure as in the main paper.
		Low-uncertainty samples are visually clean and class-consistent, while high-uncertainty samples often contain clutter, distortions, or weak class identity.
	}
	\label{fig:qual_filtering_teddy}
\end{figure}

\begin{figure}[htpb]
	\centering
	\includegraphics[width=0.45\textwidth]{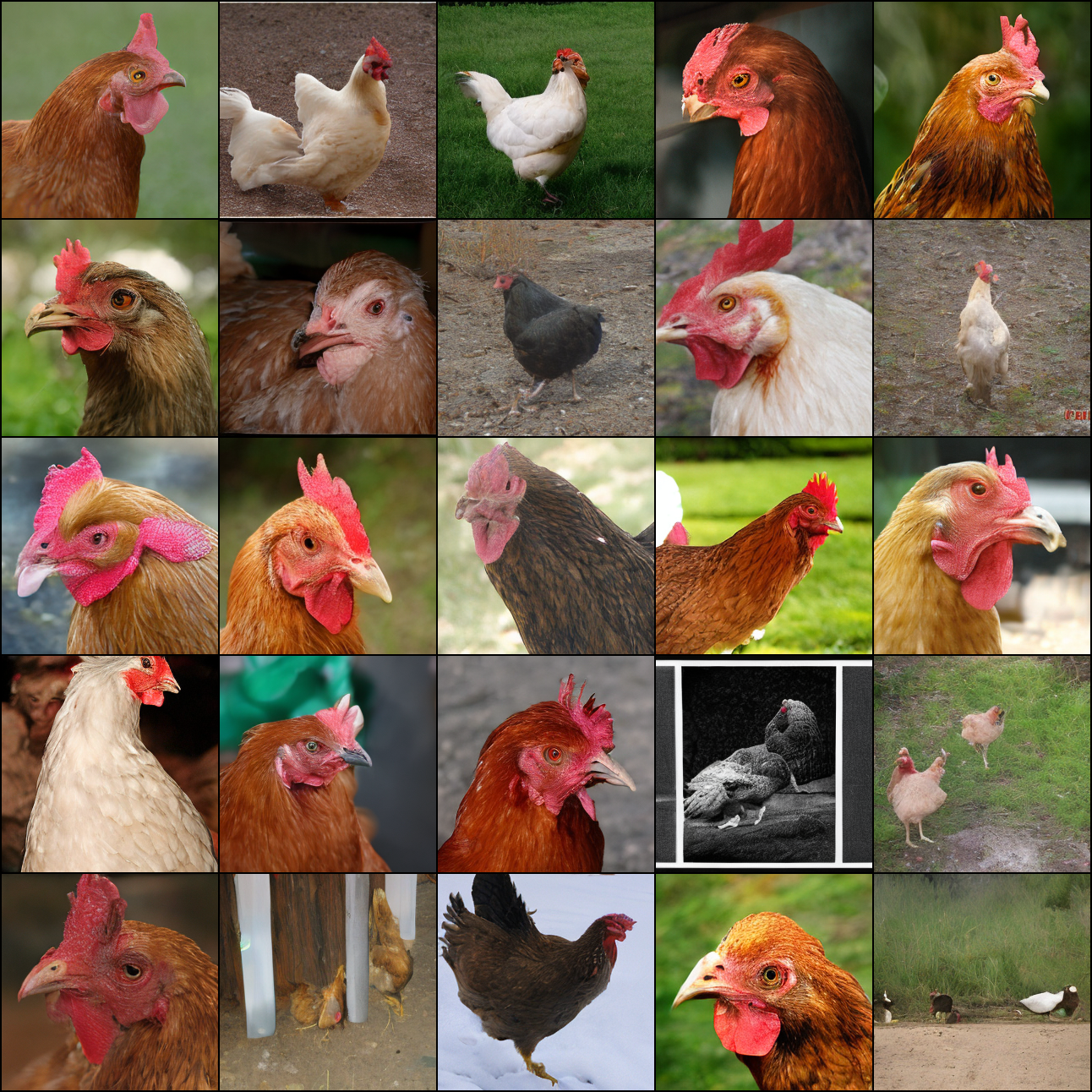}
	\hfill
	\includegraphics[width=0.45\textwidth]{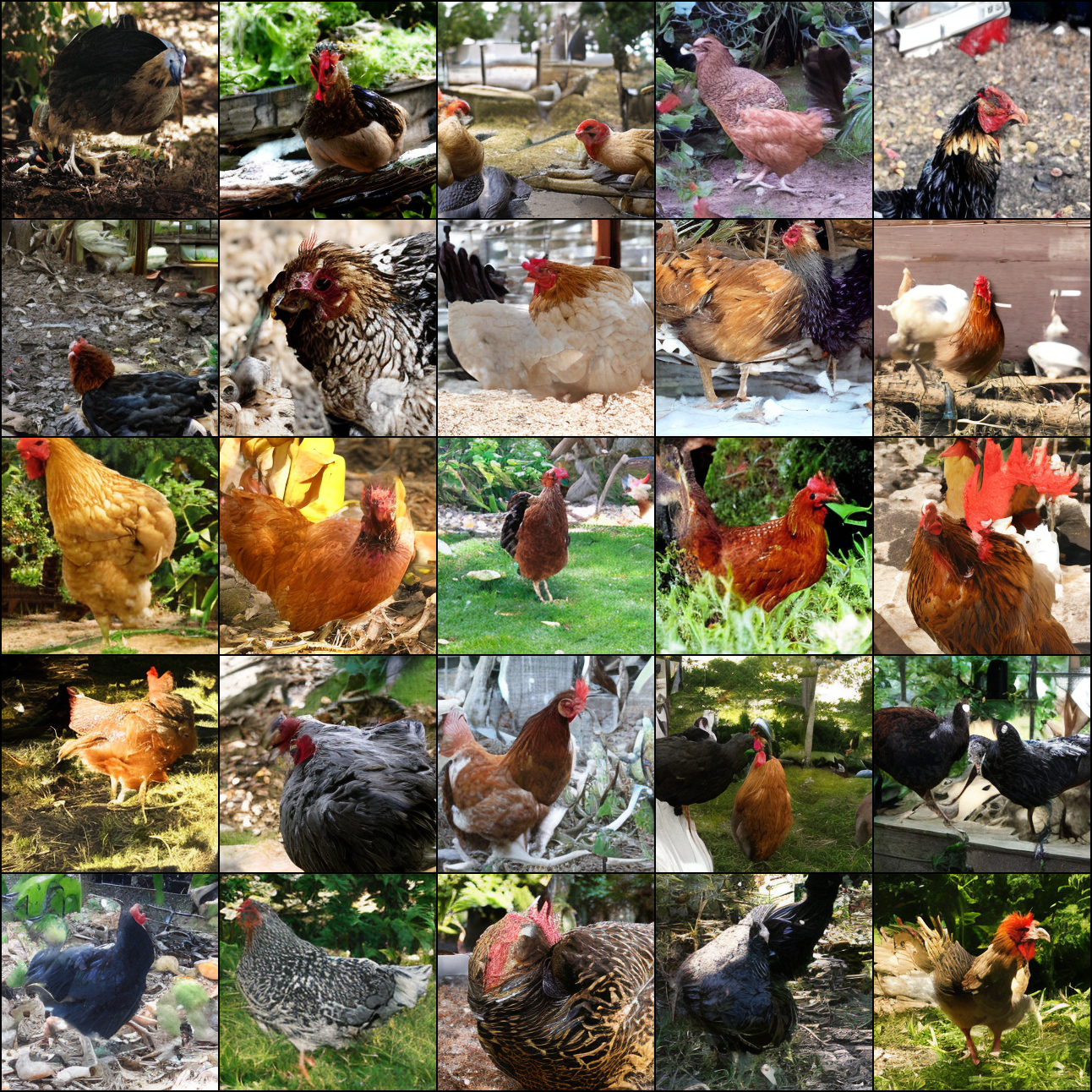}
	\caption{
		\textbf{Hen (ImageNet-256, CFG $\lambda=0.5$): lowest vs.\ highest uncertainty samples.}
		We generate 1{,}000 samples and select the 25 lowest-uncertainty images (top) and the 25 highest-uncertainty images (bottom).
		Low uncertainty corresponds to high-fidelity, easily recognizable hens (often in canonical views), whereas high uncertainty corresponds to less reliable generations with more frequent artifacts or reduced class salience, illustrating a qualitative fidelity--diversity trade-off.
	}
	\label{fig:qual_filtering_hen}
\end{figure}

We next qualitatively assess whether UA-Flow's sample-level uncertainty score distinguishes generated images by fidelity.
To this end, we focus on two ImageNet-256 classes, \texttt{hen} and \texttt{teddy\_bear}, and generate 1{,}000 samples per class using classifier-free guidance with scale $\lambda=0.5$ to show more realistic images.
For each generated image, we compute the same scalar uncertainty score used for filtering in \cref{subsec:filtering_exp}, then respectively select the 25 images with the highest uncertainty and lowest uncertainty.

\cref{fig:qual_filtering_teddy,fig:qual_filtering_hen} show clear qualitative separation between uncertainty extremes.
High-uncertainty samples frequently exhibit severe structural distortions and cluttered scenes.
Low-uncertainty samples tend to be recognizable instances of the target class with clean textures and coherent global structure.
These visual differences suggest higher perceptual fidelity in the low-uncertainty subset.

At the same time, the low-uncertainty subsets also reveal a fidelity-diversity trade-off.
For \texttt{hen}, low-uncertainty samples are dominated by canonical compositions, typically close-up head/torso views on simple backgrounds.
For \texttt{teddy\_bear}, low-uncertainty samples often depict centered plush toys under relatively uniform backgrounds.
In contrast, the high-uncertainty subsets span more diverse contexts, poses, and compositions, albeit with noticeably lower fidelity.
Overall, these results provide qualitative evidence that UA-Flow uncertainty is aligned with sample-level quality, and that uncertainty-based filtering behaves as observed in \cref{subsec:filtering_exp}: retaining low-uncertainty (high-fidelity) samples while implicitly reducing diversity.

\subsection{Calibration analysis via reconstruction on validation images}
\label{app:calibration_analysis}

Generative models lack ground-truth outputs since the model is designed to sample from the data distribution, not to reconstruct a specific target.
We therefore use partially noised real images whose clean originals are known.
These originals provide reference targets for an indirect calibration of the predicted endpoint variance.

\paragraph{Protocol.}
We encode 500 each ImageNet images for calibration and validation to latent space $\mathbf{z}_1 \in \mathbb{R}^{C \times H \times W}$ via the Stable Diffusion autoencoder ($C=4$ channels).
For each $t \in \{0.5, 0.7, 0.9\}$, we sample noise $\epsilon \sim \mathcal{N}(0,\mathbf{I})$ and compute the noisy state $\mathbf{z}_t = \alpha_t \mathbf{z}_1 + \beta_t \epsilon$.
We then run the ODE from $\mathbf{z}_t$ at time $t$ to the predicted endpoint $\hat{\mathbf{z}}_1$ at time $1.0$.
For each method, we also compute the endpoint state uncertainty $\Var[\mathbf{z}_1\mid\mathbf{z}_t]$.
The true latent $\mathbf{z}_1$ serves as the ground-truth target for calibration.

We evaluate calibration using two standard metrics computed over all $M = N \times C \times H \times W$ scalar elements (images $\times$ channels $\times$ spatial locations):
\begin{itemize}
	\item \textbf{ECE (Expected Calibration Error)}: For each element $i$, we compute the CDF value $p_i = \Phi\!\bigl((\mathbf{z}_{1,i} - \hat{\mathbf{z}}_{1,i}) / \sqrt{\Var[\mathbf{z}_{1,i}]}\bigr)$.
Here, $\Phi$ is the standard normal CDF.
For a calibrated predictive distribution, the PIT values should be uniform on $[0,1]$.
We evaluate their empirical CDF at $B+1$ evenly spaced quantile levels $q_b=b/B$, with $b=0,\dots,B$.
We report $\text{ECE} = \frac{1}{B+1}\sum_{b=0}^{B} |\hat{F}(q_b) - q_b|$, where $\hat{F}(q) = \frac{1}{M}\sum_i \mathbf{1}[p_i \le q]$.

	\item \textbf{Brier score for the 95\% coverage event}: We form the central Gaussian prediction interval $I_q$ for each element at $q=0.95$.
We then compute $\frac{1}{M}\sum_i (q - \mathbf{1}[\mathbf{z}_{1,i} \in I_q])^2$.
The event is whether the clean latent value lies inside this interval.
\end{itemize}
Post-hoc calibration is applied by fitting a scalar on a held-out calibration set (500 separate images) and rescaling the predicted variance. We use isotonic regression for the variance calibration~\citep{Kuleshov18Calib}. 

We compare UA-Flow against BayesDiff and AU.
GenUnc is excluded because it produces a scalar uncertainty per sample, whereas UA-Flow, BayesDiff, and AU all produce per-component uncertainty, making them directly comparable.
UA-Flow and BayesDiff are compared across all $t$. AU is reported only at $t=0.9$ since its uncertainty is defined only at the later steps of sampling.

\begin{table}[htbp]
	\centering
	\caption{
		\textbf{Calibration results on ImageNet-256 validation images.}
		UA-Flow achieves lower ECE and Brier score than BayesDiff across all noise levels $t$, both before and after post-hoc calibration.
		AU is reported only at $t=0.9$.
	}
	\label{tab:calibration}
	\setlength{\tabcolsep}{2pt}
	\begin{subtable}[t]{0.48\textwidth}
		\centering
		\caption{Before post-hoc calibration.}
		\label{tab:calibration_before}
		\begin{tabular}{lcccccc}
			\toprule
			& \multicolumn{2}{c}{UA-Flow} & \multicolumn{2}{c}{BayesDiff} & \multicolumn{2}{c}{AU} \\
			\cmidrule(lr){2-3} \cmidrule(lr){4-5} \cmidrule(lr){6-7}
			$t$ & ECE$\downarrow$ & Brier$\downarrow$ & ECE$\downarrow$ & Brier$\downarrow$ & ECE$\downarrow$ & Brier$\downarrow$ \\
			\midrule
			0.5 & \textbf{0.1328} & \textbf{0.4771} & 0.2381 & 0.9025 & --- & --- \\
			0.7 & \textbf{0.0992} & \textbf{0.3281} & 0.2381 & 0.9025 & --- & --- \\
			0.9 & \textbf{0.0277} & \textbf{0.1009} & 0.2386 & 0.9025 & 0.0485 & 0.1782 \\
			\bottomrule
		\end{tabular}
	\end{subtable}
	\hfill
	\begin{subtable}[t]{0.48\textwidth}
		\centering
		\caption{After post-hoc calibration.}
		\label{tab:calibration_after}
		\begin{tabular}{lcccccc}
			\toprule
			& \multicolumn{2}{c}{UA-Flow} & \multicolumn{2}{c}{BayesDiff} & \multicolumn{2}{c}{AU} \\
			\cmidrule(lr){2-3} \cmidrule(lr){4-5} \cmidrule(lr){6-7}
			$t$ & ECE$\downarrow$ & Brier$\downarrow$ & ECE$\downarrow$ & Brier$\downarrow$ & ECE$\downarrow$ & Brier$\downarrow$ \\
			\midrule
			0.5 & \textbf{0.0066} & \textbf{0.0515} & 0.0105 & 0.0558 & --- & --- \\
			0.7 & \textbf{0.0038} & \textbf{0.0492} & 0.0073 & 0.0546 & --- & --- \\
			0.9 & \textbf{0.0008} & \textbf{0.0481} & 0.0038 & 0.0505 & 0.0012 & 0.0496 \\
			\bottomrule
		\end{tabular}
	\end{subtable}
\end{table}

\paragraph{Results.}
\Cref{tab:calibration} presents calibration results before and after post-hoc calibration.
Before calibration, UA-Flow consistently outperforms BayesDiff across all noise levels, with the gap narrowing as $t$ increases.
At $t=0.9$, UA-Flow achieves ECE of 0.0277 and Brier of 0.1009, compared to 0.2386/0.9025 for BayesDiff and 0.0485/0.1782 for AU.
At high $t$, the noisy state retains most of the original signal, so the ODE effectively reconstructs the original image and calibration should be strong.
At low $t$, the signal is heavily corrupted and the trajectory resembles conditional generation, making calibration inherently harder.
This provides a natural sanity check: calibration quality should improve monotonically with $t$.
As expected, UA-Flow's calibration improves monotonically with $t$.
In contrast, BayesDiff shows nearly constant, higher ECE and Brier scores across the tested values of $t$.

After post-hoc calibration via isotonic regression, ECE and Brier scores decrease for all methods.
UA-Flow retains its advantage, reaching ECE of 0.0008 and Brier of 0.0481 at $t=0.9$, compared to 0.0038/0.0505 for BayesDiff and 0.0012/0.0496 for AU.
These results provide evidence that UA-Flow's uncertainty estimates are meaningfully calibrated and serve as a reliable measure of sample quality beyond their practical usefulness for filtering.

\subsection{Empirical validation of conditional--unconditional velocity correlation for UCFG}
\label{app:cfg_velocity_correlation}

\begin{figure}[htbp]
	\centering
	\includegraphics[width=0.8\textwidth]{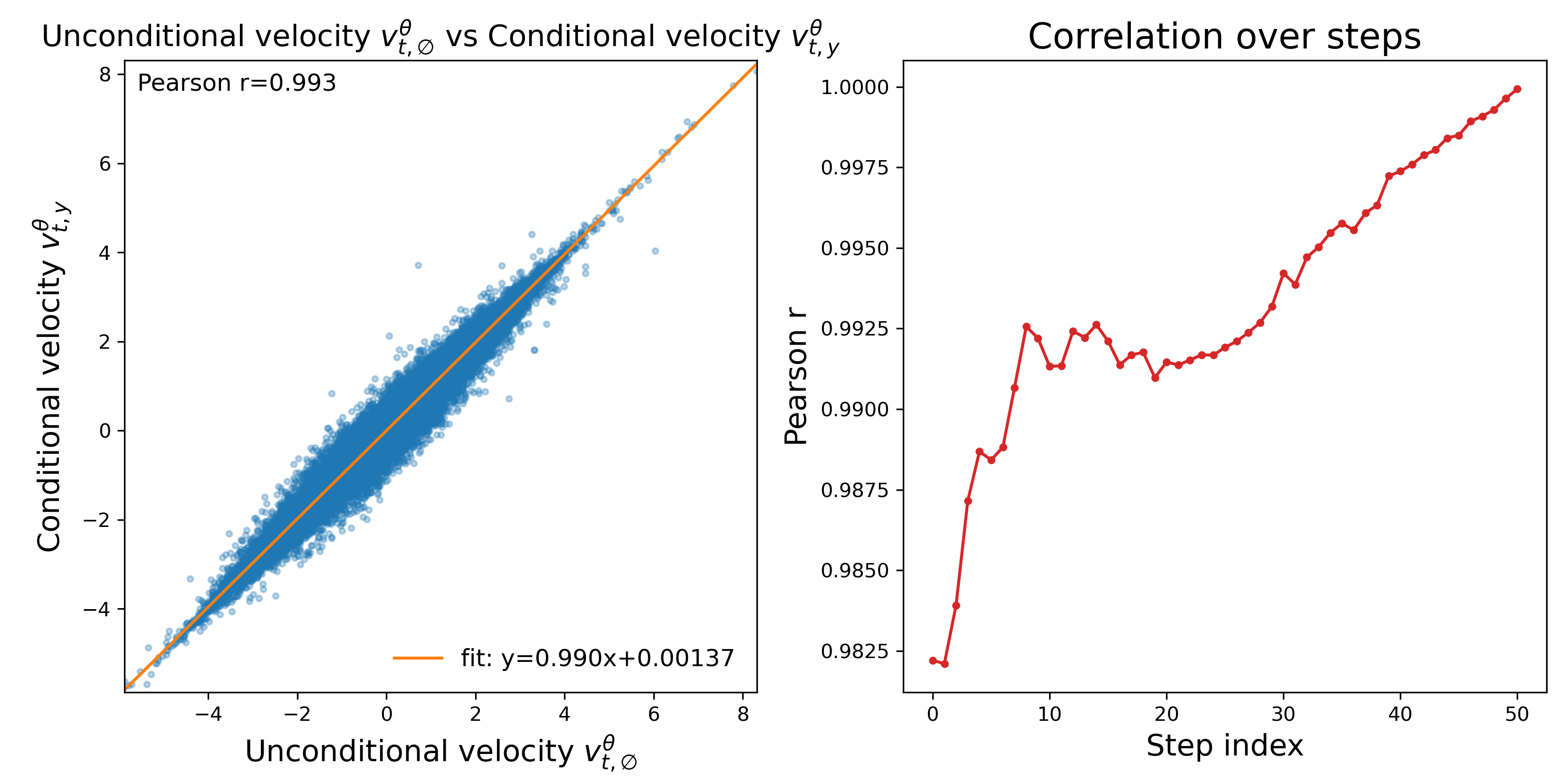}
	\caption{
		\textbf{Correlation between unconditional and conditional velocities.}
		\textit{Left:} Scatter plot of unconditional vs.\ conditional predicted mean velocities $(\bar{u}^\theta_{t,\varnothing}, \bar{u}^\theta_{t,y})$, using $10^6$ randomly sampled elements across all sampling steps and 1{,}000 generated samples without CFG ($\lambda=0$).
		\textit{Right:} Pearson correlation between $\bar{u}^\theta_{t,\varnothing}$ and $\bar{u}^\theta_{t,y}$ at each sampling step, aggregated over all elements and samples; the correlation exceeds $0.98$ at every step.
	}
	\label{fig:cfg_velocity_correlation}
\end{figure}

Uncertainty-aware classifier-free guidance (UCFG) combines conditional and unconditional predictions and approximates the variance of the extrapolated velocity by setting their correlation coefficient to $r_i=1$ (see \cref{eq:cfg_var_approx}).
To validate this assumption empirically, we measure the correlation between the predicted conditional and unconditional velocities along the sampling trajectory.

We generate 1{,}000 samples without CFG (i.e. $\lambda = 0$) using the same sampling configuration as in our main experiments.
At each intermediate sampling step, we evaluate the model twice at the current state: once with the class condition $y$ and once with the null condition $\varnothing$, obtaining the element-wise mean velocities $\bar{u}^\theta_{t,y}(\bar{\mathbf{x}}_t)$ and $\bar{u}^\theta_{t,\varnothing}(\bar{\mathbf{x}}_t)$, respectively.
To visualize their global relationship across time and spatial locations, we randomly subsample $10^6$ element pairs $\big(\bar{u}^\theta_{t,\varnothing}, \bar{u}^\theta_{t,y}\big)$ from all steps and samples.

Figure~\ref{fig:cfg_velocity_correlation} (left) shows an almost linear relationship between the unconditional and conditional velocities, with Pearson correlation $r=0.993$ over the subsampled elements pooled across all sampling steps.
Figure~\ref{fig:cfg_velocity_correlation} (right) reports the per-step Pearson correlations, which are consistently close to $1.0$ and exceed $0.98$ at every step.
Together, these results empirically support the approximation $r_i=1$ used to compute the UCFG velocity variance in \cref{eq:cfg_var_approx}.

\section{Ablations on Uncertainty Estimation and Guidance}
\label{app:additional_results:uncertainty_ablations}

\subsection{Ablations on the Uncertainty Estimation Pipeline}
\label{app:additional_results:uncertainty_ablations:variance}
\label{app:probe_s}
\label{app:variance_propagation_method}
\label{app:sparse_uncertainty}
\label{app:aggregation_ratio}

\begin{figure}[htbp]
	\centering
	\begin{subfigure}[t]{0.8\textwidth}
		\centering
		\includegraphics[width=\textwidth]{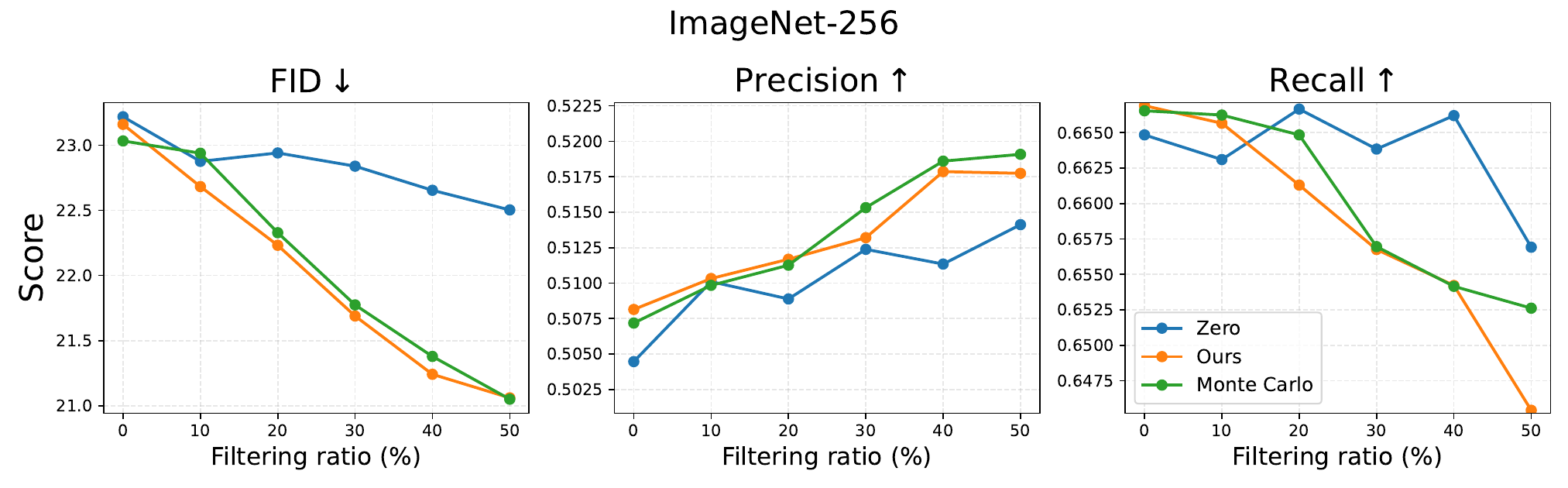}
		\caption{Covariance approximation in \cref{eq:cov_approximation}.}
		\label{fig:filtering_covariance_options_imagenet256}
	\end{subfigure}\\[0.5ex]
	\begin{subfigure}[t]{0.75\textwidth}
		\centering
		\includegraphics[width=\textwidth]{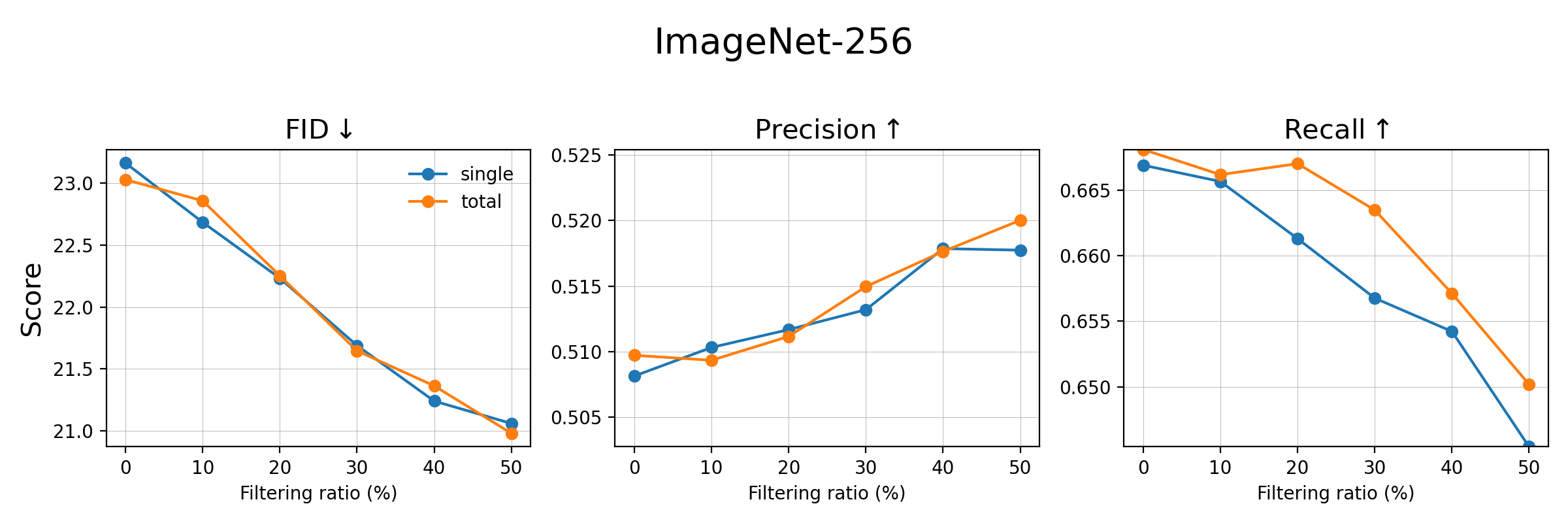}
		\caption{Including the additional variance term $\Var(\bar{u}^\theta_t(\mathbf{x}_t))$ in \cref{eq:variance_approx}.}
		\label{fig:filtering_variance_term_imagenet256}
	\end{subfigure}
	\begin{subfigure}[t]{0.22\textwidth}
		\centering
		\includegraphics[width=\textwidth]{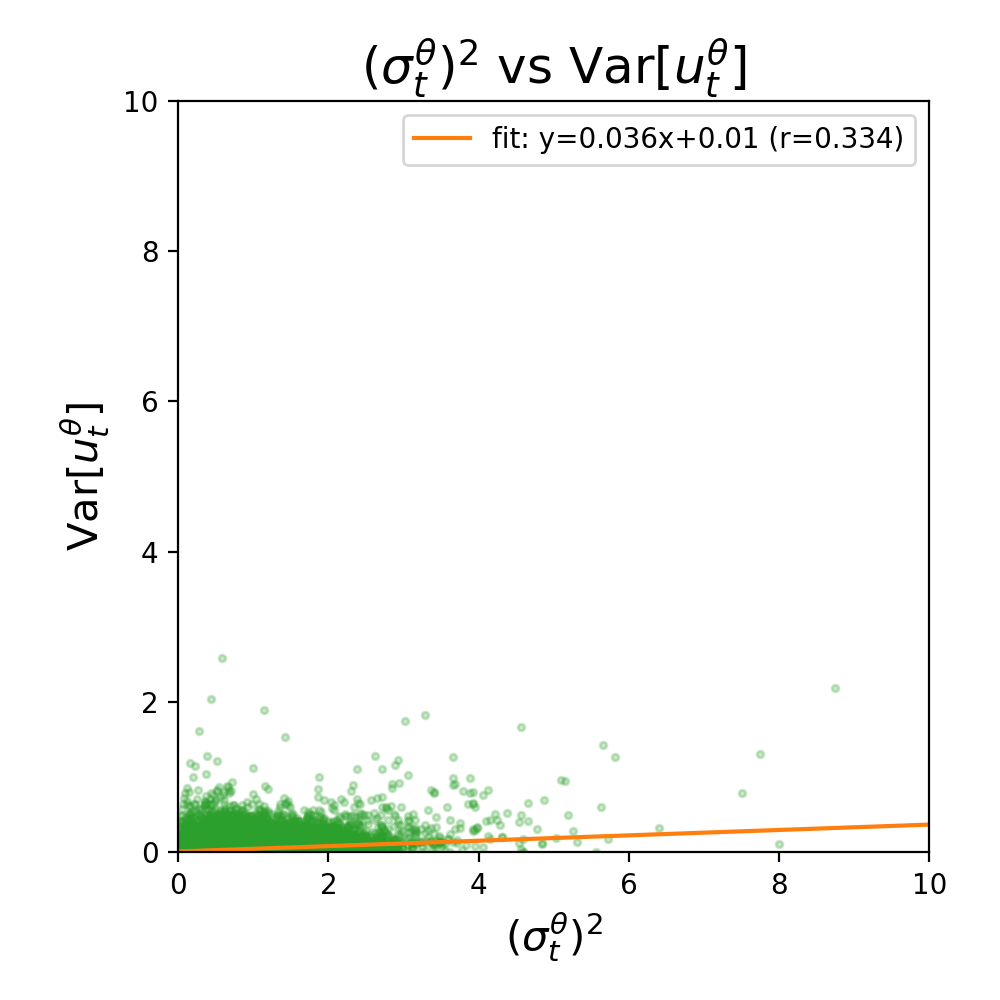}
		\caption{Variance comparison.}
		\label{fig:filtering_variance_term_scatter_imagenet256}
	\end{subfigure}
	\\
	\begin{subfigure}[t]{0.8\textwidth}
		\centering
		\includegraphics[width=\textwidth]{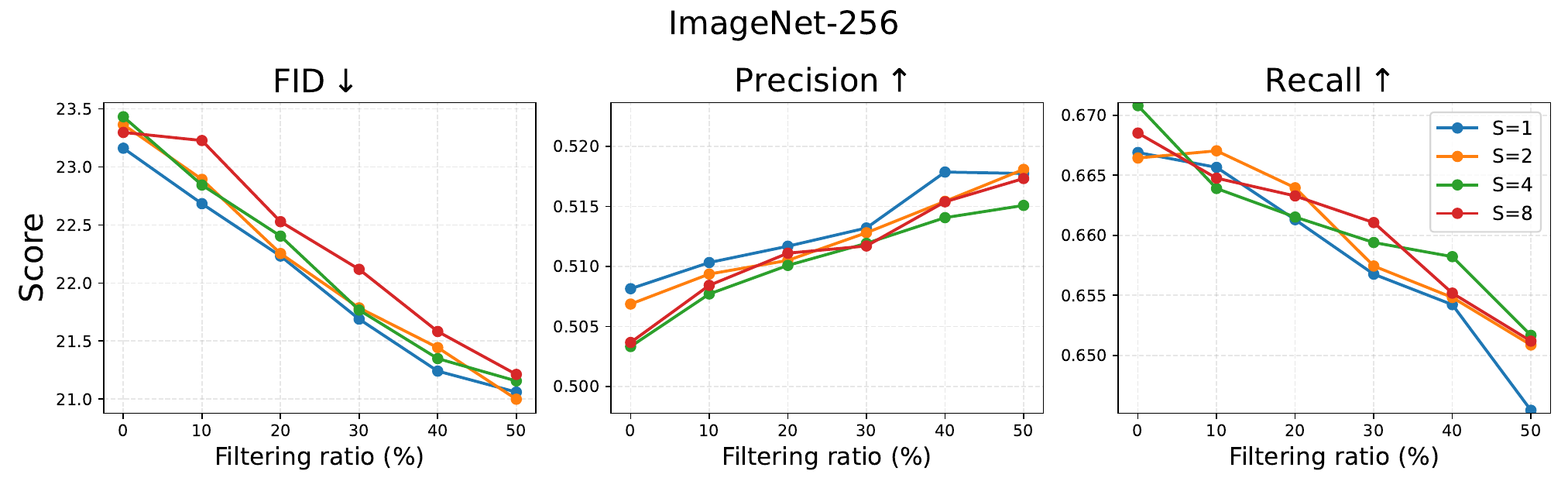}
		\caption{Number of Hutchinson probes $S\in\{1,2,4,8\}$ in \cref{eq:cov_approx_jvp}.}
		\label{fig:filtering_random_probes_imagenet256}
	\end{subfigure}
	\\[0.5ex]
	\begin{subfigure}[t]{0.8\textwidth}
		\centering
		\includegraphics[width=\textwidth]{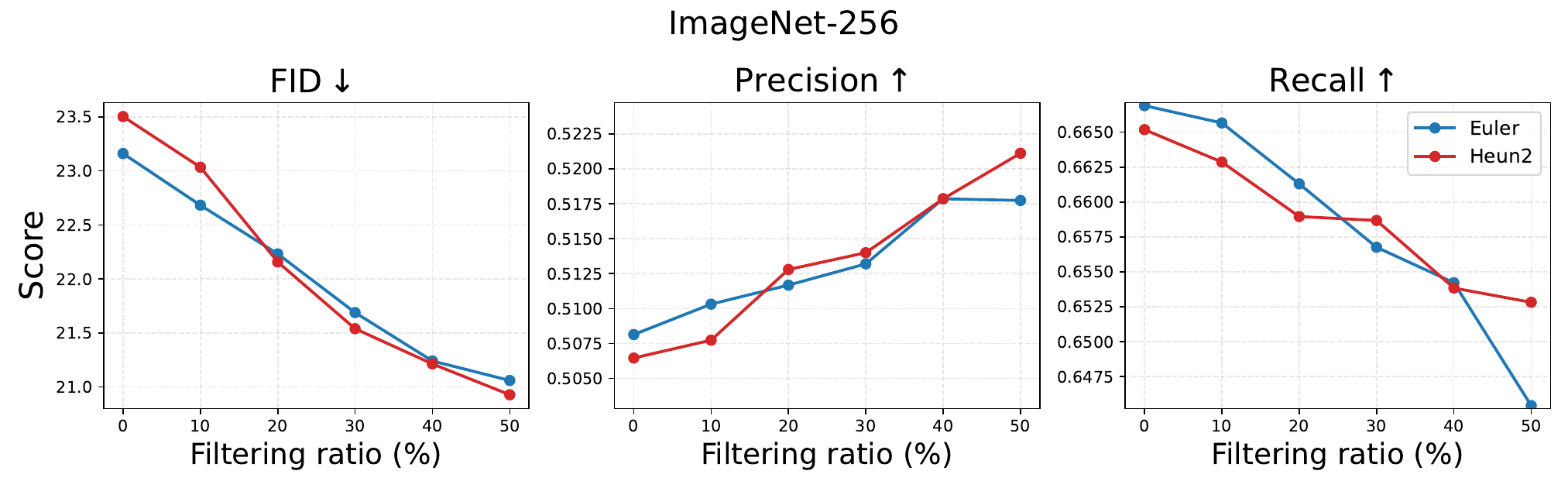}
		\caption{Variance propagation method: Euler (default) vs.\ Heun2.}
		\label{fig:filtering_variance_propagation_imagenet256}
	\end{subfigure}
	\caption{
		\textbf{Ablations on UA-Flow's uncertainty estimation pipeline (ImageNet-256, UA-Flow, no guidance).}
		Each panel sweeps a single design choice while keeping the rest of the filtering pipeline fixed, and reports FID, precision, and recall as a function of the filtering ratio.
		Panel~(c) additionally shows a scatter plot of $\Var(\bar{u}^\theta_t(\mathbf{x}_t))$ (MC estimate with $K=10$) versus $(\sigma^\theta_t(\bar{\mathbf{x}}_t))^2$ across sampling steps and spatial locations.
		Continued on the next page.
	}
	\label{fig:uncertainty_ablations_combined}
\end{figure}

\begin{figure}[htbp]\ContinuedFloat
	\centering
	\setcounter{subfigure}{5}
	\begin{subfigure}[t]{0.8\textwidth}
		\centering
		\includegraphics[width=\textwidth]{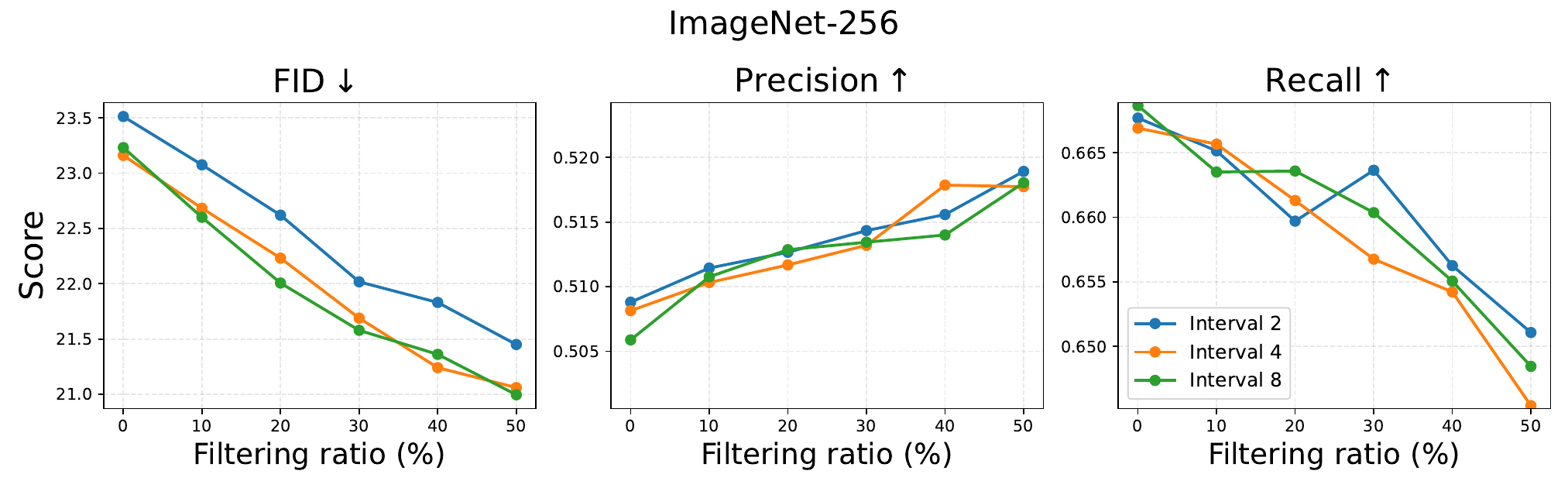}
		\caption{Sparse uncertainty updates: every 2, 4 (default), or 8 steps.}
		\label{fig:filtering_sparse_uncertainty_imagenet256}
	\end{subfigure}\\[0.5ex]
	\begin{subfigure}[t]{0.8\textwidth}
		\centering
		\includegraphics[width=\textwidth]{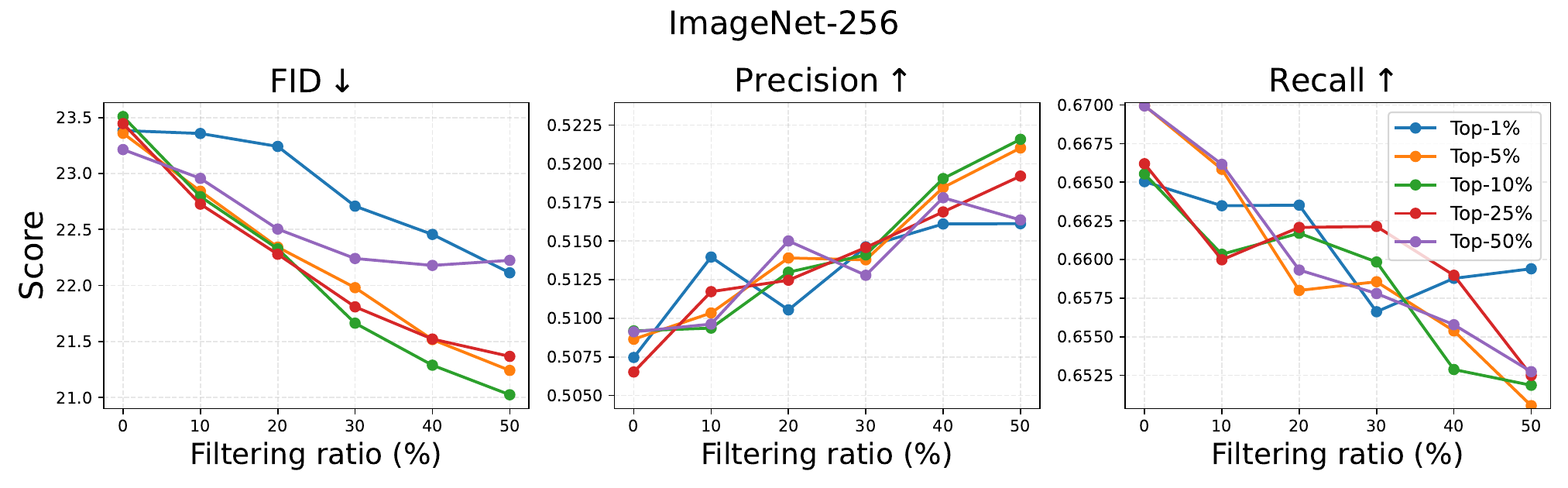}
		\caption{Top-$k\%$ aggregation ratio for the sample-level uncertainty score, $k\in\{1,5,10,25,50\}\%$.}
		\label{fig:filtering_uncertainty_aggregation_imagenet256}
	\end{subfigure}
	\caption[]{\textit{Ablations on UA-Flow's uncertainty estimation pipeline (continued).}}
\end{figure}

\begin{figure}[htbp]\ContinuedFloat
	\centering
	\begin{subfigure}[t]{0.8\textwidth}
		\centering
		\includegraphics[width=\textwidth]{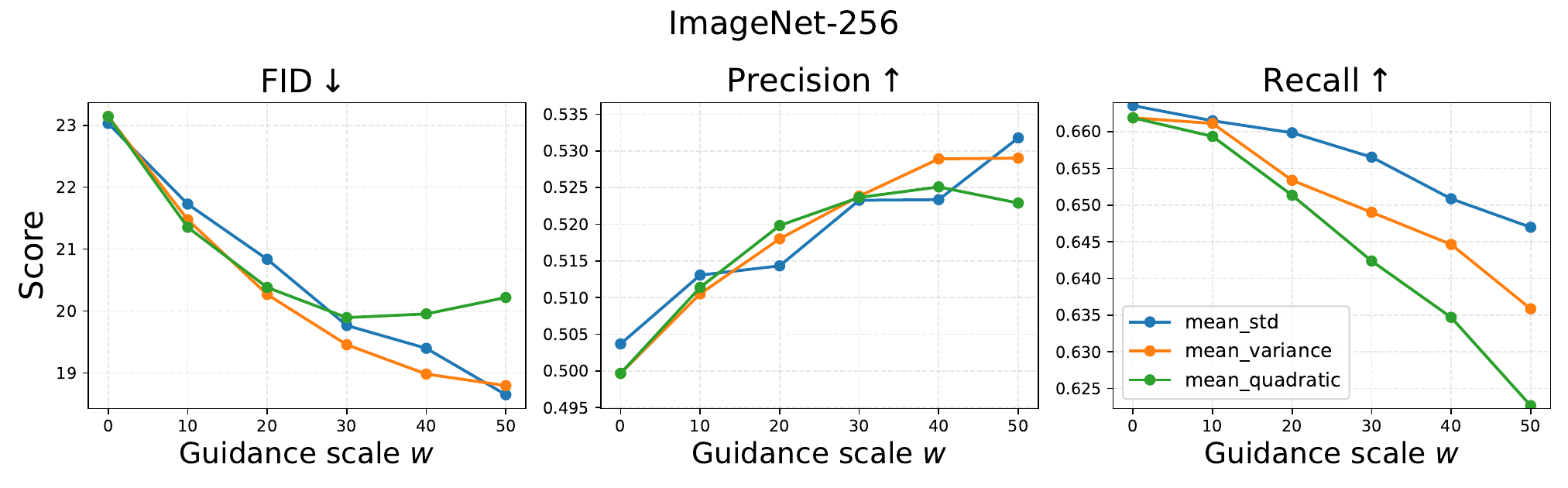}
		\caption{Guidance function $f(\sigma^2)$: (i) $-\big(\tfrac{1}{n}\sum_{i=1}^{n}\sigma_i\big)^2$ (\texttt{mean\_std})
		(ii) $-\big(\tfrac{1}{n}\sum_{i=1}^{n}\sigma_i^2\big)^2$ (\texttt{mean\_variance}, default)
		(iii) $-\big(\tfrac{1}{n}\sum_{i=1}^{n}\sigma_i^4\big)^2$ (\texttt{mean\_quadratic}).}
		\label{fig:ucg_f_sigma_imagenet256}
	\end{subfigure}\\[0.5ex]
	\begin{subfigure}[t]{0.8\textwidth}
		\centering
		\includegraphics[width=\textwidth]{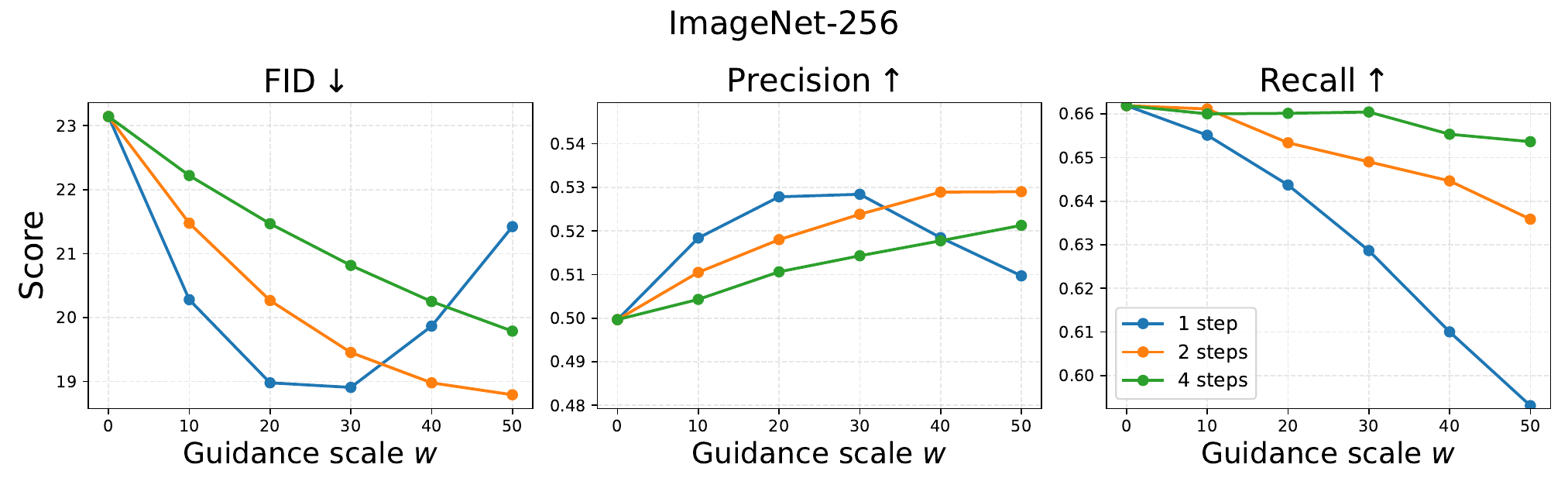}
		\caption{Guidance interval $\in\{1,2,4\}$ sampling steps.}
		\label{fig:ucg_interval_imagenet256}
	\end{subfigure}
	\caption[]{\textit{Ablations on UG (continued; ImageNet-256, UA-Flow, $\lambda=0$).}}
\end{figure}

We vary one component at a time while otherwise following the filtering protocol in \cref{subsec:filtering_exp} (ImageNet-256, UA-Flow, no guidance). \Cref{fig:uncertainty_ablations_combined} reports FID, precision, and recall.

\paragraph{Covariance approximation (\cref{fig:filtering_covariance_options_imagenet256}).}
We compare omitting the covariance term (Option 1), our one-probe JVP approximation (Option 2), and a $10$-sample Monte Carlo estimator (Option 3)~\citep{Kou23Bayesdiff}. Omitting covariance performs worst, whereas Options 2 and 3 yield similar curves. We therefore use Option 2, which requires one probe rather than ten velocity evaluations per propagation step.

\paragraph{Variance term $\Var(\bar{u}^\theta_t(\mathbf{x}_t))$ (\cref{fig:filtering_variance_term_imagenet256,fig:filtering_variance_term_scatter_imagenet256}).}
We compare the default predicted variance $(\sigma^\theta_t(\bar{\mathbf{x}}_t))^2$ with its sum with $\Var(\bar{u}^\theta_t(\mathbf{x}_t))$, estimated using $K=10$ Monte Carlo samples per step. Adding the latter leaves the filtering curves nearly unchanged despite a $K\times$ cost. Across $10^6$ locations and steps, it is small relative to the predicted term: $\Var(\bar{u}^\theta_t(\mathbf{x}_t)) \approx 0.036\,(\sigma^\theta_t(\bar{\mathbf{x}}_t))^2 + 0.01$ ($r=0.334$). We therefore omit it.

\paragraph{Number of Hutchinson probes $S$ (\cref{fig:filtering_random_probes_imagenet256}).}
We approximate the covariance with Hutchinson's estimator using $S\in\{1,2,4,8\}$ probes. The resulting curves are nearly identical, so we use the single-probe setting $S=1$.

\paragraph{Variance propagation method (\cref{fig:filtering_variance_propagation_imagenet256}).}
We compare the default first-order Euler update with a Heun2 update that averages variance increments at the current and predicted next states. They yield closely matched curves; Heun2 provides only a small FID gain at high filtering ratios. We use Euler because it requires one fewer uncertainty evaluation per step.

\paragraph{Sparse uncertainty updates (\cref{fig:filtering_sparse_uncertainty_imagenet256}).}
We propagate uncertainty every 2, 4, or 8 sampling steps, rather than at every step. All settings yield consistent trends. We use every-four-step updates, matching BayesDiff~\citep{Kou23Bayesdiff} and reducing propagation evaluations by $4\times$ relative to per-step updates without affecting filtering quality.

\paragraph{Uncertainty aggregation ratio (\cref{fig:filtering_uncertainty_aggregation_imagenet256}).}
We compare top-$k\%$ aggregation of the uncertainty map ($\mathrm{CVaR}_\alpha$) for $k\in\{1,5,10,25,50\}\%$, using $k=10\%$ by default. Results are stable for $k\in[5\%,25\%]$; $1\%$ emphasizes too few elements, whereas $50\%$ can dilute the tail signal.

\subsection{Ablations on UG}
\label{app:f_sigma}
\label{app:ucg_interval}

\paragraph{Effect of $f(\sigma^2)$ (\cref{fig:ucg_f_sigma_imagenet256}).}
UG uses the gradient of a scalar function $f$ of the predicted variance. We compare \texttt{mean\_std}, \texttt{mean\_variance} (default), and \texttt{mean\_quadratic}, which aggregate element-wise standard deviations, variances, and squared variances, respectively. \texttt{mean\_std} and \texttt{mean\_variance} yield similar curves, while \texttt{mean\_quadratic} produces slightly sharper FID and precision peaks. We use \texttt{mean\_variance}, indicating that UG is robust to the choice of $f$.

\paragraph{Effect of guidance interval (\cref{fig:ucg_interval_imagenet256}).}
We apply the UG correction in \cref{eq:uncertainty_classifier_guidance} every 1, 2 (default), or 4 sampling steps. All intervals improve FID over the unguided baseline. Applying UG every step reaches its optimum at $w=30$ but degrades at larger scales; the 2- and 4-step intervals shift the optimum upward without an apparent peak-out. We therefore apply UG every two steps.

\subsection{Effect of the Bias-Correction Term $U_t(\mathbf{x}_t, \mathbf{x}_1)$}
\label{app:bias_correction}

\begin{figure}[htbp]
	\centering
	\begin{subfigure}[t]{0.75\textwidth}
		\centering
		\includegraphics[width=\textwidth]{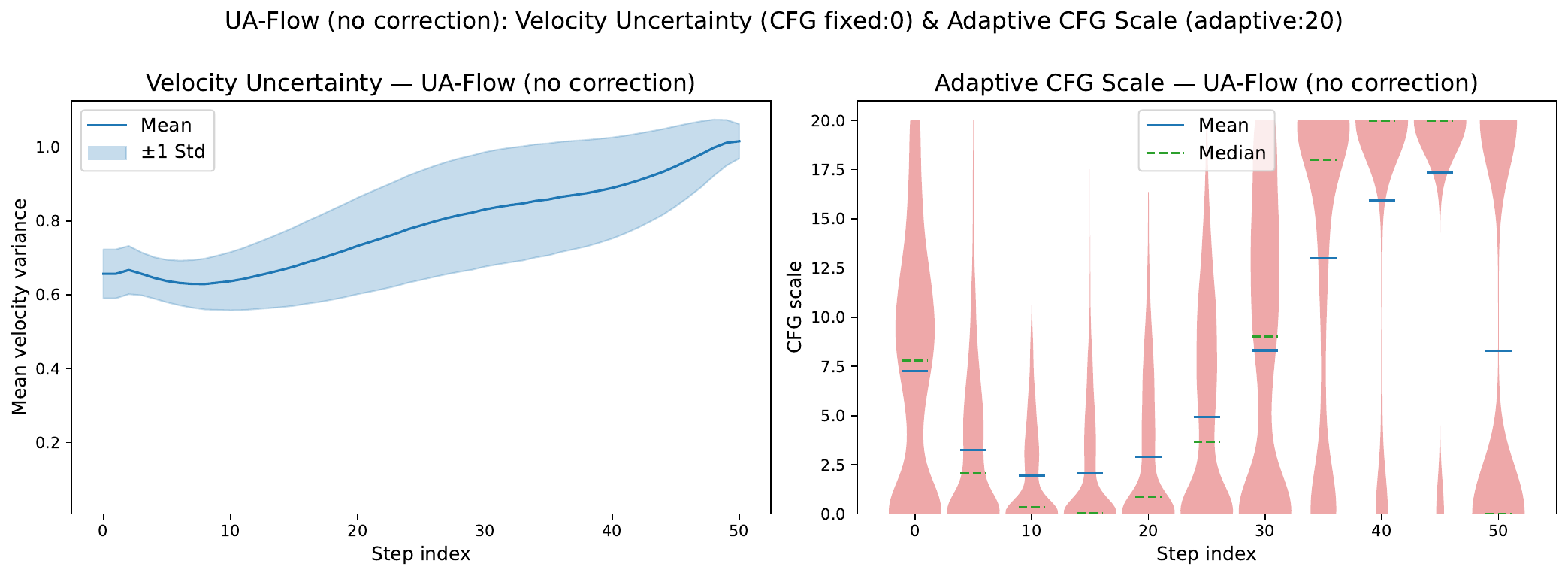}
		\caption{Without correction.}
		\label{fig:bias_correction_no}
	\end{subfigure}
	\begin{subfigure}[t]{0.75\textwidth}
		\centering
		\includegraphics[width=\textwidth]{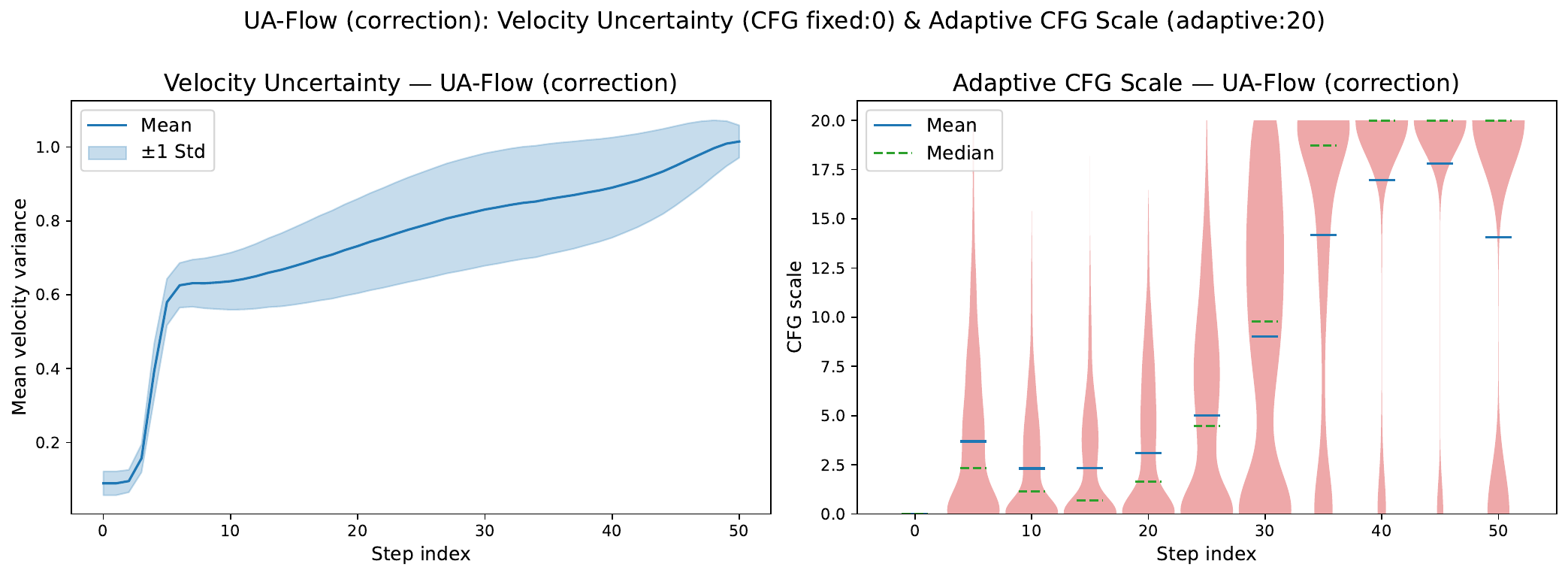}
		\caption{With correction (ours).}
		\label{fig:bias_correction_yes}
	\end{subfigure}
	\caption{
		\textbf{Velocity uncertainty magnitude ($\lambda=0$) and adaptive CFG scale ($\lambda_{\max}=20$) over sampling steps, with and without the bias-correction term $U_t$ (ImageNet-256, 1{,}000 images).}
		Without correction, the variance head overestimates uncertainty at the very early steps, while the median of the adaptive CFG scale is lower during the intermediate phase where coarse image structure is formed.
		With correction, velocity uncertainty is lower at the early steps and the adaptive CFG scale is higher during the intermediate phase.
	}
	\label{fig:bias_correction}
\end{figure}

\begin{table}[t]
	\centering
	\setlength{\tabcolsep}{3.5pt}
	\renewcommand{\arraystretch}{1.05}
	\caption{
		\textbf{Bias-correction ablation (ImageNet-256).}
		Metrics for the UA-Flow model trained with $U_t=0$. $^\dagger$Results are averaged over 3 seeds.
	}
	\label{tab:bias_correction}
	\begin{tabular}{@{}cccc@{}}
		\toprule
		Task & FID$\downarrow$ & Prec.$\uparrow$ & Rec.$\uparrow$ \\
		\midrule
		--- & 22.87 & 0.5063 & \textbf{0.6683} \\
		Filtering (50\%) & 20.66 & 0.5158 & 0.6462 \\
		UG ($\lambda{=}0$, $w{=}50$) & 18.44 & 0.5296 & 0.6389 \\
		Best UCFG$^\dagger$ ($\lambda_{\max}{=}1.25$) & $4.67_{\pm.05}$ & $0.7578_{\pm.0040}$ & $0.5063_{\pm.0029}$ \\
		Best fixed CFG$^\dagger$ ($\lambda{=}0.75$) & $\mathbf{4.53_{\pm.04}}$ & $\mathbf{0.7821_{\pm.0066}}$ & $0.4920_{\pm.0037}$ \\
		\bottomrule
	\end{tabular}
\end{table}

The bias-correction term in \cref{eq:cufm}, $U_t(\mathbf{x}_t, \mathbf{x}_1) := \hat{u}_t(\mathbf{x}_t)^2 - u_t(\mathbf{x}_t \mid \mathbf{x}_1)^2$, accounts for the gap between the marginal velocity estimate and the conditional target velocity. We train a matched UA-Flow checkpoint with $U_t=0$ and evaluate filtering, UG, and UCFG on ImageNet-256. A stop-gradient velocity-prediction alternative was excluded because training was unstable.

\paragraph{Filtering and UG.}
Without correction, filtering and UG still improve over the unguided baseline, with FID reductions comparable to the corrected model (filtering: $\Delta$2.21 vs.\ $\Delta$2.36; UG: $\Delta$4.43 vs.\ $\Delta$4.35). This suggests that their relative uncertainty rankings remain useful.

\paragraph{UCFG.}
The correction is critical for UCFG. Without it, the best UCFG (FID $4.67 \pm 0.05$ at $\lambda_{\max}{=}1.25$) does not surpass fixed CFG (FID $4.53 \pm 0.04$ at $\lambda{=}0.75$); with it, UCFG (FID $4.30 \pm 0.01$ at $\lambda_{\max}{=}1.5$) outperforms fixed CFG (FID $4.48 \pm 0.01$ at $\lambda{=}0.75$). \Cref{fig:bias_correction} shows that removing the correction overestimates uncertainty early but produces slightly weaker adaptive CFG during the structure-forming intermediate phase, limiting UCFG's benefit.

\section{Experiments on Non-Image Domains}
\label{app:non_image_experiments}

To examine whether UA-Flow's uncertainty estimation generalizes beyond image generation, we evaluate it on non-image domains where flow matching has been successfully applied.

\subsection{2D Checkerboard}
\label{app:non_image:checkerboard}

\begin{figure}[htbp]
	\centering
	\includegraphics[width=0.9\textwidth]{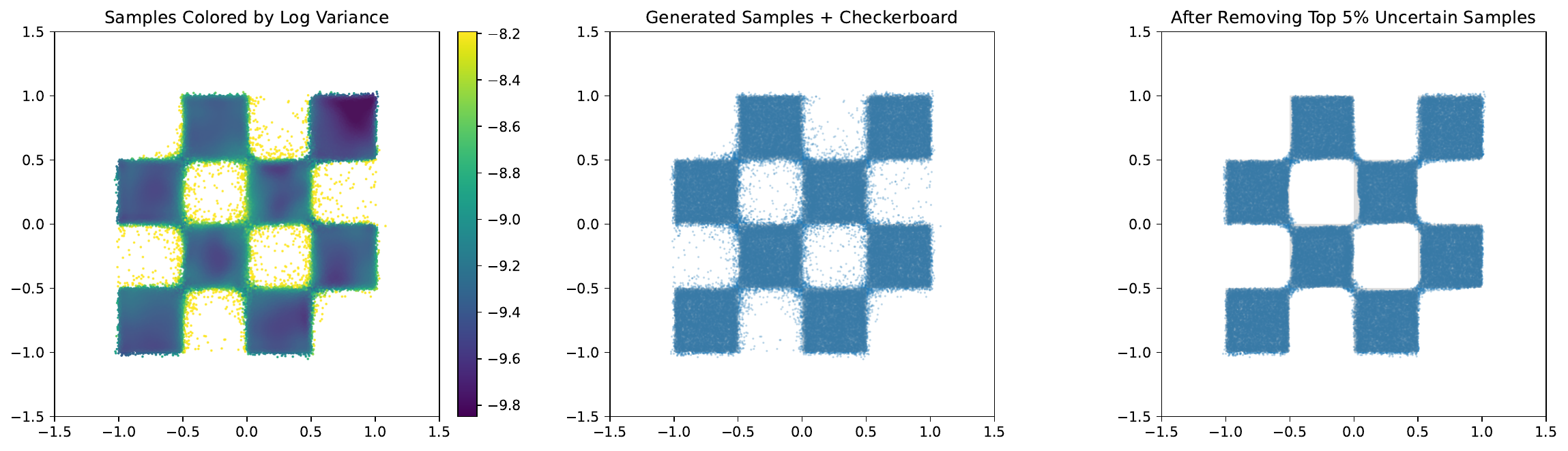}
	\caption{
		\textbf{Generated 2D checkerboard samples and uncertainty.}
		\emph{Left}: samples colored by log state variance.
		\emph{Middle}: generated samples overlaid on the ground-truth checkerboard grid.
		\emph{Right}: samples after removing the top 5\% highest-uncertainty samples.
		High-uncertainty samples concentrate near or outside the grid boundaries, and removing them visibly improves coverage.
	}
	\label{fig:uncertainty_heatmap}
\end{figure}

\begin{table}[t]
	\centering
	\caption{
		\textbf{Coverage after uncertainty-based filtering} on the 2D checkerboard.
		Coverage is the fraction of retained samples that fall inside the checkerboard grid.
		Filtering by UA-Flow's uncertainty monotonically improves coverage.
	}
	\label{tab:checkerboard_coverage}
	\begin{tabular}{lc}
		\toprule
		Filtering ratio & Coverage$\uparrow$ \\
		\midrule
		0\% (no filtering) & 0.9572 \\
		1\%   & 0.9668 \\
		2.5\% & 0.9793 \\
		5\%   & 0.9876 \\
		10\%  & \textbf{0.9936} \\
		\bottomrule
	\end{tabular}
\end{table}

\paragraph{Setup.}
We train a flow matching model on the 2D checkerboard distribution to verify that UA-Flow assigns high uncertainty to samples that fall outside the target support.
The velocity network is a 3-layer MLP trained with batch size 4{,}096 for 100{,}000 iterations at learning rate $10^{-3}$, followed by 100{,}000 fine-tuning iterations at learning rate $10^{-5}$.
At evaluation, we generate $10^5$ samples using the Euler method with 100 NFE and compute the sample-level uncertainty score by reducing the propagated element-wise state variance to a scalar.
We define \emph{coverage} as the fraction of retained samples that fall inside the checkerboard grid.

\paragraph{Results.}
\Cref{fig:uncertainty_heatmap} visualizes the generated samples colored by their log state variance.
High-uncertainty samples cluster near or outside the grid boundaries, while low-uncertainty samples concentrate well within the checkerboard cells.
Quantitatively, the mean uncertainty score of samples outside the grid is $8.0\times$ higher than that of samples inside.
This difference supports the use of the uncertainty score to identify out-of-support samples in this setting.

\Cref{tab:checkerboard_coverage} reports coverage after filtering out the most uncertain samples.
Without filtering, coverage is 95.7\%.
Removing the top 5\% highest-uncertainty samples raises coverage to 98.8\%, and removing 10\% reaches 99.4\%.
These results provide evidence that UA-Flow's uncertainty score can identify samples that deviate from the target support on a simple 2D distribution.

\subsection{Time Series}
\label{app:non_image:timeseries}

\begin{figure}[htbp]
	\centering
	\includegraphics[width=\textwidth]{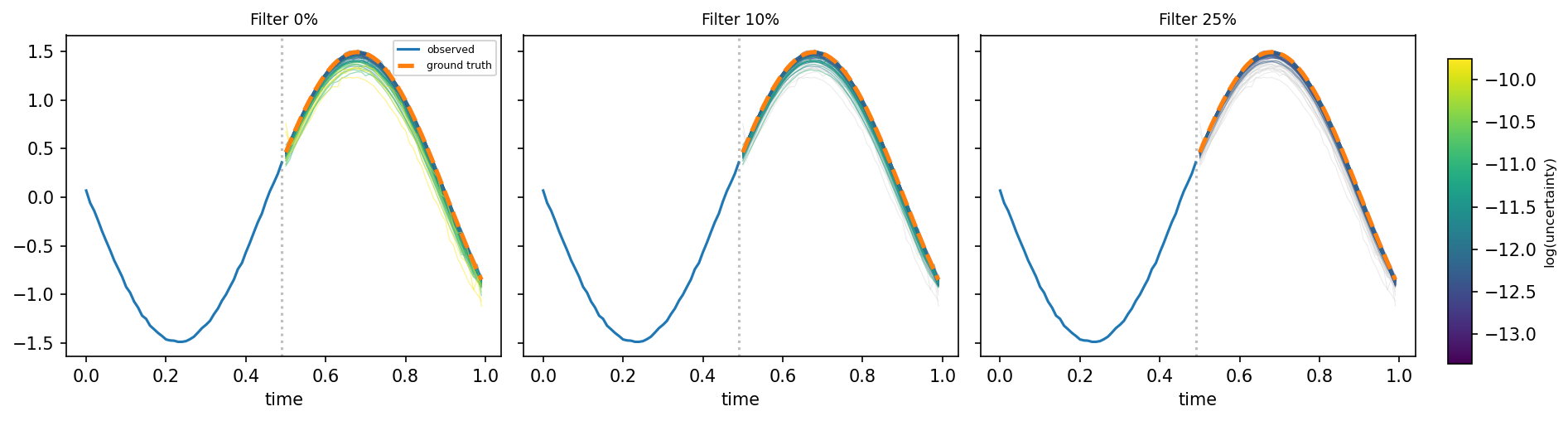}
	\caption{
		\textbf{Conditional time-series predictions and uncertainty-based filtering.}
		Each panel shows generated predictions (colored curves) given an observed input (blue) and ground truth curve (orange dash).
		From left to right: no filtering, 10\% filtering, and 25\% filtering.
		Removing high-uncertainty predictions eliminates incorrect forecasts.
	}
	\label{fig:timeseries}
\end{figure}

\begin{table}[htbp]
	\centering
	\caption{
		\textbf{MSE after uncertainty-based filtering} on the sine-wave time-series task.
		MSE is computed between each predicted trajectory and the ground-truth continuation, averaged over retained samples.
		Filtering by UA-Flow's uncertainty monotonically reduces prediction error.
	}
	\label{tab:timeseries_filtering}
	\begin{tabular}{lc}
		\toprule
		Filtering ratio & MSE ($\times 10^{-4}$)$\downarrow$ \\
		\midrule
		0\% (no filtering) & 14.65 \\
		10\%  & 3.87 \\
		25\%  & 2.11 \\
		50\%  & 1.77 \\
		90\%  & \textbf{0.78} \\
		\bottomrule
	\end{tabular}
\end{table}

\paragraph{Setup.}
We consider a conditional time-series prediction task on synthetic sine-wave data.
Each sequence is a noisy sine wave with frequency sampled uniformly from $[0.5, 2.0]$, amplitude from $[0.5, 1.5]$, and additive Gaussian noise with standard deviation $0.01$.
Sequences are discretized at $\Delta t = 0.01$.
The first 50 steps serve as the conditional input and the subsequent 50 steps as the prediction target.
The velocity network is a 1D UNet trained with batch size 4{,}096 for 40{,}000 iterations at learning rate $10^{-4}$, followed by 10{,}000 fine-tuning iterations at learning rate $10^{-5}$.
At evaluation, we draw 1{,}000 random conditional inputs and generate 100 predictions per input using the Euler method with 100 NFE.
The sample-level uncertainty score is computed by summing the propagated element-wise state variance over all predicted time steps.

\paragraph{Results.}
\Cref{tab:timeseries_filtering} reports MSE between predicted and ground-truth trajectories after filtering out the most uncertain predictions.
Filtering monotonically reduces MSE: removing the top 10\% cuts error by $3.8\times$ (from $14.65 \times 10^{-4}$ to $3.87 \times 10^{-4}$), and removing 90\% yields a further reduction to $0.78 \times 10^{-4}$.
\Cref{fig:timeseries} illustrates this qualitatively: without filtering, some trajectories diverge from the ground truth, whereas the retained predictions after filtering closely track the true continuation.
These results confirm that UA-Flow's uncertainty serves as a reliable per-sample quality signal for conditional time-series generation, consistent with the filtering behavior observed in the image and 2D checkerboard experiments.

\subsection{Robot Policy in Push-T Environment}
\label{app:non_image:pusht}

\begin{table}[htbp]
	\centering
	\caption{
		\textbf{Uncertainty-triggered early termination on Push-T robot manipulation.}
		During a rollout, if the state uncertainty of the generated action chunk exceeds a threshold $\tau$, the rollout is terminated and excluded from evaluation; otherwise the chunk is executed.
		Results are averaged over the rollouts that complete without early termination, out of $1{,}000$ total. A rollout is successful when its maximum reward exceeds 1.0.
	}
	\label{tab:pusht}
	\begin{tabular}{lcccccc}
		\toprule
		Threshold $\tau$ & Included & Excluded & Excl.\ \% & Success \%$\uparrow$ & Mean MaxR$\uparrow$ & Std MaxR \\
		\midrule
		no removal            & 1000 &   0 &  0.0\% & 62.7\% & 0.9514 & 0.1506 \\
		\midrule
		0.0010                &  122 & 878 & 87.8\% & \textbf{91.0}\% & \textbf{0.9702} & 0.1253 \\
		0.0025                &  371 & 629 & 62.9\% & 77.6\%          & 0.9597          & 0.1424 \\
		0.0050                &  708 & 292 & 29.2\% & 66.8\%          & 0.9536          & 0.1526 \\
		0.0075                &  854 & 146 & 14.6\% & 63.6\%          & 0.9536          & 0.1504 \\
		0.0100                &  921 &  79 &  7.9\% & 63.3\%          & 0.9521          & 0.1511 \\
		\bottomrule
	\end{tabular}
\end{table}

\paragraph{Setup.}
This preliminary study examines how chunk-level uncertainty relates to closed-loop outcomes in Push-T~\citep{Chi25Diffusionpolicy, Zhang24FP}, where a policy pushes a T-shaped block to a target pose. We use a 1D UNet conditioned on a $76\times76$ image and 2D agent position (observation horizon 2; prediction horizon 16; action chunk size 8), trained for 1{,}000 epochs with batch size 256 and learning rate $10^{-4}$.
At each decision step, we sum the propagated state variance across action dimensions to obtain an uncertainty score for one generated action chunk.
We test whether this score can serve as an online abort signal, demonstrating the potential of UA-Flow’s uncertainty estimates for robotics applications.


\paragraph{Uncertainty-triggered early termination.}
We terminate a rollout when its current chunk uncertainty exceeds a threshold $\tau$; \cref{tab:pusht} reports the retained rollouts among 1{,}000 trials. As $\tau$ tightens, both the exclusion rate and retained success rate increase. At $\tau=0.001$, excluding $87.8\%$ of rollouts raises retained success from the $62.7\%$ baseline to $91.0\%$.

\section{Samples and Uncertainties}
\label{app:samples_and_uncertainties}

\begin{figure}[t]
	\centering
	\setlength{\tabcolsep}{1pt}
	\begin{tabular}{@{}cc@{}}
		\includegraphics[width=0.46\textwidth]{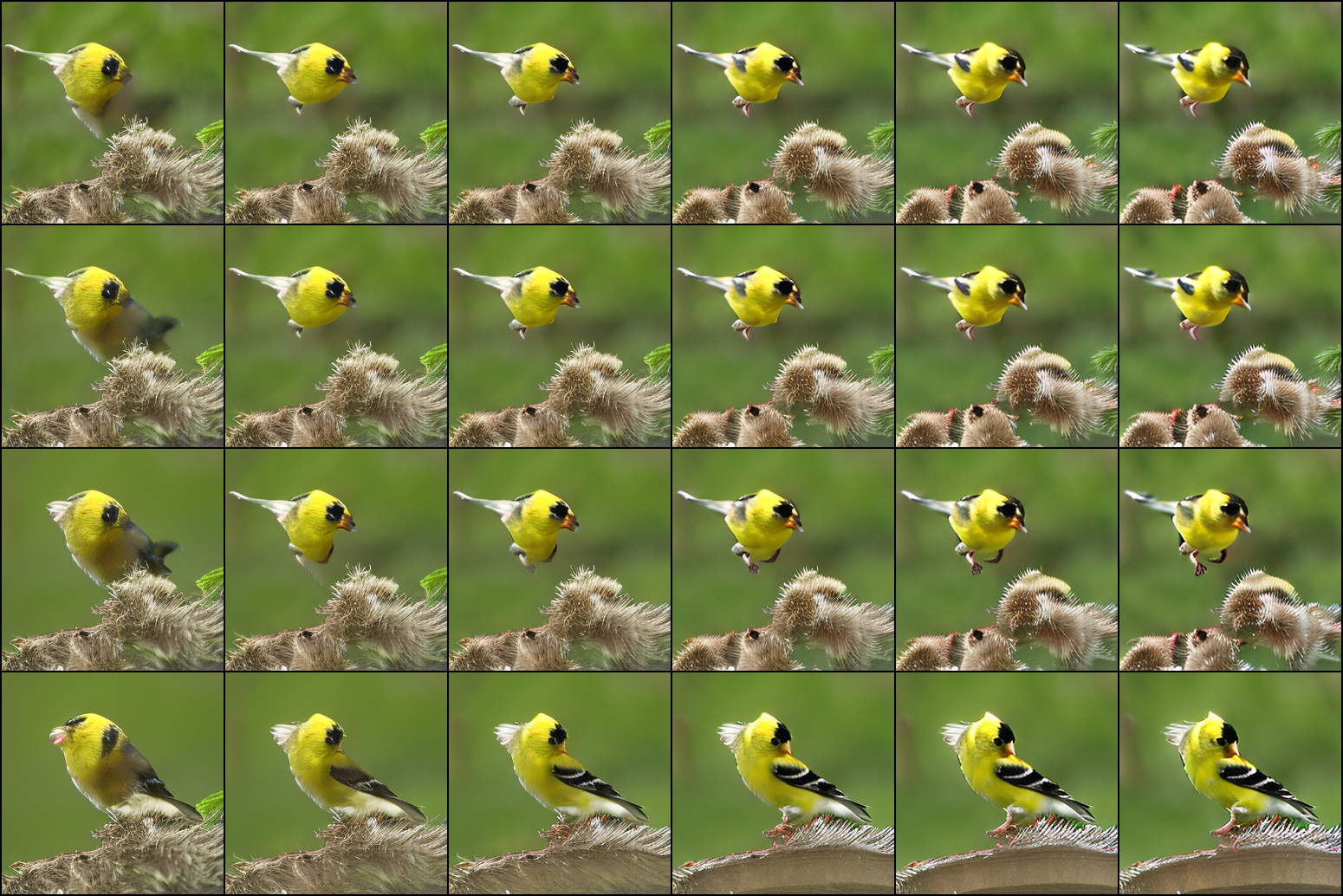} &
		\includegraphics[width=0.46\textwidth]{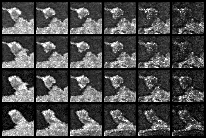} \\[-0.8ex]
		\includegraphics[width=0.46\textwidth]{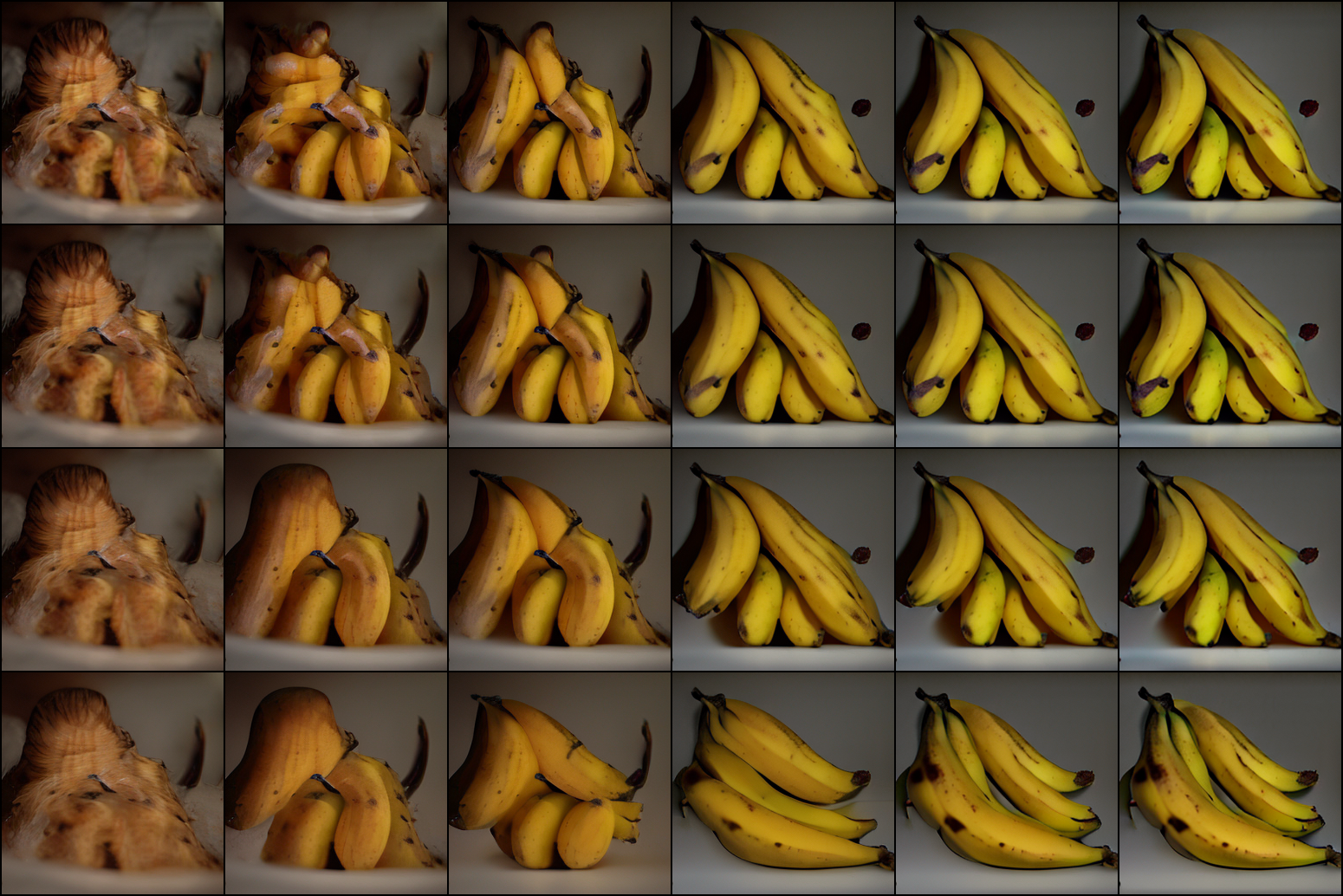} &
		\includegraphics[width=0.46\textwidth]{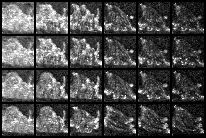} \\[-0.8ex]
		\includegraphics[width=0.46\textwidth]{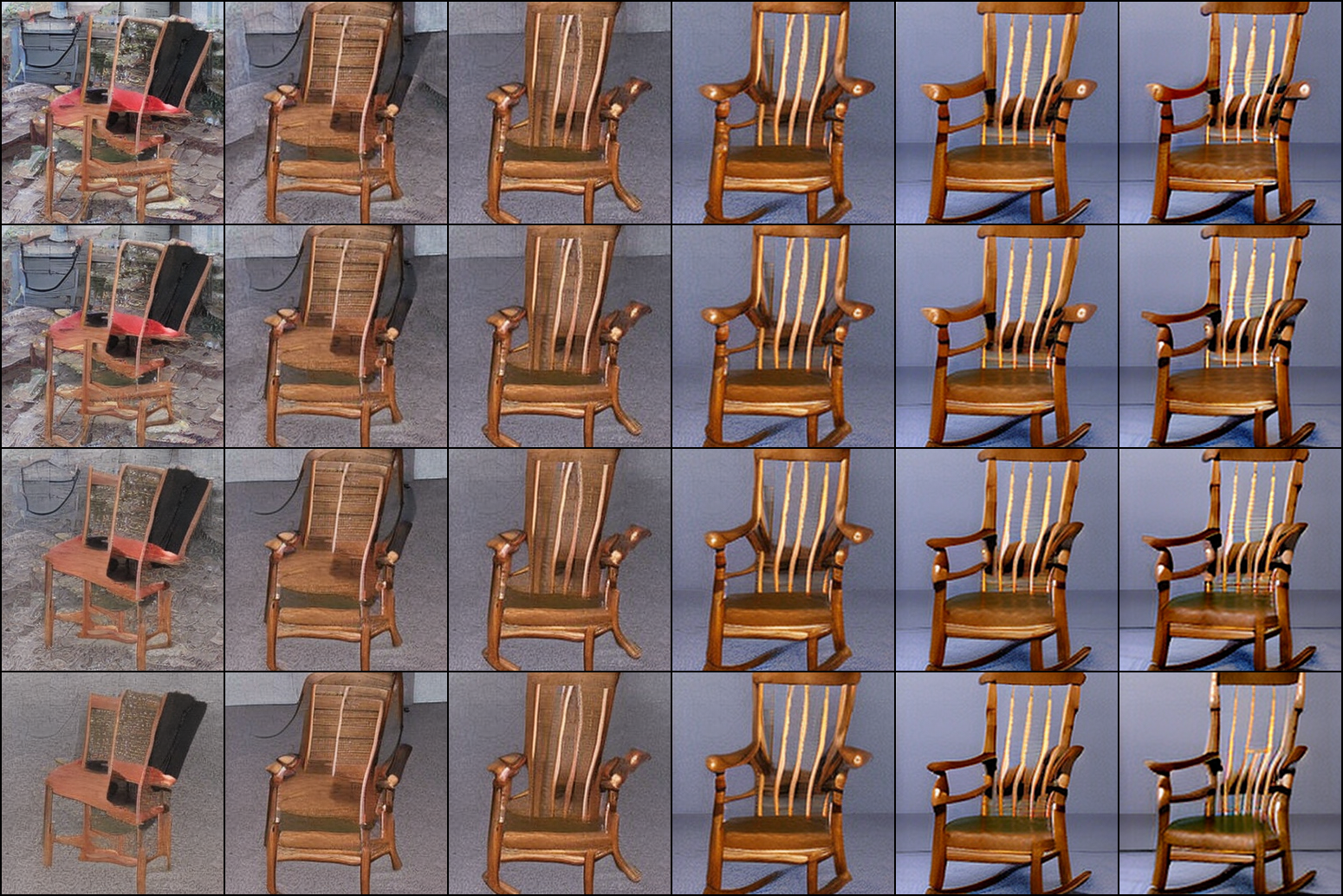} &
		\includegraphics[width=0.46\textwidth]{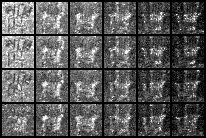} \\[-0.8ex]
		\includegraphics[width=0.46\textwidth]{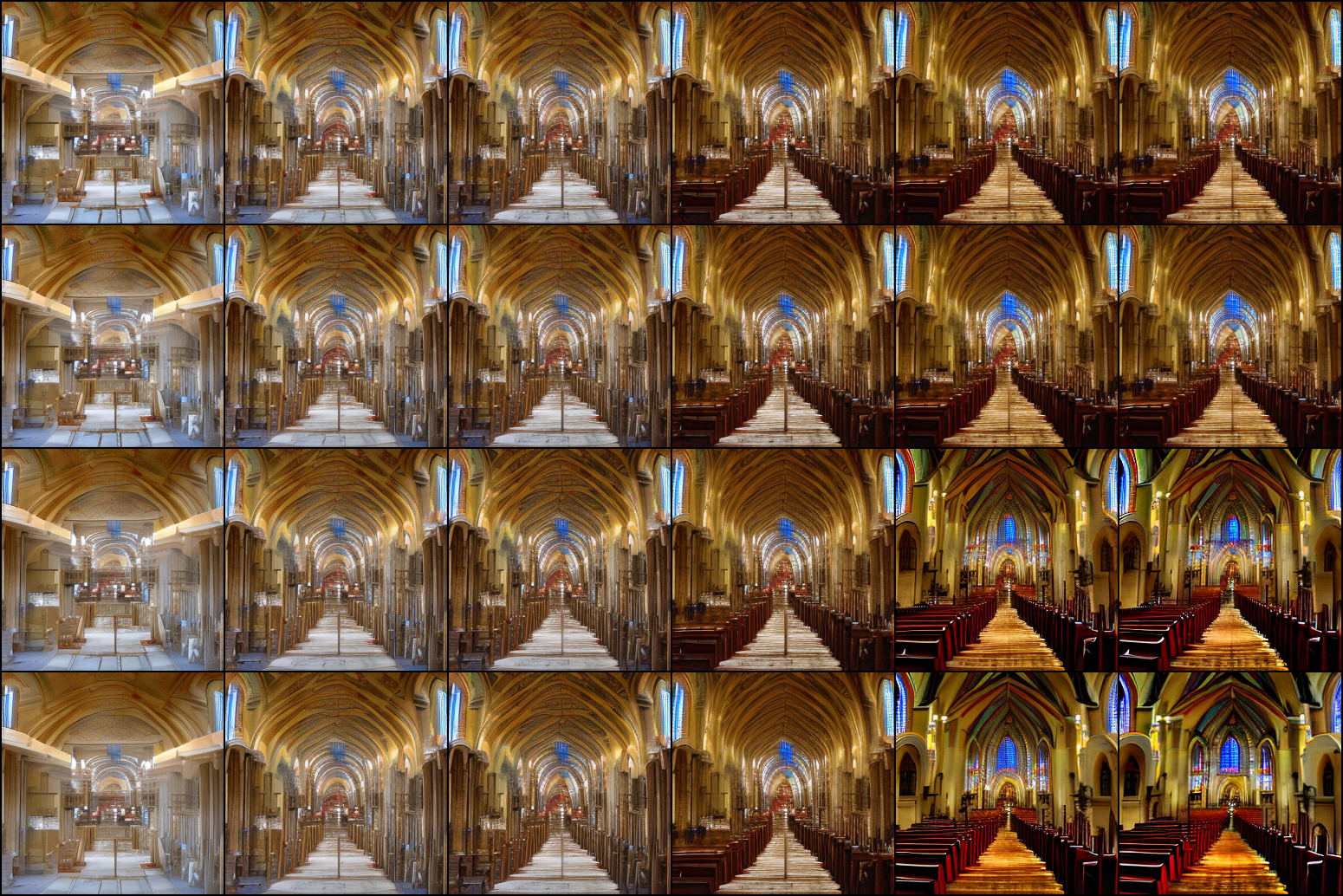} &
		\includegraphics[width=0.46\textwidth]{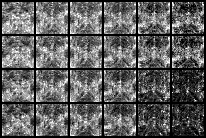} \\
	\end{tabular}
	\caption{
		\textbf{Selected ImageNet-256 sample grids under uncertainty-aware guidance sweeps.}
		Rows sweep the UG scale $w \in \{0, 10, 30, 50\}$ and columns sweep the maximum UCFG scale $\lambda_{\max} \in \{0, 1, 2, 5, 10, 20\}$.
		Left: generated samples. Right: predicted latent pixel-wise uncertainty maps.
	}
	\label{fig:samples_and_uncertainties}
\end{figure}

\end{document}